\documentclass{EJAM}
\usepackage{upgreek}
\usepackage{amsfonts}
\usepackage{amsmath}      
\usepackage{cleveref}
\usepackage{subcaption}
\usepackage{algorithm}
\usepackage{algorithmic}

\newsavebox{\astrutbox}
\sbox{\astrutbox}{\rule[-5pt]{0pt}{20pt}}

\makeatletter
\def\draftnote{\relax}      
\def\trimmarks{\relax}      
\makeatother

\makeatletter
\let\headerps@out\relax
\makeatother

\makeatletter

\renewcommand{\textit}[1]{{\fontshape{it}\selectfont #1}}
\renewcommand{\textbf}[1]{{\fontseries{bx}\selectfont #1}}
\makeatother

\begin{document}

\makeatletter
\let\ps@myplain\ps@empty
\makeatother
\renewcommand{\printhistory}{}

\lefttitle{Wen {\it et al.}}
\righttitle{}

\papertitle{PREPRINT}

\title{$\beta$-number and Its Applications to Data Science}

\author{Pengxu Wen$^1$}
\author{Ziwei Nie$^1$}
\author{Liwen Zou$^1$}
\author{Ying Cao$^1$}
\author{Xiaoping Yang$^1$}

\affil{$^1$School of Mathematics, Nanjing University, Nanjing, P.R. China\\
\textbf{Corresponding author:} Ziwei Nie and Xiaoping Yang; Email: \href{mailto:nieziwei@nju.edu.cn}{nieziwei@nju.edu.cn}; \href{mailto:xpyang@nju.edu.cn}{xpyang@nju.edu.cn}}

\begin{abstract}
In this paper, we introduce a novel metric, $\beta$-number, designed to measure the "$p$-th order $d$-dimensional oscillation degree" of a set. 
For a given set $K$ in the Euclidean space $\mathbb{R}^n$, $\beta$-number is computed by identifying an optimal $d$-dimensional plane $L$ that minimizes the average $p$-th order distance of all points in $K$ to $L$. This metric serves as a versatile tool in data science, functioning both as a dimensionality reduction technique to find the optimal $p$-th order $d$-dimensional subspace and as a measure to characterize data oscillation for classification. 
We conduct a thorough exploration of $\beta$-number through theoretical analysis and numerical experiments. Initially, we establish the upper and lower bounds of $\beta$-number for any set $K$, along with its scaling law. We then elucidate the relationships between $\beta$-number and its related parameters. 
A key finding of this study is that in discrete settings, when $p=2$, the $d$-dimensional plane determined by $\beta$-number coincides with that obtained via Principal Component Analysis (PCA), positioning $\beta$-number as a generalization of PCA. 
Furthermore, we propose two algorithms for calculating $\beta$-number in both continuous and discrete settings. We validate our theoretical results through a series of numerical experiments in low-dimensional cases. 
Finally, we demonstrate the potential of $\beta$-number in practical data analysis through a dimensionality reduction task and an image classification task.
\end{abstract}


\begin{msc} 
62H25 (Primary); 65D18, 68U05, 62H30, 15A60 (Secondary)
\end{msc}

\maketitle

\section{Introduction}
\subsection{Definition of $\beta$-number}

In 1990, in order to study the 1-dimensional Traveling Salesman Problem, Peter Jones introduced an indicator describing the flatness of infinite sets, known as the Jones's $\beta$-number \cite{Jones_1990}. It provides a multiscale geometric criterion that determines whether a set $E \subseteq \mathbb{C}$ can be contained within a curve of finite length. For a set $K$ lying in $\mathbb{R}^n$ or in a Hilbert space $H$, and for a cube or ball $Q$, the $\beta$-number defined by Jones is as follows:
\begin{align}
    \beta^{\infty}_{K}(Q) 
    &=\dfrac{\text { width of thinnest cylinder containing } K \cap Q}{\operatorname{diam}(Q)}  \label{Eq 1} \\
    &=\dfrac{2}{\operatorname{diam}(Q)} \inf _{L \operatorname{line}} \sup _{x \in K \cap Q} \operatorname{dist}(x, L) \label{Eq 2}\text {. }
    \end{align}

In 1991, David and Semmes, in their investigation of the boundedness of singular integral operators $T_f(x)$ on a set $E \subset \mathbb{R}^n$, explained that Jones's $\beta$-number is the $L^\infty$ version, and introduced a generalization as an $L^p$ version of $\beta$-number \cite{David_Semmes_1991}. For a set $K$ lying in $\mathbb{R}^n$ or a Hilbert space $H$, given $1 \leq p < \infty$, a closed ball $B(o,r) \subset \mathbb{R}^n$, and an integer $0 < d < n$, the $L^p$ version of $\beta$-number is defined as follows:
\begin{equation}\label{Eq 3}
        \beta^{p}_{K}(B(o,r))
        =\inf _{L}\left(\frac{1}{r^{d}} \int_{K \cap B(x, r)}\left(\frac{\operatorname{dist}(y, L)}{r}\right)^{p}d y\right)^{\frac{1}{p}},
    \end{equation}
where the infimum is taken over all $d$-dimensional planes $L$. 

Over time, $\beta$-number has become an important metric in geometric analysis. In order to apply $\beta$-number in data analysis, we propose a modified definition of $\beta$-number:
\begin{definition}\label{Def 1}
    Consider a set $K \subseteq \mathbb{R}^n$, which can be a discrete point set with $m$ elements or a bounded set with Hausdorff dimension $k$, where $0 < m < \infty$ and $1\leq k \leq n$.
    Let $B(o, r)$ denote the minimal enclosing sphere that contains all points in the set $K$, where $o$ represents the center and $r$ denotes the radius. Assume $1 \leq d \leq n$ and $1 \leq p < \infty$, where $k, d, p$ are integers. Then, the $L^ \infty$ version of $\beta$-number is defined as follows:
\begin{align}
    \beta_{\infty}(K)
    &=\dfrac{\text { width of thinnest cylinder containing } K }{\operatorname{diam}(B)}
     \label{Eq 4} \\
     &=\dfrac{1}{r} \inf _{L \operatorname{line}} \sup _{x \in K } \operatorname{dist}(x, L)
     \text {. } \label{Eq 5}
\end{align}
The $L^p$ version of $\beta$-number for both continuous and discrete cases is defined as follows:
\begin{equation}\label{Eq 6}
    \beta_{p,d}(K) \colon=
    \inf _{L}\beta_{p,d}(K,L) \colon=
    \left\{
    \begin{aligned}
    &\inf _{L}\left(\frac{1}{V(K) \cdot r^{d}} \int_{K}\left(\frac{\operatorname{dist}(y, L)}{r}\right)^{p} d y\right)^{\frac{1}{p}} && \text{if } \dim_{H}(K)=k \\
    &\inf _{L}\left(\frac{1}{m \cdot r^{d}} \sum_{i=1}^{m}\left(\frac{\operatorname{dist}\left(y_{i},  L\right)}{r}\right)^{p}\right)^{\frac{1}{p}} && \text{if\ \ } |K| = m
    \end{aligned}
    \right.
\end{equation}
where the infimum is taken over all d-dimensional planes $L$, and $V(K)$ denotes the volume of the set $K$ with Hausdorff dimension $k$.
\end{definition}

\begin{remark}
    In \cref{Def 1}, the $L^\infty$ version of $\beta$-number represents the limiting case of the $L^p$ version as $p$ approaches infinity for $d=1$.
\end{remark}

\subsection{A brief survey related to $\beta$-number}
Peter Jones first proposed the concept of $\beta$-number and provided a simple necessary and sufficient condition for a bounded set $K \subset{\mathbb{C}}$  to lie on a rectifiable curve \cite{Jones_1990}. David and Semmes then proposed the $L ^ p$ version of $\beta$-number to find a quantitative characterization of the qualitative concept of "rectifiability" \cite{David_Semmes_1991}. The development of quantitative rectifiability in the pioneering works of Jones, David, and Semmes has led to the study of the connection between rectifiability and the boundedness of square functions and singular integrals \cite{Tolsa_2015}. 
In 1992, Okikolu extended Jones' results from $K \subset{\mathbb{R}^2} $ to $K \subset{\mathbb{R}^n}$ ($n \geq 2$) \cite{Okikiolu_1992}. Raanan Schul conducted a Hilbert space version of this theory in 2007, which is equivalent to obtaining dimension free estimates \cite{Schul_2007}. These results formed the basis of a theory now called "quantitative rectifiability", which has been extended by many authors
\cite{Azzam_Tolsa_2015_2,Badger_Schul_2016,David_Semmes_1993,Lerman_2003}. More recently, many studies have focused on using $\beta$-number and its variants to delineate when measures and sets in Euclidean space are rectifiable, further applying this to geometric measure analysis \cite{BadgerLiZimmerman+2023,Badger_Naples_Vellis_2019,Hyde_2024,High-dimensional2,High-dimensional1,Nin_Obinna_2022,Tolsa_2015}.

Notably, in the fields of subspace approximation and coreset construction, a wealth of existing studies have centered on the core goal of "finding a $d$-dimensional plane $L$ to minimize the $L^p$-distances from all points in a set $K$ to $L$", leading to the establishment of systematic theoretical frameworks and efficient algorithmic systems for subspace approximation. It is worth noting that this optimization objective shares a certain similarity with the core optimization problem in the definition of our $\beta$-number \cite{clarkson2015input,sohler2018strong,tukan2022new,woodruff2014sketching}. However, the core focus of our work differs fundamentally from the aforementioned studies: we introduce the classical $\beta$-number concept (originating from the field of geometric measure theory) into data science, innovatively defining it as a quantitative indicator for characterizing the "intrinsic geometric oscillation degree" of datasets, with the primary aim of developing practical analytical tools tailored to data science scenarios.

\subsection{Aim and structure of this paper}
   Although a large number of  studies have focused on quantitatively characterizing rectifiability with $\beta$-number, there is a notable lack of research dedicated to the analysis of $\beta$-number itself. The definition, calculation, and potential applications of $\beta$-number remain largely unclear. Meanwhile, despite the certain similarity between the optimization objective in subspace approximation and $\beta$-number, no existing studies have introduced $\beta$-number as a geometric feature indicator into data science scenarios. The aims of this paper are as follows: first, to conduct a comprehensive theoretical analysis of the proposed $\beta$-number, with an emphasis on clarifying its ability to capture specific data features; second, to develop efficient algorithms for identifying optimal subspaces and computing $\beta$-number, and to carry out a comparative study with Principal Component Analysis (PCA); additionally, to verify the practical utility of $\beta$-number in various data science tasks, clarify its advantages in data reduction and classification, and explore its potential as a tool for characterizing data features.

The structure of the paper is as follows. First, we conduct a mathematical analysis in \cref{Sect2}, including the lower and upper bounds of $\beta$-number, the scaling law of $\beta$-number, and the relationships between $\beta$-number and its related parameters. In \cref{Sect3}, we describe the data features reflected by $\beta$-number and define it as the "$p$-th order $d$-dimensional oscillation degree" of a set $K$. Subsequently, we compare our $\beta$-number with principal component analysis (PCA) and derive some interesting conclusions. After that, we propose numerical computation algorithms in \cref{Sect4}. Moreover, we conduct various experiments in \cref{Sect5}. For low-dimensional cases, we present the visualization experimental results of the optimal subspace $L$ during the $\beta$-number computation. 
Subsequently, we validate the aforementioned conclusions through numerical experiments to further elucidate the significance of $\beta$-number. Finally, in \cref{Sect6}, we apply $\beta$-number to tasks involving data dimensionality reduction and image classification, exploring its potential as both a dimensionality reduction technique and a novel metric for characterizing data features.

\section{Theoretical Analysis}\label{Sect2}
In this section, we first analyze the lower and upper bounds of the two forms of $\beta$-number in \cref{subsec 2.1}. For the $L^\infty$ version of $\beta$-number, the lower and upper bounds can be easily obtained from \cref {Def 1}. For the $L^p$ version of $\beta$-number, there are many variables that affect its value, and we can obtain certain results when specific variables are determined. We present the scaling law of $\beta$-number in \cref{subsec 2.2}. The relationships between the values of $\beta$-number and its parameters will be discussed in \cref{subsec 2.3}.

\subsection{Lower and upper bounds of $\beta$-number}\label{subsec 2.1}
\begin{theorem}
If $K$ is a finite discrete point set in $\mathbb{R}^n$ or a bounded set with Hausdorff dimension $k$ in $\mathbb{R}^n$, then
    \begin{equation}
     0 \leq\beta_{\infty}(K) \leq 1.
 \end{equation}
\end{theorem}
\begin{proof}
    From (\ref{Eq 4}), since $B$ is the minimum enclosing ball of $K$, it follows from the properties of enclosing shapes that the width of the narrowest cylinder containing $K$ must be less than or equal to the diameter of $B$. This establishes the inequality $0 \leq \beta_{\infty}(K) \leq 1$. 
    On one hand, when $K$ is concentrated on a line, according to (\ref{Eq 5}), the optimal line $L$ is the line along which $K$ is concentrated, and $\beta_{\infty}(K)$ achieves its infimum of 0. On the other hand, when $K$ is a ball with Hausdorff dimension $n$ or a sphere with Hausdorff dimension $n-1$ in $\mathbb{R}^n$, the value of $\beta_{\infty}(K)$ reaches its supremum of 1.
\end{proof}
The value of the $L^p$ version of the $\beta$-number is influenced by several factors, including the Hausdorff dimension and data distribution of the set $K$, the dimension $d$ of $L$, and the parameter $p$. The interplay among these factors results in a complicated upper bound for the $L^p$ version of $\beta$-number.
\begin{theorem}\label{The 2.2}
If $K \subseteq \mathbb{R}^n$ is a bounded set with Hausdorff dimension $k$ or a finite discrete point set, for given $p, d$, then
\begin{equation}
     0 \leq\beta_{p,d}(K) < \left(\frac{1}{r^d}\right)^{\frac{1}{p}}.
\end{equation}
\end{theorem}
\begin{proof}
First, we consider the lower bound of $\beta_{p,d}$. If all points in $K$ are concentrated in a $d$-dimensional linear subspace (or a lower-dimensional linear subspace), according to (\ref{Eq 6}), $K$ can be embedded into some $d$-dimensional linear subspace $L$. Therefore, for any point $y$ in $K$, we have $\operatorname{dist}(y, L) = 0$. In this case, the $L^p$ version of $\beta$ number for $K$ achieves its infimum of 0, thus:
\begin{equation*}
    \beta_{p,d}(K) \geq 0.
\end{equation*}
Next, we consider the upper bound of $\beta_{p,d}$. If $K$ is a bounded set, for the integral $\int_{K} \left(\frac{\operatorname{dist}(y, L)}{r}\right)^{p} dy$, by the mean value theorem for integrals and properties of $L$, although there may be some points in $K$ such that $\operatorname{dist}(y,L) > r$, the average value of $\frac{\operatorname{dist}(y, L)}{r}$ can still remain below 1, thus satisfying the integral inequality:
\begin{equation*}    
 \int_{K}\left(\frac{\operatorname{dist}(y, L)}{r}\right)^{p} dy < V(K).
\end{equation*}
If $K$ is a finite discrete point set, similarly, it follows that:
\[
\sum_{i=1}^{m}\left(\frac{\operatorname{dist}\left(y_{i}, L\right)}{r}\right)^{p} < m.
\]
Therefore, we have:
\begin{equation*}
    \beta_{p,d}(K) < \left(\frac{1}{r^d}\right)^{\frac{1}{p}}.
\end{equation*}
\end{proof}

Furthermore, when $k = n$, if the smallest enclosing ball of $K$ is $K$ itself, the symmetry of the sphere ensures that for any given $p$ and $d$, the center of the sphere must lie on the $d$-dimensional plane $L$. In low-dimensional spaces, the $\beta$-number of $K$ can be directly obtained by integrating.

For instance, when $n=k=2$, $d=1$, suppose $K$ is a disk with a radius of $r$ in $\mathbb{R}^2$. For any given $p$, $L$ is any straight line passing through the center of the circle. Thus, we can integrate directly to obtain:
\begin{equation}\label{Eq.9}
\begin{aligned}
    \beta_{p,1}(K)
    &=\inf _{L}\left(\frac{1}{\pi r^2 \cdot r^{1}} \int_{K}\left(\frac{\operatorname{dist}(y, L)}{r}\right)^{p} d y\right)^{\frac{1}{p}}
    =\left(\frac{1}{\pi r^3} \int_{0}^{r} \int_{0}^{2 \pi} \rho \cdot\left(\frac{\rho|\sin \theta|}{r}\right)^{p} d \theta d \rho\right)^{\frac{1}{p}}\\
    &=\left(\frac{1}{\pi(p+2)r} \cdot\left(\int_{0}^{2 \pi}|\sin \theta|^{p} d \theta\right)\right)^{\frac{1}{p}}.
\end{aligned}
\end{equation}
When $p$ and $r$ are any given positive integers, the above expression is a constant. 

Similarly, when $n=k=3$, suppose $K$ is a ball with a radius of $r$ in $\mathbb{R}^3$. For any given $p$, it can be concluded that:
\begin{equation}\label{Eq 10}
     \beta_{p,1}(K) =
     \left(\frac{3}{r^4} \cdot \int_{0}^{r}\left(\frac{1}{p+2}\left(1-z^{2}\right)^{\frac{p+2}{2}}\right) d z\right)^{\frac{1}{p}},
\end{equation}
\begin{equation}\label{Eq 11}
     \beta_{p,2}(K) =
     \left(\frac{3}{2\pi r^{5}} 
     \int_{0}^{2 \pi}|\sin \theta|^{p} d \theta\cdot \int_{0}^{r}\left(\frac{1}{p+2}\left(1-x^{2}\right)^{\frac{p+2}{2}}\right) d x\right)^{\frac{1}{p}}.
 \end{equation}
 
\subsection{The scaling law of $\beta$-number}\label{subsec 2.2}
Let $K_1$ be a set in $\mathbb{R}^n$, which can either be a bounded set with Hausdorff dimension of $k$ or a finite discrete point set. Denote the minimal enclosing ball of $K_1$ as $B_1$, with its center $o_1$ and radius $r_1$. Scaling each point in the set $K_1$ by a factor of $a$ with respect to the center $o_1$, where $a$ is a positive constant, we obtain the set $K_2$. Specifically, this means multiplying the distance of each point from \(o_1\) by \(a\). Consequently, the radius of the minimal enclosing ball of $K_2$ is then $r_2 = a r_1$. In this subsection, we will discuss the relationship between the $\beta$-numbers of $K_1$ and $K_2$.

First, for the $L^\infty$ version of $\beta$-number, we have the following result:
\begin{theorem}
For any set $K_1$ in $\mathbb{R}^n$, which can be a bounded set with Hausdorff dimension $k$ or a finite discrete point set, let $o_1$ be the center of the minimum enclosing ball of $K_1$ with radius $r_1$. Scaling $K_1$ by a positive factor of $a$ with respect to $o_1$ yields the set $K_2$. Then:
\begin{equation}
    \beta_{\infty}(K_1) = \beta_{\infty}(K_2).
\end{equation}
\end{theorem}
\begin{proof}
    Since $K_2$ is obtained by scaling $K_1$ by a factor of $a$, the width of the narrowest cylinder containing $K_2$ is also $a$ times the width of the narrowest cylinder containing $K_1$. Meanwhile, $r_2 = ar_1$. According to (\ref{Eq 4}), the $L^{\infty}$ form of the $\beta$-number remains invariant under scaling, i.e., $\beta_{\infty}(K_1) = \beta_{\infty}(K_2)$.
\end{proof}
Next, for the $L^p$ version of $\beta$-number we have the following result:
\begin{theorem}
If $K_{1} \subseteq \mathbb{R}^n$ is a bounded set with Hausdorff dimension $k$ or a finite discrete point set, let $o_1$ be the center of the minimum enclosing ball of $K_1$ with radius $r_1$. Scaling $K_1$ by a positive factor of $a$ with respect to $o_1$ yields the set $K_2$. For given $p, d$, then:
\begin{equation}
    \beta_{p,d}(K_2) = a^{-d/p} \cdot \beta_{p,d}(K_1).
\end{equation}
\end{theorem}
\begin{proof}
For the $L^p$ version of $\beta$-number in the continuous case, we proceed with the proof in the following two steps.

\textbf{Step 1:}  
We begin by proving that for sets $ K_1 $ and $ K_2 $, if $ L $ is any $ d $-dimensional plane, and $ L^{'} $ is obtained by scaling $ L $ by a factor of $ a $ around $ o_1 $, then:
\begin{equation*}
    \beta_{p,d}(K_2, L^{'}) = a^{-d/p} \cdot \beta_{p,d}(K_1, L).
\end{equation*}
First, for any point $ y_1 \in K_1 $, its corresponding scaled point $ y_1^{'} \in K_2 $ satisfies:
\begin{equation*}
\frac{\operatorname{dist}(y_1^{'}, L^{'})}{r_2} = \frac{a \cdot \operatorname{dist}(y_1, L)}{a \cdot r_1} = \frac{\operatorname{dist}(y_1, L)}{r_1}.
\end{equation*}
Since the points in $ K_1 $ and $ K_2 $ are in one-to-one correspondence, we have:
\begin{equation*}
\frac{1}{V(K_2)} \cdot \int_{K_2}\left(\frac{\operatorname{dist}(y, L^{'})}{r_2}\right)^p d y = \frac{1}{V(K_1)} \cdot \int_{K_1}\left(\frac{\operatorname{dist}(y, L)}{r_1}\right)^p d y.
\end{equation*}
Using the definition of the $\beta$-number:
\begin{equation*}
\beta_{p,d}(K_2, L^{'}) = \left(\frac{1}{r_2^d} \cdot \frac{1}{V(K_2)} \int_{K_2}\left(\frac{\operatorname{dist}(y, L^{'})}{r_2}\right)^p d y\right)^{1/p}.
\end{equation*}
Substituting $ r_2 = a \cdot r_1 $ and simplifying:
\begin{equation*}
\beta_{p,d}(K_2, L^{'}) = \left(\frac{1}{(a \cdot r_1)^d} \cdot \frac{1}{V(K_1)} \cdot \int_{K_1}\left(\frac{\operatorname{dist}(y, L)}{r_1}\right)^p d y\right)^{1/p} = a^{-d/p} \cdot \beta_{p,d}(K_1, L).
\end{equation*}
\textbf{Step 2:}  
We aim to prove that if $ L_1 $ and $ L_2 $ are the optimal $ d $-dimensional planes for $ K_1 $ and $ K_2 $, respectively, then the plane $ L_1^{'} $ obtained by scaling $ L_1 $ with respect to the center $ o_1 $ is equivalent to the optimal $ d $-dimensional plane $ L_2 $ for $ K_2 $, i.e.,
\begin{equation*}
\beta_{p,d}(K_2, L_1^{'}) = \beta_{p,d}(K_2, L_2).
\end{equation*}
Let $ L_2^{'} $ be obtained by scaling $ L_2 $ by a factor of $ 1/a $ around the center $ o_1 $. By the definition of the $\beta$-number, we have $\beta_{p,d}(K_1, L_1) \leq \beta_{p,d}(K_1, L_2^{'})$. Using the result of Step 1, we multiply both sides of the inequality by $ a^{-d/p} $, obtaining:
\begin{equation*}
\beta_{p,d}(K_2, L_1^{'}) \leq \beta_{p,d}(K_2, L_2).
\end{equation*}
On the other hand, by the definition of the $\beta$-number, we also have $\beta_{p,d}(K_2, L_2) \leq \beta_{p,d}(K_2, L_1^{'})$.
Combining the two inequalities above, we conclude:
\begin{equation*}
 \beta_{p,d}(K_2, L_2)=\beta_{p,d}(K_2, L_1^{'}) = a^{-d/p} \cdot \beta_{p,d}(K_1, L_1).
\end{equation*}
Similar to the proof above, for the $L^p$-version of $\beta$-number in the discrete case, we obtain the same result.
\end{proof}

In summary, for a set $K$ in $\mathbb{R}^n$, when it is scaled with respect to the center of its minimum enclosing ball, the $L^\infty$ version of $\beta$-number remains unchanged, while the $L^p$ version of $\beta$-number is scaled by a constant factor. Furthermore, the optimal plane $L$ undergoes a corresponding scaling. Therefore, for any set $K$ in $\mathbb{R}^n$, no matter how complicated, we can scale it to obtain a set $K'$ such that its minimum enclosing ball is a unit ball. Subsequently, we can analyze $K'$, identify its optimal $d$-dimensional linear subspace, and compute its corresponding $\beta$-number. Finally, by applying inverse scaling, we can determine the optimal $d$-dimensional linear subspace and $\beta$-number for the original set $K$.

\subsection{The relationships between $\beta$-number and its parameters}\label{subsec 2.3}
For the $L^\infty$ version of $\beta$-number, its value depends solely on the set $K$. However, for the $L^p$ version of $\beta$-number, as mentioned in \cref{subsec 2.1}, its value is influenced by multiple factors, including the Hausdorff dimension and the data distribution of set $K$, the dimension $d$ of $L$, and the parameter $p$. In this section, we investigate the relationship between the $L^p$ version of $\beta$-number and the parameter $p$ as well as the dimension $d$.

\subsubsection{The relationship between $\beta$-number value and $p$}
Given a set $K \subset \mathbb{R}^n$ and a dimension $d$, as $p$ varies, the $d$-dimensional optimal subspace $L$ of $K$ typically changes with $p$, except in certain special cases. However, the variation of $L$ with respect to $p$ does not exhibit a regular pattern, posing significant challenges to the analysis of the relationship between the $\beta$-number and $p$. Therefore, in this section, we consider the case where $L$ is fixed, and study the discrete $L^p$ version of $\beta$-number for $K$, i.e.,
\begin{equation*}
    \beta_{p,d}(K, L) = \left( \frac{1}{m \cdot r^d} \sum_{i=1}^{m} \left( \frac{\operatorname{dist}(y_{i}, L)}{r} \right)^p  \right)^{\frac{1}{p}},
\end{equation*}
and investigate its monotonicity with respect to the parameter $p$.
\begin{theorem}\label{The 2.5}
If $K \subset{\mathbb{R}}^n$ is a discrete point set with $m$ elements, and its minimum enclosing ball radius $r$ is greater than or equal to 1, then for any given dimension $d$, $\beta_{p,d}(K, L)$ increases as the parameter $p$ increases.
\end{theorem}
\begin{proof}
    First, decompose $\beta_{p,d}(K, L)$ as follows:
\[
\beta_{p,d}(K, L) = \left( \frac{1}{m \cdot r^d} \sum_{i=1}^m \left( \frac{\operatorname{dist}(y_i, L)}{r} \right)^p \right)^{\frac{1}{p}} = \left( \frac{1}{r^d} \right)^{\frac{1}{p}} \cdot \left( \frac{1}{m} \sum_{i=1}^m \left( \frac{\operatorname{dist}(y_i, L)}{r} \right)^p \right)^{\frac{1}{p}}.
\]
Since $r \geq 1$, it follows directly that $\left( \frac{1}{r^d} \right)^{\frac{1}{p}}$ increases as $p$ increases. Next, we prove that
$\left( \frac{1}{m} \sum_{i=1}^m \left( \frac{\operatorname{dist}(y_i, L)}{r} \right)^p \right)^{\frac{1}{p}}$ also increases with $p$.

When $L$ is fixed, $\frac{\operatorname{dist}(y_i, L)}{r}$ is constant for each $y_i \in K$. Let $a_i = \frac{\operatorname{dist}(y_i, L)}{r}$. The task reduces to proving that $\left( \frac{1}{m} \sum_{i=1}^m a_i^p \right)^{\frac{1}{p}}$ increase monotonically with $p$. Specifically, for any $p_1, p_2$ with $0 < p_1 \leq p_2$ and positive constants $a_1, a_2, \ldots, a_m > 0$, we need to show:
\[
\left( \frac{a_1^{p_1} + \cdots + a_m^{p_1}}{m} \right)^{\frac{1}{p_1}} \leq \left( \frac{a_1^{p_2} + \cdots + a_m^{p_2}}{m} \right)^{\frac{1}{p_2}}.
\]
Let $u_i = a_i^{p_1}$ and $\alpha = \frac{p_2}{p_1} \geq 1$. The inequality becomes:
\[
\left( \frac{u_1 + \cdots + u_m}{m} \right)^\alpha \leq \frac{u_1^\alpha + \cdots + u_m^\alpha}{m}.
\]
Define $g(t) = t^\alpha$. Since $g(t)$ is convex, according to Jensen's inequality, $g\left( \frac{u_1 + \cdots + u_m}{m} \right) \leq \frac{1}{m} \left( g(u_1) + \cdots + g(u_m) \right).$
Thus, the above inequality holds. Consequently, $\left( \frac{1}{m} \sum_{i=1}^m a_i^p \right)^{\frac{1}{p}}$ increases with $p$.

In conclusion, both $\left( \frac{1}{r^d} \right)^{\frac{1}{p}}$ and $\left( \frac{1}{m} \sum_{i=1}^m a_i^p \right)^{\frac{1}{p}}$ increase monotonically with $p$, and since both are non-negative, $\beta_{p,d}(K, L)$ also increases with $p$.
\end{proof}
\begin{remark}
    The monotonicity property described in \cref{The 2.5} holds when the minimum enclosing ball radius $r$ of the discrete point set $K$ is greater than or equal to 1. However, when the radius $r$ does not exceed 1, this monotonicity may not be preserved. In such cases, the behavior of $\beta_{p,d}(K, L)$ as a function of $p$ can vary, and additional analysis may be required to understand the specific conditions under which monotonicity might still hold or fail. This highlights the importance of the radius constraint in ensuring the monotonicity of $\beta_{p,d}(K, L)$ with respect to $p$.
\end{remark}

\subsubsection{The relationship between $\beta$-number value and $d$}
Given all other variables fixed, for a set $K$ with a minimum enclosing ball radius greater than or equal to 1, the relationship between the $L^p$-version of $\beta$-number and the dimension $d$ of the plane $L$ is as follows.
\begin{theorem}
If $K \subset{\mathbb{R}}^n$ is a discrete point set or a bounded set with Hausdorff dimension $k$, and its minimum enclosing ball radius $r$ is greater than or equal to 1, then for any given parameter $p$, the $L^p$ version of $\beta$-number for $K$ decreases as the dimension $d$ increases.
\end{theorem}
\begin{proof}
Taking a bounded set $K$ with Hausdorff dimension $k$ as an example, when the dimension $ d $ is 1, assume that the line $ L_1 $ is the optimal 1-dimensional line such that $\beta_{p,1}(K) := \beta_{p,1}(K, L_1)
$. Let $ L_1' $ be any line that intersects but is not equal to $L_1 $. These two lines $L_1$ and $L_1'$ together form a 2-dimensional plane $L_2'$. For any point $y \in K$, we have the following inequality:
\begin{equation*}
\operatorname{dist}(y, L_2') \leq \operatorname{dist}(y, L_1).
\end{equation*}
Thus, we obtain:
\begin{equation*}
\beta_{p,1}(K) = \left( \frac{1}{V(K) \cdot r^1} \int_K \left( \frac{\operatorname{dist}(y, L_1)}{r} \right)^p dy \right)^{\frac{1}{p}} \geq \left( \frac{1}{V(K) \cdot r^1} \int_K \left( \frac{\operatorname{dist}(y, L_2')}{r} \right)^p dy \right)^{\frac{1}{p}}.
\end{equation*}
However, when $d = 2$, the plane $ L_2' $ may not be the optimal 2-dimensional plane for $ K $. Let $ L_2 $ be the optimal 2-dimensional plane for $ K $. By the definition of $ \beta $-number, we have:
\begin{equation*}
\beta_{p,2}(K) \leq \left( \frac{1}{V(K) \cdot r^2} \int_K \left( \frac{\operatorname{dist}(y, L_2')}{r} \right)^p dy \right)^{\frac{1}{p}} \leq \left( \frac{1}{V(K) \cdot r^1} \int_K \left( \frac{\operatorname{dist}(y, L_2')}{r} \right)^p dy \right)^{\frac{1}{p}}.
\end{equation*}
Thus, $ \beta_{p,1}(K) \geq \beta_{p,2}(K) $. By similar reasoning, as the dimension $ d $ increases, we get:
\begin{equation*}
\beta_{p,1}(K) \geq \beta_{p,2}(K) \geq \cdots \geq \beta_{p,n}(K) = 0.
\end{equation*}
If $K$ is a finite discrete set, then in the discrete case, the proof process of the $L^p$ version of $\beta$-number is similar to that described above for the continuous case. In conclusion, for any given parameter $ p $, as the dimension $ d $ increases, the $ L^p $-version of $ \beta $-number of $ K $ decreases.
\end{proof}
\begin{remark}
When the minimum enclosing ball radius $r$ of $K$ is less than  1, this monotonicity property may not hold. In such a case, the behavior of the $L^p$ version of $\beta$-number with respect to the dimension $d$ requires further analysis to determine specific conditions under which monotonicity might still hold.

\end{remark}

\section{Data Features Described by $\beta$-number}\label{Sect3}
 According to (\ref{Eq 5}) and the theoretical analysis presented in the previous section, it is evident that the $L^\infty$ version of $\beta$-number of set $K$ emphasizes the information of specific boundary points of the set and  does not capture the overall properties of the dataset. In this section, we focus on the characteristics of the data that can be captured by the $L^{p}$ version of $\beta$-number.

\subsection{$p$-th order $d$-dimensional oscillation degree of a set}\label{sub 3.1}
We can intuitively observe from (\ref{Eq 6}) that the $L^{p}$ version of $\beta$-number for a set seeks an optimal lower dimensional linear subspace ($d$-dimensional plane) that minimizes the $p$-power average distances from the set to it. As David and Semmes \cite{David_Semmes_1991} mentioned, the $L^{p}$ version of $\beta$-number measures how well the set $K$ can be approximated by a $d$-plane when $p$ is set to 1. Similarly, the $L^{p}$ version of $\beta$-number identifies an optimal $p$-order $d$-dimensional subspace $L$ for the set $K$ in $\mathbb{R}^n$, and calculates the loss incurred when $K$ is projected to $L$, thereby measuring how well $K$ can be approximated by a $d$-plane in the $p$-order case. Given $p$ and $d$, a smaller $\beta$-number of the set $K$ implies that $K$ becomes flatter, more concentrated around a certain $d$-dimensional plane $L$, experiences a smaller $p$-order loss during projection onto $L$, and preserves more information after the projection. Therefore, we refer to this feature as the "$p$-th order $d$-dimensional oscillation degree" of $K$. For two sets $K_1$ and $K_2$ in $\mathbb{R}^n$, where $dim(L)=d<n$, if $\beta^{p}_{K_1}>\beta^{p}_{K_2}$, it implies that $K_2$ can be better approximated by a $d$-dimensional space compared to $K_1$, experiences less loss when projected into a $d$-dimensional space. Therefore, we say that $K_1$ oscillates more than $K_2$.

\subsection{Comparisons with Principal Component Analysis (PCA)}\label{sub 3.2}
As we all know, Principal Component Analysis (PCA) \cite{Hotelling1933AnalysisOA,jolliffe2002principal,jolliffe2016review} is the most widely used data dimensionality reduction algorithm in unsupervised machine learning. It maps $n$-dimensional features onto $k$-dimensional features, which are new orthogonal features also known as principal components. 

If $K \subset \mathbb{R}^n$ is a discrete point set containing $m$ elements, its $L^p$ version of $\beta$-number minimizes the average $p$-order distance from $K$ to a $d$-dimensional subspace, thereby identifying the optimal d-dimensional subspace $L$. In this context, the process of applying PCA to reduce the dimensionality of $K$ can be understood as seeking a $d$-dimensional subspace that maximizes the product of the variances along the principal component directions. This goal is equivalent to minimizing the mean squared distance from $K$ to this subspace. Therefore, PCA can be formulated as the following optimization problem:
\begin{equation}\label{Eq 18}
    PCA_{d}(K) = \inf_F \frac{1}{m}\sum_{i=1}^{m} \operatorname{dist}(y_i, F)^2,
\end{equation}

where the infimum is taken over all  $d$-dimensional planes $F$ passing through the center of $K$. In fact, the value of $PCA_{d}(K)$ corresponds to the sum of the $n-d$ non-principal eigenvalues of the covariance matrix of $K$.

By comparing (\ref{Eq 6}) and (\ref{Eq 18}), it can be observed that they share a strong structural similarity. Thus, from the perspective of dimensionality reduction, the magnitude of the $\beta$-number can be interpreted as the $p$-order dimensionality reduction loss when the set $K$ in $\mathbb{R}^n$ is projected onto a $d$-dimensional plane $L$. Furthermore, the following conclusions can be drawn.
\begin{theorem}\label{The 3.1}
If $K \subset \mathbb{R}^n$ is a discrete set containing $m$ points, let $r$ denote the radius of the smallest enclosing ball of $K$. For any given dimension $d$, when $p=2$, the optimal $d$-dimensional subspace $L$ corresponding to the $L^2$ version of $\beta$-number is equivalent to the optimal $d$-dimensional subspace of PCA. Furthermore, the following relationship holds:
\begin{equation}
    \beta_{2,d}(K) = \left( \frac{1}{r^{d+2}} \cdot PCA_{d}(K) \right)^{1/2}.
\end{equation}
\end{theorem}
\begin{proof}
    Assume $L_1$ and $L_2$ are the optimal $d$-dimensional planes for the $L^2$ version of the $\beta$-number and PCA, respectively. Let $\bar{y}$ be the centroid of $K$. According to the definitions of $\beta_{2,d}(K)$ and $PCA_{d}(K)$, when $p=2$, the difference between these optimization problems lies in the selection criteria for the $d$-dimensional space. $L_2$ must pass through $\bar{y}$, while $L_1$ does not have this restriction. Thus, if we can prove that when $p=2$, $\bar{y}$ must also lie on $L_1$, then we have:
\[
\beta_{2,d}(K) = \inf_{L}\left(\frac{1}{m \cdot r^{d}} \sum_{i=1}^{m}\left(\frac{\operatorname{dist}(y_{i}, L)}{r}\right)^{2}\right)^{\frac{1}{2}} = \left(\frac{1}{r^{d+2}} \cdot PCA_{d}(K)\right)^{\frac{1}{2}}.
\]
The rest is to prove that when $p=2$, $\bar{y}$ must lie on $L_1$.

Assume $\bar{y}$ is not on $L_1$. Define a plane $L_1'$ parallel to $L_1$ and passing through $\bar{y}$. Let $F$ be the orthogonal complement of $L_1'$ such that $\mathbb{R}^n = F \oplus L_1'$. Here, $F$ is an $(n-d)$-dimensional subspace containing $\bar{y}$. $F$ intersects $L_1'$ at $\bar{y}$, and intersects $L_1$ at point $q$. Project the point set $K$ onto $F$, resulting in the projected set $Q = \{ q_i \}_{i=1}^{m}$. Thus, we have:
\[
\beta_{2,d}(K) = \beta_{2,d}(K,L_1) = \left(\frac{1}{r^{d+2}}\right)^{\frac{1}{2}} \cdot \left(\frac{1}{m} \sum_{i=1}^{m}\left(\operatorname{dist}(q_{i}, q)\right)^{2}\right)^{\frac{1}{2}}
\]
\[
\beta_{2,d}(K,L_1^{'}) = \left(\frac{1}{r^{d+2}}\right)^{\frac{1}{2}} \cdot \left(\frac{1}{m} \sum_{i=1}^{m}\left(\operatorname{dist}(q_{i}, \bar{y})\right)^{2}\right)^{\frac{1}{2}}
\]
According to the definition of the $\beta$-number, we have $\beta_{2,d}(K,L_1) \leq \beta_{2,d}(K,L_1')$. Since $\bar{y}$ is the centroid of both $K$ and $Q$, we have $\sum_{i=1}^{m}\left(\operatorname{dist}(q_{i}, \bar{y})\right)^{2} < \sum_{i=1}^{m}\left(\operatorname{dist}(q_{i}, q)\right)^{2}$. Thus, $\beta_{2,d}(K,L_1') < \beta_{2,d}(K,L_1)$, which is a contradiction. Therefore, when $p=2$, $\bar{y}$ must lie on $L_1$, and the optimal $d$-dimensional plane for the $\beta$-number is equivalent to the optimal $d$-dimensional plane from PCA, with $\beta_{2,d}(K) = \left(\frac{1}{r^{d+2}} \cdot PCA_{d}(K)\right)^{\frac{1}{2}}$.
\end{proof}
Compared to PCA, we assert that the subspace we have identified is more reasonable, as the subspace derived from PCA is constrained to pass through the centroid of the set. However, in cases where $p \neq 2$, the optimal subspace determined by the $L^{p}$ version of $\beta$-number is free from this constraint. Additionally, we present numerical experiments in \cref{Sect5} that demonstrate when $p = 2$, for any set $K$, the $d$-dimensional optimal subspace $L$ identified by $\beta$-number is exactly consistent with the subspace spanned by the eigenvectors corresponding to the first $d$ principal components in PCA.
From the perspective of dimensionality reduction, $\beta_{p,d}(K)$ represents the $p$-th order dimensionality reduction loss when projecting the set $K$ onto a $d$-dimensional space. Therefore, we consider the $\beta$-number to be a higher-order generalization of PCA.
\begin{remark}
    $\beta_{p,d}(K)$ is defined as the "$p$-th order $d$-dimensional oscillation degree" of the set $K$, representing the minimum $p$-th order dimensionality reduction loss incurred when projecting the set $K$ onto a $d$-dimensional plane.
\end{remark}

\section{Numerical algorithms}\label{Sect4}
Let $K \subset{\mathbb{R}^n}$ be a discrete point set with $m$ elements or a bounded set with Hausdorff dimension $k$. To approximate two forms of $\beta$-number for $K$, we employ the following three-step approach. First, we discretize the integral form of the $\beta$-number in the continuous case, transforming it into a discrete form for subsequent computation. Then, we determine the center and radius of the minimum enclosing ball of the set $K$, and normalize the set $K$ to the $n$-dimensional unit ball, facilitating subsequent calculations. Second, we reformulate the problem of finding the optimal $d$-dimensional plane $L$ in $\beta$-number as an optimization problem, thereby determining the optimal solution. Finally, if $K$ is a bounded set with Hausdorff dimension $k$, we partition the region of $K$ to convert it into a weighted finite discrete point set, and then approximate $\beta$-number using the same method as for a discrete point set with $m$ elements. The detailed implementation can be found in \cref{Appendix A}.

In summary, for a discrete point set in $\mathbb{R}^n$, our \cref{alg1:discrete beta_number} proceeds as follows:
\begin{algorithm}[H]
\caption{Computation of two forms of $\beta$-number for a discrete set $K$}
\label{alg1:discrete beta_number}
\begin{algorithmic}
    \STATE \textbf{Input:} $K=\{y_i\}_{i=1}^{m}$, $d$, $p$.
    
    \STATE \textbf{Output:}  $\beta_{\infty}(K)$,
    $\beta_{p,d}(K)$
    
    \STATE \textbf{Initialization:} Compute $B(o,r)$, let $K' = \frac{1}{r}(K - o)$.
    
    \STATE \textbf{Optimization Problem:} Solve the optimization problem as in \cref{subsec A.2}:
    \[
    \beta_{\infty}(K')
    =\min_{A_1, b_1} \left( \max_{y_i \in K'} \left( \frac{\| A_1^T (A_1 A_1^T)^{-1} (b_1 - A_1 y_i) \|_2}{r} \right) \right)
    \]
    \[\beta_{p,d}(K') 
    =\min_{A_2, b_2} \left( \frac{1}{m \cdot r^{d}} \sum_{i=1}^{m} \left( \frac{\| A_2^T (A_2 A_2^T)^{-1} (b_2 - A_2 y_i) \|_2}{r}  \right)^{p} \right)^{\frac{1}{p}}. 
    \]
    obtain $L_{\infty}^{'}: A_1y = b_1$,
    $L_{p,d}^{'}: A_2y = b_2$,
    $\beta_{\infty}(K')$,
    $\beta_{p,d}(K')$.
    \STATE \textbf{Inversely scale:} $L_{\infty} = rL_{\infty}^{'} + o$,
    $L_{p,d} = rL_{p,d}^{'} + o$,
    $\beta_{\infty}(K) = \beta_{\infty}(K')$,
    $\beta_{p,d}(K) = r^{-d/p} \cdot \beta_{p,d}(K')$.
\end{algorithmic}
\end{algorithm}

For a bounded set with Hausdorff dimension $k$ in $\mathbb{R}^n$, our \cref{alg2:discrete beta_number} proceeds as follows:
\begin{algorithm}[H]
\caption{Computation of two forms of $\beta$-number for a bounded set $K$ with Hausdorff dimension $k$}
\label{alg2:discrete beta_number}
\begin{algorithmic}
    \STATE \textbf{Input:} $K$, $d$, $p$.
    
    \STATE \textbf{Output:}  $\beta_{\infty}(K)$,
    $\beta_{p,d}(K)$
    
    \STATE \textbf{Initialization:} Compute $B(o,r)$, let $K' = \frac{1}{r}(K - o)$.

    \STATE \textbf{Domain Partition:}
    Partition $K'$ as $K'=\bigcup_{i=1}^{n} \Delta_{i}$, along with a centroid set $\tilde K=\{y_i\}_{i=1}^{n}$ and a weight set $W=\{w_i|w_i=|\Delta_{i}|/V(K')\}_{i=1}^{n}$.
    
    \STATE \textbf{Optimization Problem:} Solve the optimization problem as in \cref{subsec A.2}:
    \[
    \beta_{\infty}(\tilde K) = \min_{A_1, b_1} \left( \max_{y_i \in \tilde K} \left( \frac{\| A_1^T (A_1 A_1^T)^{-1} (b_1 - A_1 y_i) \|_2}{r} \right) \right)
    \]
    \[\beta_{p,d}(\tilde K) = \min_{A_2, b_2} \left( \frac{1}{r^{d}} \sum_{i=1}^{n} \left( \frac{w_{i} \cdot \| A_2^T (A_2 A_2^T)^{-1} (b_2 - A_2 y_i) \|_2}{r} \right)^{p} \right)^{\frac{1}{p}}. 
    \]
    obtain $\tilde L_{\infty}: A_1y = b_1$,
    $\tilde L_{p,d}^{'}: A_2y = b_2$,
    $\beta_{\infty}(\tilde K)$,
    $\beta_{p,d}(\tilde K)$.
    \STATE \textbf{Inversely scale:} $L_{\infty} = r\tilde L_{\infty} + o$,
    $L_{p,d} = r\tilde L_{p,d} + o$,
    $\beta_{\infty}(K) = \beta_{\infty}(\tilde K)$,
    $\beta_{p,d}(K) = r^{-d/p} \cdot \beta_{p,d}(\tilde K)$.
\end{algorithmic}
\end{algorithm}

\section{Numerical experiments}\label{Sect5}
In this section, we first utilize the numerical methods proposed in \cref{Sect4} to compute $\beta$-number of a given set $K$ within low-dimensional spaces ($n \leq 3$) and to visualize the corresponding line or plane $L$. Subsequently, we compare and visualize the optimal subspaces identified through $\beta$-number with those derived from Principal Component Analysis (PCA), highlighting the advantages of our approach. Our numerical experiments provide empirical support for the preceding theoretical analysis. It is noteworthy that all sets $K$ used in our experiments have been normalized to the unit sphere.

\subsection{Low dimensional cases ($n\leq3$)}
For a given set $K$ in low-dimensional spaces, we analyze the computation of its two forms of 
$\beta$-number through classification. If $K$ is a bounded set with Hausdorff dimension $k$, we need to perform region partitioning based on $k$, transforming it into a discrete point set with varying weights for approximate calculations. Conversely, if $K$ is a finite discrete point set, the weights of all points in $K$ are identical, allowing for straightforward approximate calculations. We present our results on both toy examples and a publicly available image dataset in \cref{Appendix B}.

\subsection{Comparisons with PCA}
Firstly, for a set $K$ with Hausdorff dimension $k$, when calculating its $\beta$-number, we perform a region subdivision and use the volume proportion of each subdivided unit as the weight for the corresponding centroid points, thereby obtaining a set of centroid points. Consequently, the results of \cref{The 3.1} are also applicable to the relationship between the $L^2$ version of the $\beta$-number for the set $K$ and Weighted Principal Component Analysis (WPCA) \cite{cazes1988weighted,jolliffe2016review}. We will subsequently validate our viewpoint through experimental evidence.

For $K$ in $\mathbb{R}^2$ and $\mathbb{R}^3$, which can be a discrete point set with $m$ elements or a bounded set with Hausdorff dimension $k$, the relationship between the $L^2$ version of $\beta$-number and PCA is illustrated in the following \cref{fig12}:
\begin{figure}[htbp]
    \centering
    \begin{tabular}{ccc}
        \multicolumn{1}{p{0.25\textwidth}}{\centering \textbf{One-dimensional \\ subspace in $\mathbb{R}^2$}} & 
        \multicolumn{1}{p{0.25\textwidth}}{\centering \textbf{One-dimensional \\ subspace in $\mathbb{R}^3$}} & 
        \multicolumn{1}{p{0.25\textwidth}}{\centering \textbf{Two-dimensional \\ subspace in $\mathbb{R}^3$}} \\
        \subcaptionbox{
        \parbox{0.2\textwidth}{\centering
            $\beta_{2,1}(K)$=0.4993\\
            $PCA_1(K)$=0.2493
        }}{\includegraphics[width=0.25\textwidth]{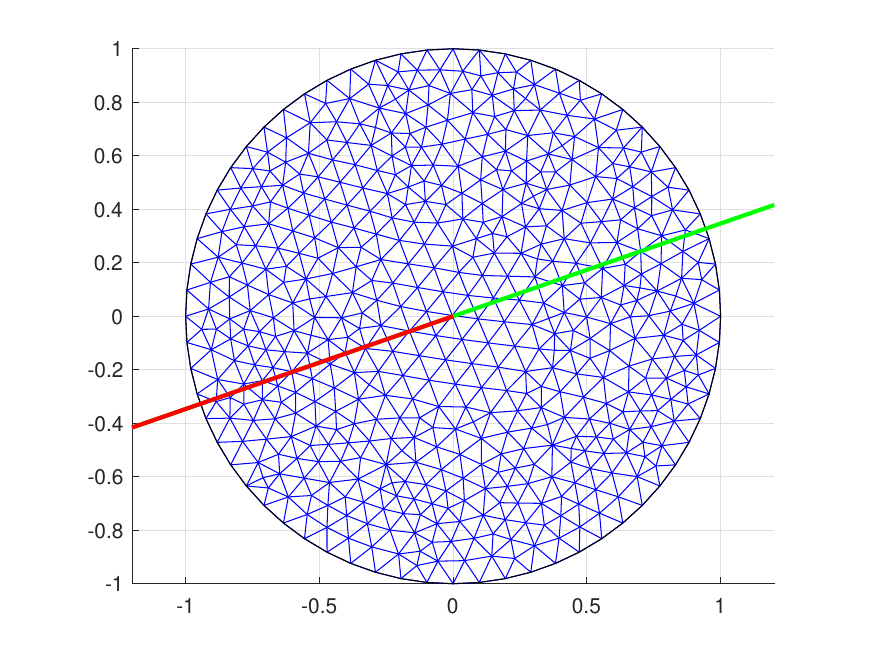}} &
        \subcaptionbox{
        \parbox{0.2\textwidth}{\centering
            $\beta_{2,1}(K)$=0.6301\\
            $PCA_1(K)$=0.3971
        }}{\includegraphics[width=0.25\textwidth]{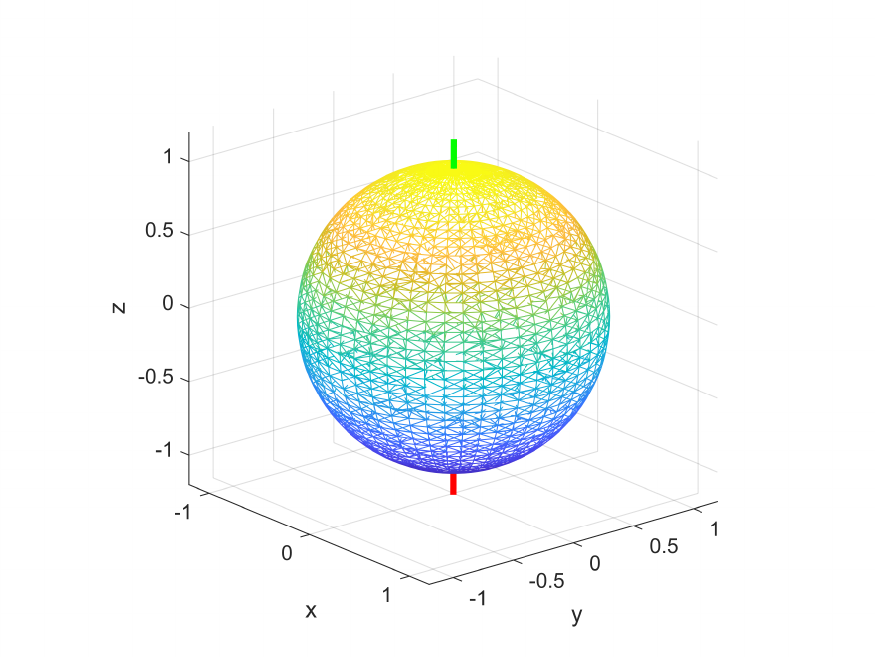}} &
        \subcaptionbox{
        \parbox{0.2\textwidth}{\centering
            $\beta_{2,2}(K)$=0.4456\\
            $PCA_2(K)$=0.1985
        }}{\includegraphics[width=0.25\textwidth]{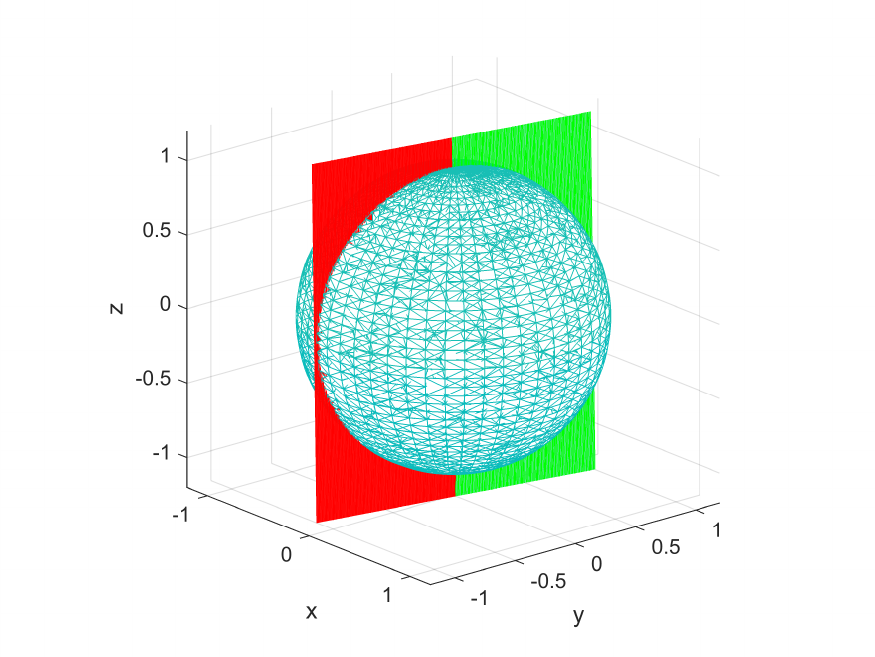}} \\
        \subcaptionbox{
        \parbox{0.2\textwidth}{\centering
            $\beta_{2,1}(K)$=0.2775\\
            $PCA_1(K)$=0.0770
        }}{\includegraphics[width=0.25\textwidth]{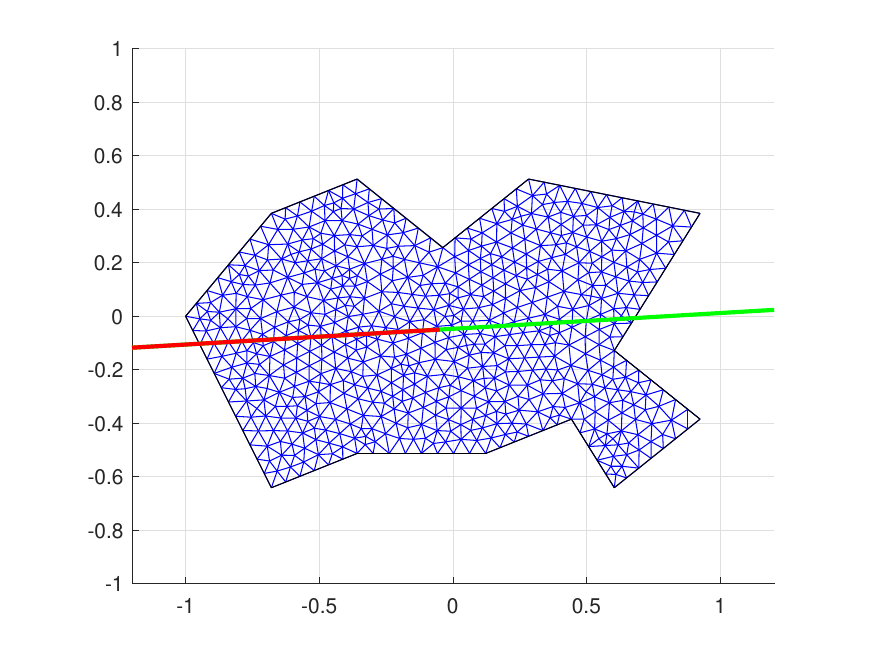}} &
        \subcaptionbox{\parbox{0.2\textwidth}{\centering
            $\beta_{2,1}(K)$=0.3860\\
            $PCA_1(K)$=0.1490
        }}{\includegraphics[width=0.25\textwidth]{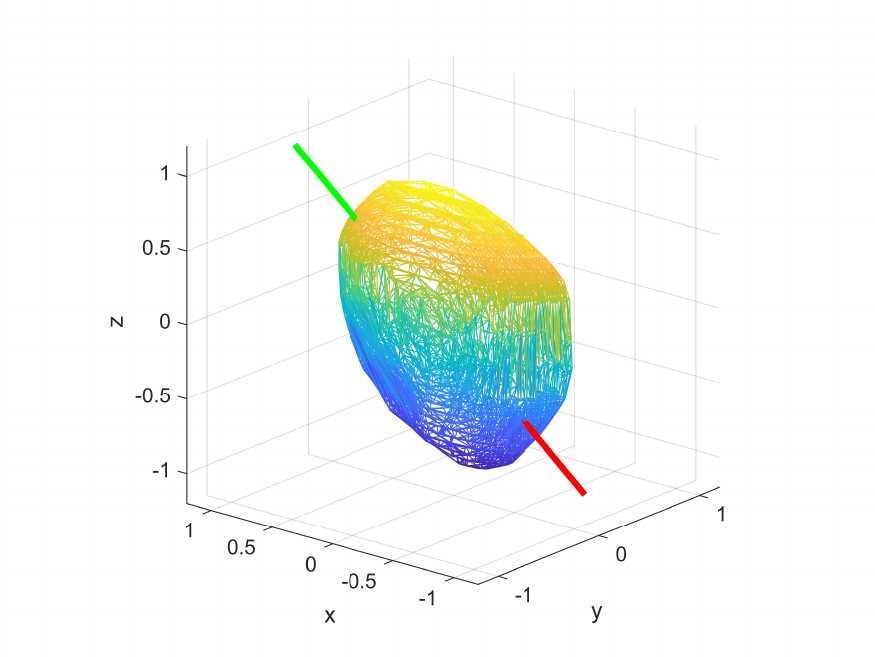}} &
        \subcaptionbox{
        \parbox{0.2\textwidth}{\centering
            $\beta_{2,2}(K)$=0.1052\\
            $PCA_2(K)$=0.0111
        }}{\includegraphics[width=0.25\textwidth]{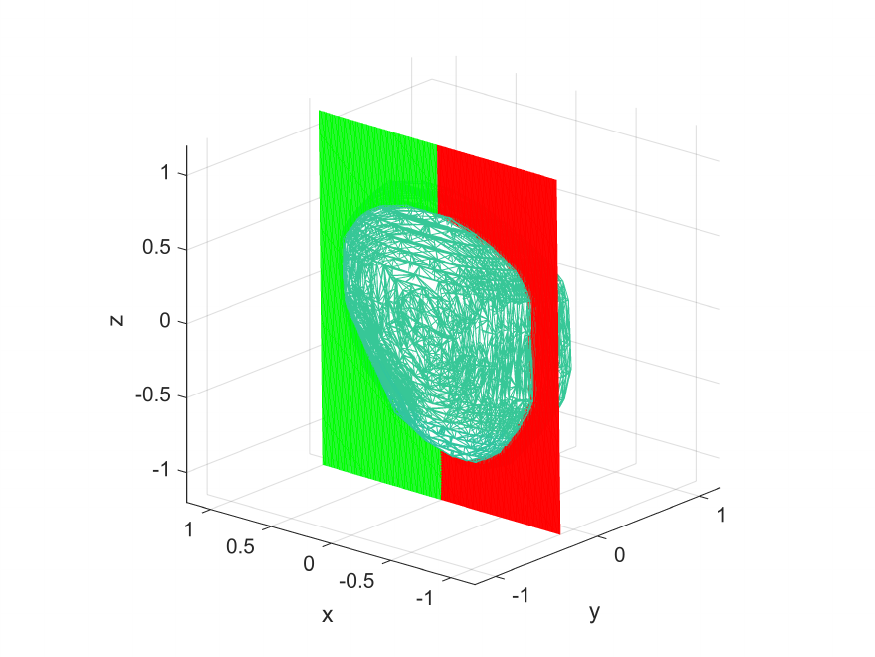}} \\
        \subcaptionbox{
        \parbox{0.2\textwidth}{\centering
            $\beta_{2,1}(K)$=0.4138\\
            $PCA_1(K)$=0.1712
        }}{\includegraphics[width=0.25\textwidth]{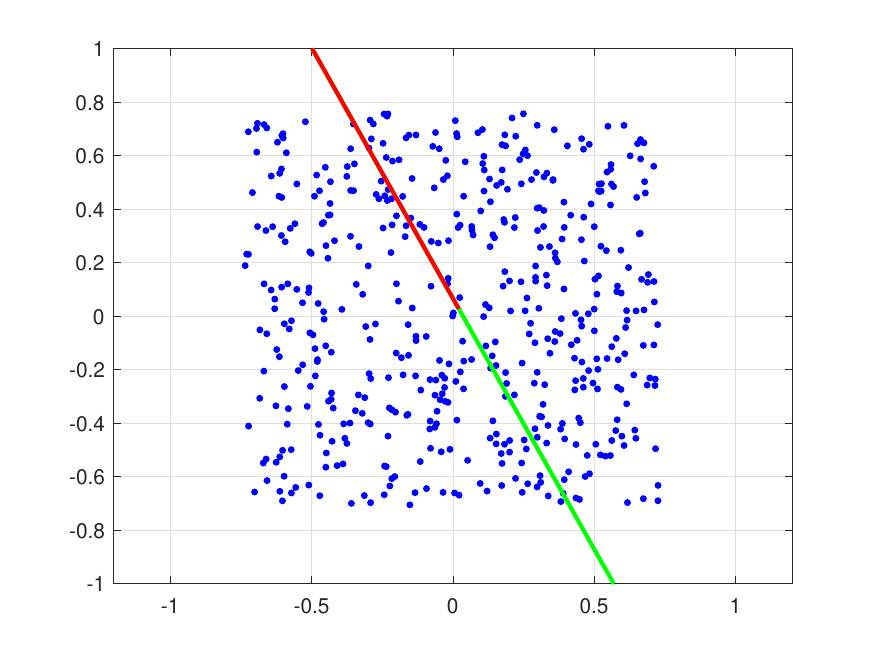}} &
        \subcaptionbox{
        \parbox{0.2\textwidth}{\centering
            $\beta_{2,1}(K)$=0.5172\\
            $PCA_1(K)$=0.2675
        }}{\includegraphics[width=0.25\textwidth]{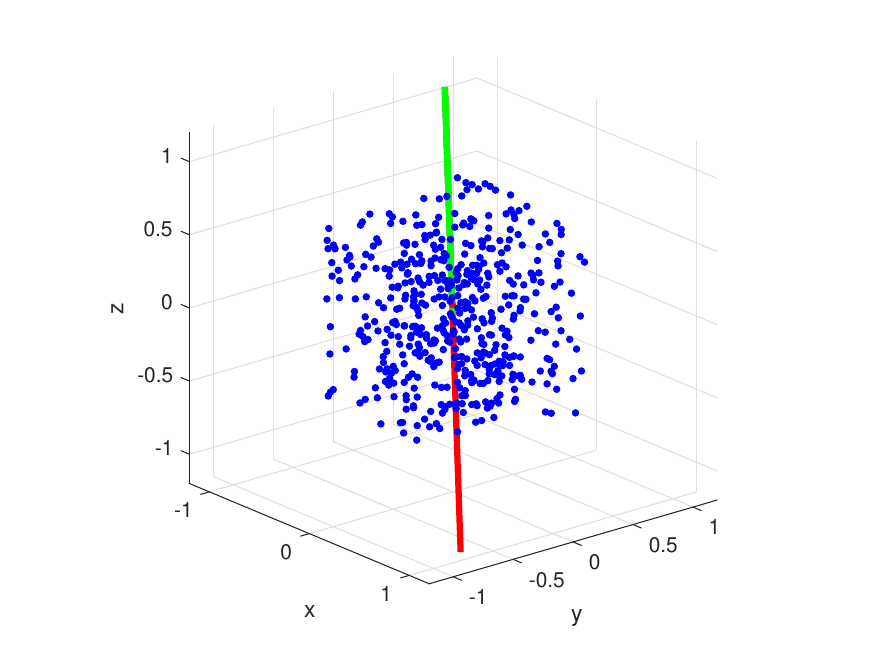}} &
        \subcaptionbox{
        \parbox{0.2\textwidth}{\centering
            $\beta_{2,2}(K)$=0.3598\\
            $PCA_2(K)$=0.1294
        }}{\includegraphics[width=0.25\textwidth]{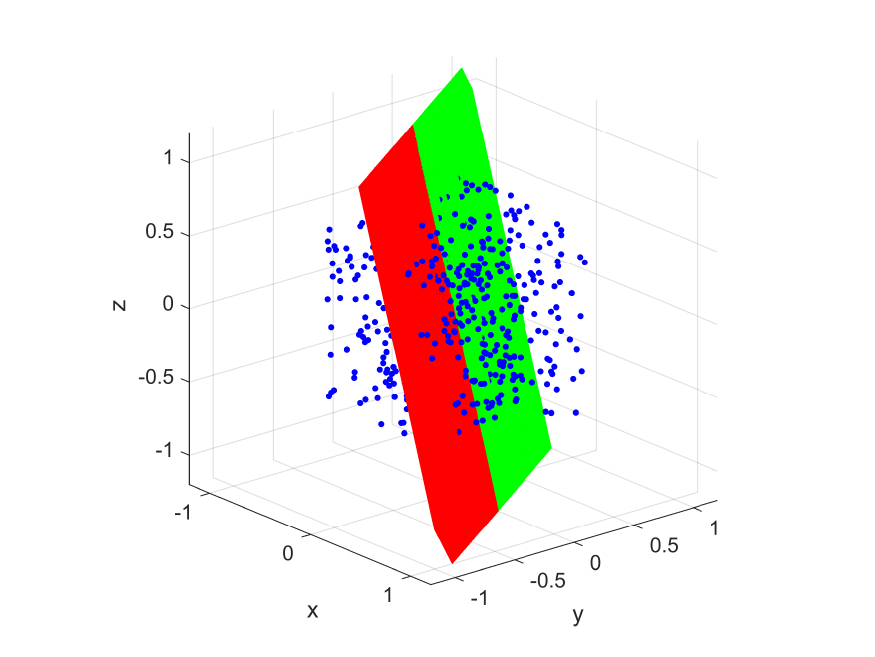}} \\
    \end{tabular}
    \caption{Comparison of subspaces between PCA and $\beta$-number. In each figure, the dimensionality-reduced subspace of $\beta$-number is represented in red, while that of PCA is represented in green, with the centroid of $K$ located at the intersection of these two subspaces. The corresponding $\beta$-number value and $PCA_d(K)$ value are displayed at the bottom of each figure.
}
    \label{fig12}
\end{figure}

From \cref{fig12}, it can be observed that whether $K$ is a finite discrete point set or a bounded set with Hausdorff dimension $k$, when $r=1$ and $p=2$, the optimal $d$-dimensional subspace in the $L^2$ version is identical to the dimensionality-reduced subspace formed by the eigenvectors corresponding to the top $d$ largest eigenvalues in PCA. Furthermore, it holds that $\beta_{2,d}(K) = \left( PCA_{d}(K) \right)^{1/2}$, which is consistent with the conclusion of \cref{The 3.1}.

In summary, $\beta$-number can be regarded as a higher-order extension of PCA. It minimizes the $p$-th order distance loss when reducing the set $K$ to a $d$-dimensional subspace, rather than solely relying on variance to assess the retention or loss of information. Furthermore, by adjusting the value of $p$, we can obtain various $p$-th order $d$-dimensional subspaces. Unlike the subspaces identified by PCA, our $p$-th order $d$-dimensional subspaces are not constrained to contain the centroid of the set when $p \neq 2$. Thus, we observe that the subspaces obtained through the $\beta$-number are more reasonable.

It is important to note that PCA enhances computational efficiency by transforming variance calculations into matrix operations, thereby significantly reducing runtime. In contrast, the computation of $\beta$-number involves $p$-th order distances and necessitates optimization algorithms to identify optimal planes, leading to longer runtimes compared to PCA. However, for large datasets, WPCA requires substantial memory for matrix storage and computation due to the varying weights of centroid points. Conversely, our $\beta$-number approach demands relatively less memory.

\section{Applications to Data Science}\label{Sect6}
In this section, we apply $\beta$-number to data science. As previously mentioned, the $L^p$ version of $\beta$-number for a set seeks an optimal $d$-dimensional plane that minimizes the $p$-power average distances from the set to the plane. Thus, using $\beta$-number, we can not only identify the optimal $p$-th order $d$-dimensional plane for dimensionality reduction of a data point set, but also compute the $p$-th order $d$-dimensional oscillation degree for each sample for data classification. In \cref{subsec 6.1}, we perform dimensionality reduction on a three-dimensional data point set and compare it with PCA. In \cref{subsec 6.2}, we calculate $\beta$-number for images from different categories in a publicly available image datasets for image classification.
 
\subsection{Data dimensionality reduction based on $\beta$-number}\label{subsec 6.1}
In this section, we apply $\beta$-number to data dimensionality reduction tasks. In order to provide a more intuitive demonstration of the dimensionality reduction effect of our $\beta$-number and compare it with PCA, we construct a three-dimensional classification dataset $K$. By calculating $\beta$-number of $K$, we identify the optimal $p$-th order two-dimensional reduction plane. This plane is then compared with the optimal two-dimensional reduction plane obtained via PCA, along with their respective classification performances.

\begin{figure}[htbp]
    \centering
    \begin{tabular}{ccc}
         \\
        \subcaptionbox{Dataset $K$\label{fig13a}}{\includegraphics[width=0.3\textwidth]{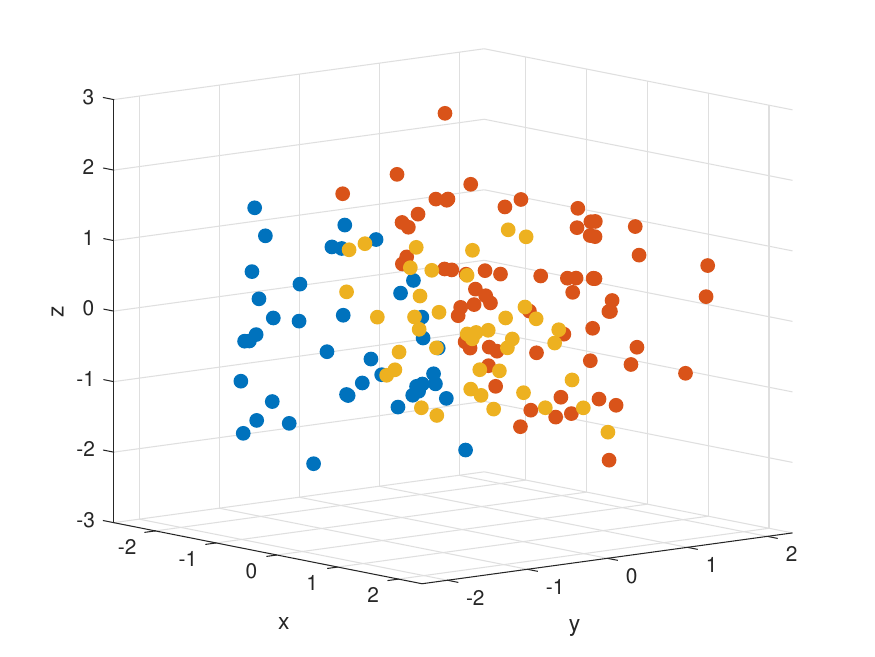}} &
        
        \subcaptionbox{Two dimensionality reduction planes}{\includegraphics[width=0.3\textwidth]{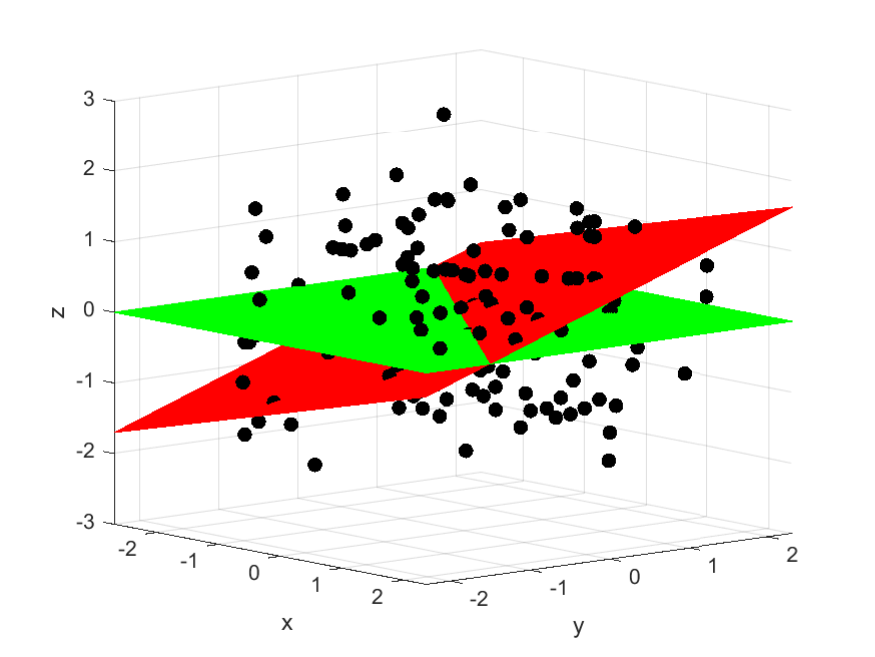}} &     \subcaptionbox{Projected points on $L_1$}{\includegraphics[width=0.3\textwidth]{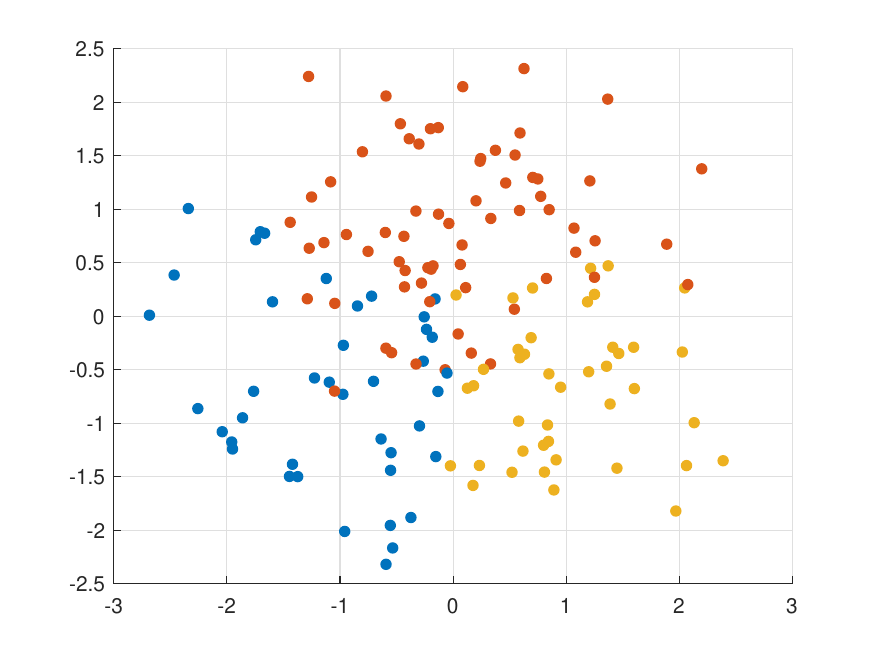}} \\
        \subcaptionbox{Projected points on $L_2$}{\includegraphics[width=0.3\textwidth]{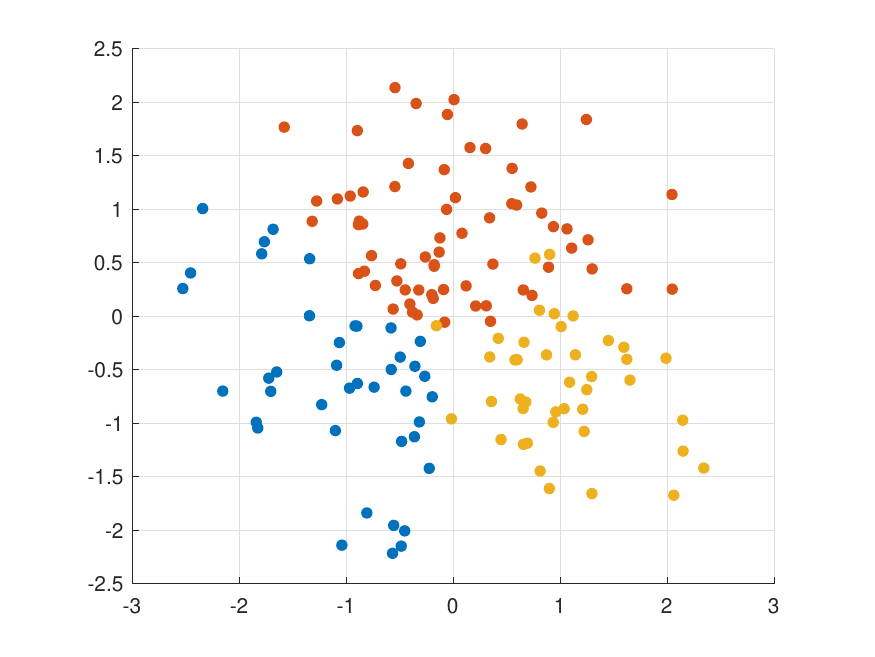}} &
        \subcaptionbox{Decision boundary of $K_1$}{\includegraphics[width=0.3\textwidth]{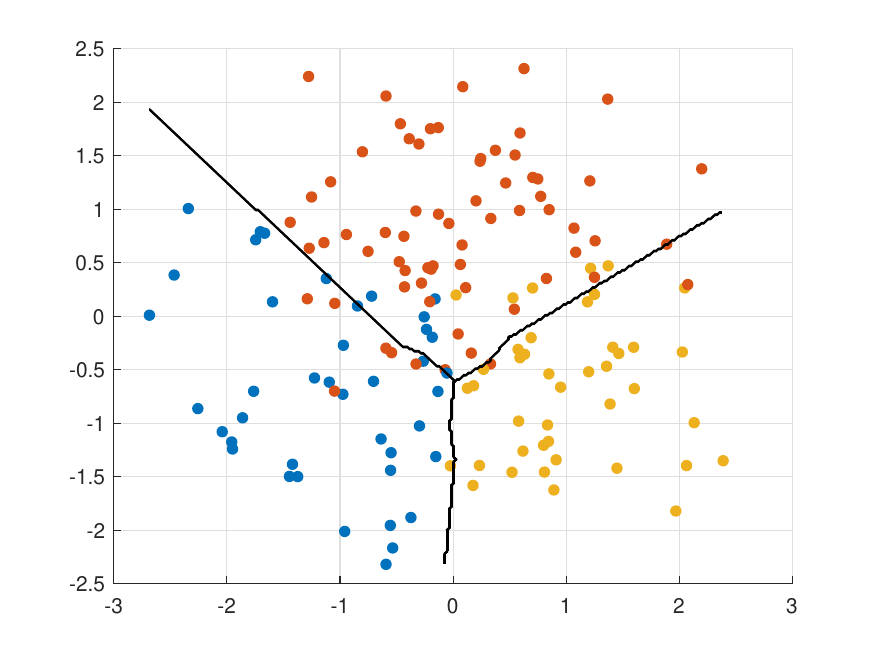}} &
        \subcaptionbox{Decision boundary of $K_2$}{\includegraphics[width=0.3\textwidth]{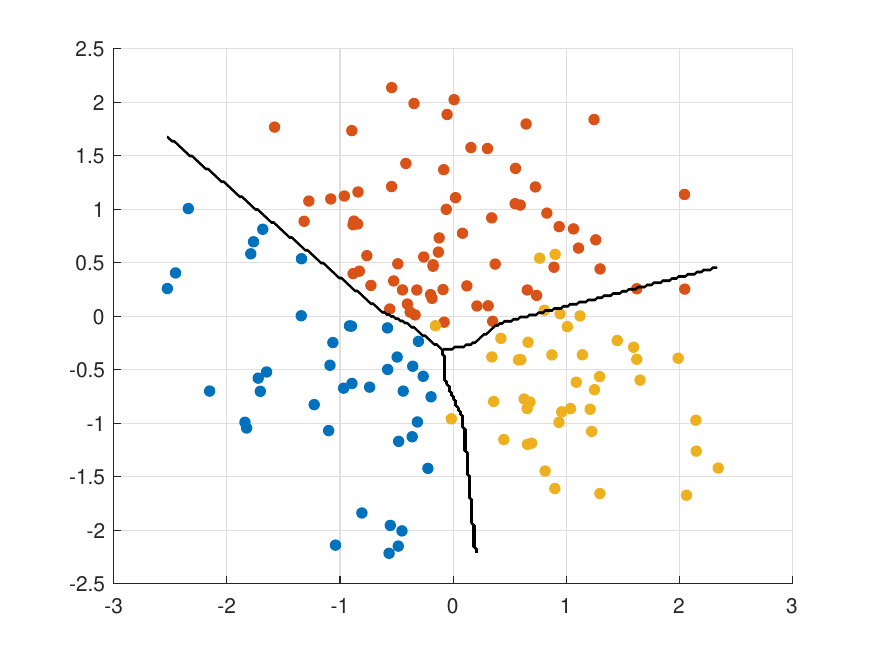}} \\
    \end{tabular}
    \caption{(\textbf{a}) Illustration of the 3D point set $K$, with points of different categories represented in different colors. (\textbf{b}) Illustration of two dimensionality reduction planes, where the green represents the PCA reduced plane $L_1$, and the red represents the $\beta$-number reduced plane $L_2$. (\textbf{c}) Illustration of the projected point set $K_1$ on $L_1$. (\textbf{d}) Illustration of the projected point set $K_2$ on $L_2$. (\textbf{e}) Illustration of the SVM classification decision boundary for $K_1$. (\textbf{f}) Illustration of the SVM classification decision boundary for $K_2$.
}
    \label{fig13}
\end{figure}

Initially, the dataset $K$ is composed of three classes of points, as shown in \cref{fig13a}, with each class represented by a different color. Subsequently, we set $p=1$ and employ both PCA and $\beta$-number to reduce $K$ to two dimensions, resulting in two reduction planes, $L_1$ and $L_2$. To demonstrate that the 1-order 2-dimensional reduction plane derived from our $\beta$-number is more advantageous for classifying the dataset $K$ than the optimal two-dimensional reduction plane obtained from PCA, we employ Support Vector Machine (SVM)\cite{SVM,scholkopf2002SVM} to classify the projection sets $K_1$ and $K_2$ on $L_1$ and $L_2$, respectively, outputting classification accuracy and plotting the SVM decision boundaries. The results are illustrated in \cref{fig13}.

From \cref{fig13}, we can observe that $K_2$ is indeed more separable than $K_1$. The SVM classification results show an accuracy of 86.00\% on $K_1$ and 95.33\% on $K_2$. This example demonstrates that, for certain datasets, $\beta$-number-based dimensionality reduction can achieve better results than PCA. 

\subsection{Image classification based on $\beta$-number}\label{subsec 6.2}
In this section, we apply $\beta$-number to image classification tasks. For a given image classification dataset, we explore the potential of $\beta$-number as a novel metric by calculating it for images from different categories.

\begin{figure}[htbp]
    \centering
    \begin{tabular}{ccc}
         \\
        \subcaptionbox{}{\includegraphics[width=0.4\textwidth]{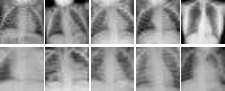}} &

        \subcaptionbox{\label{fig14b}}{\includegraphics[width=0.25\textwidth]{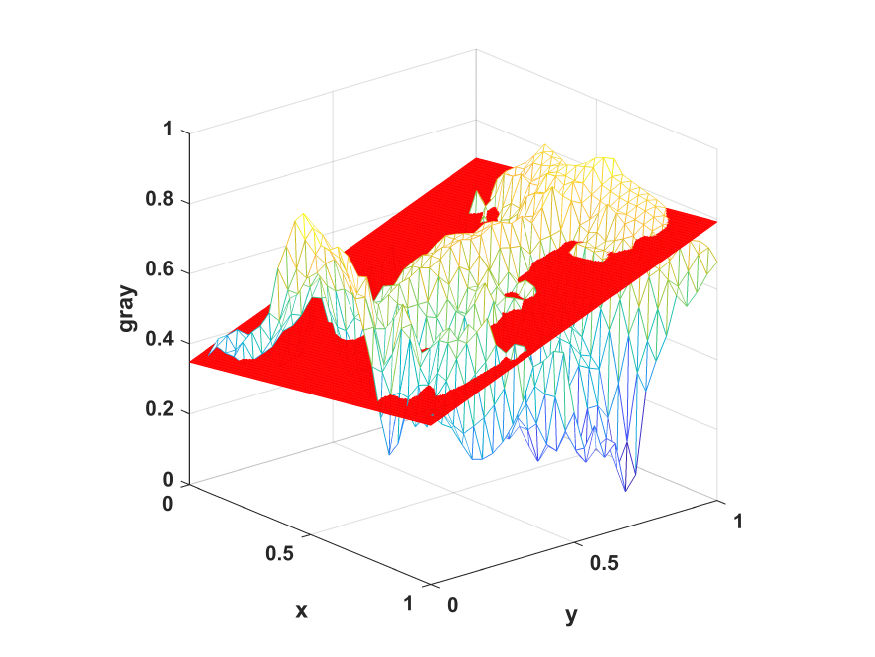}} &
        
        \subcaptionbox{\label{fig14c}}{\includegraphics[width=0.25\textwidth]{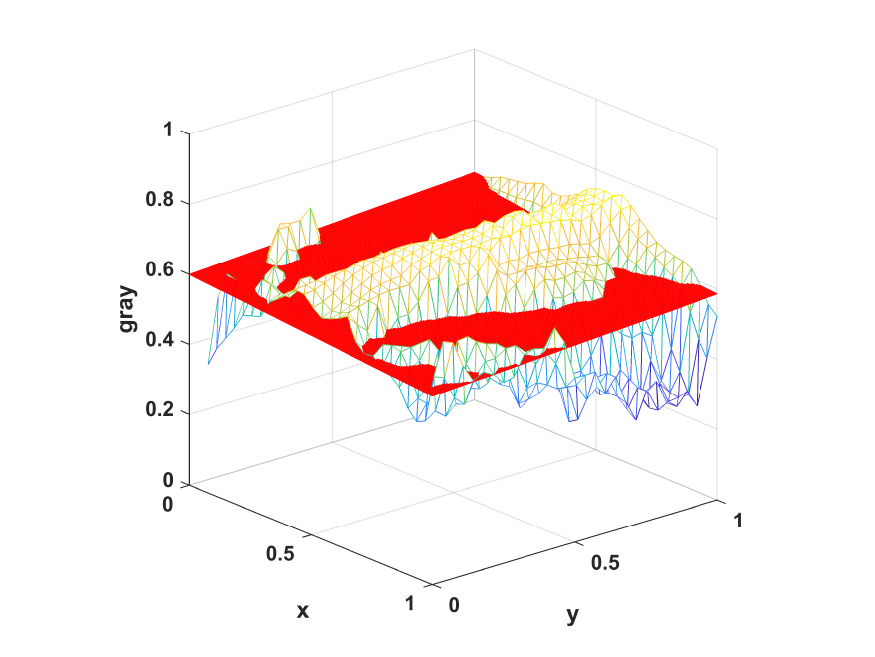}}   
    \end{tabular}
    \caption{(\textbf{a}) Schematic of the PneumoniaMNIST dataset. The first row presents normal images, with their $\beta_{1,2}(K)$ values from left to right being: 0.3002, 0.3200, 0.3402, 0.3608, and 0.3827; the second row presents pneumonia images, with their $\beta_{1,2}(K)$ values from left to right being: 0.1784, 0.2024, 0.2211, 0.2408, and 0.2603. (\textbf{b}) The fitted plane of  a typical normal image's grayscale surface when $p=1$. The corresponding value of $\beta_{1,2}(K)$ is 0.3608. (\textbf{c}) The fitted plane of a typical pneumonia image's grayscale surface when $p=1$. The corresponding value of $\beta_{1,2}(K)$ is 0.2408.
}
    \label{fig14}
\end{figure}

PneumoniaMNIST, a subset of the MedMNIST dataset \cite{medmnistv2}, is specifically designed for the binary classification task of distinguishing between pneumonia and normal chest X-ray images. This dataset comprises 5,856 pediatric chest X-ray images, each with a resolution of $28 \times 28$ pixels. In this binary classification task, due to the presence of white regions in the lung areas of pneumonia-affected patients that closely resemble the background, the overall grayscale variation in such images tends to be lower. To address this, we utilize $\beta$-number to capture the grayscale oscillation within each image. 
Specifically, for each image, we begin by extracting the planar coordinates and grayscale values of all pixels. To address the inconsistency between the dimensions of the planar coordinates and grayscale values, both are independently normalized to the range of $[0,1]$. This process generates a three-dimensional grayscale point set, denoted as $K$, which represents the grayscale surface. Next, $\beta$-number is applied to fit an optimal two-dimensional plane and calculate the grayscale oscillation of the point set $K$, resulting in the values of $\beta_{1,2}(K)$, as illustrated in \cref{fig14}.

\begin{figure}[ht]
    \centering
    \begin{subfigure}{0.4\textwidth}
        \centering
        \includegraphics[width=1\textwidth]{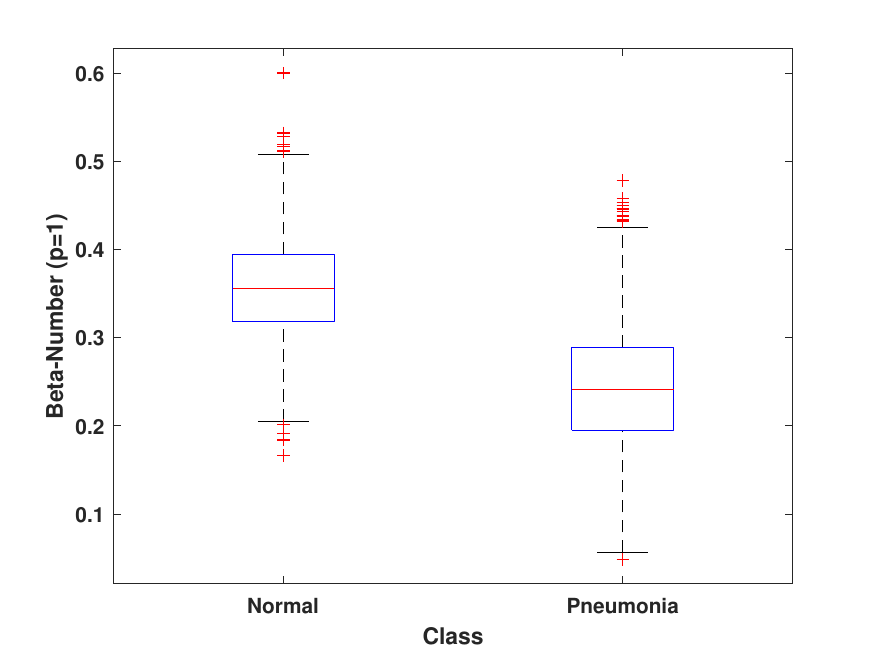}
        \caption{Boxplot of $\beta$-number ($p$=1) by class}
    \end{subfigure}%
    \begin{subfigure}{0.4\textwidth}
        \centering
        \includegraphics[width=1\textwidth]{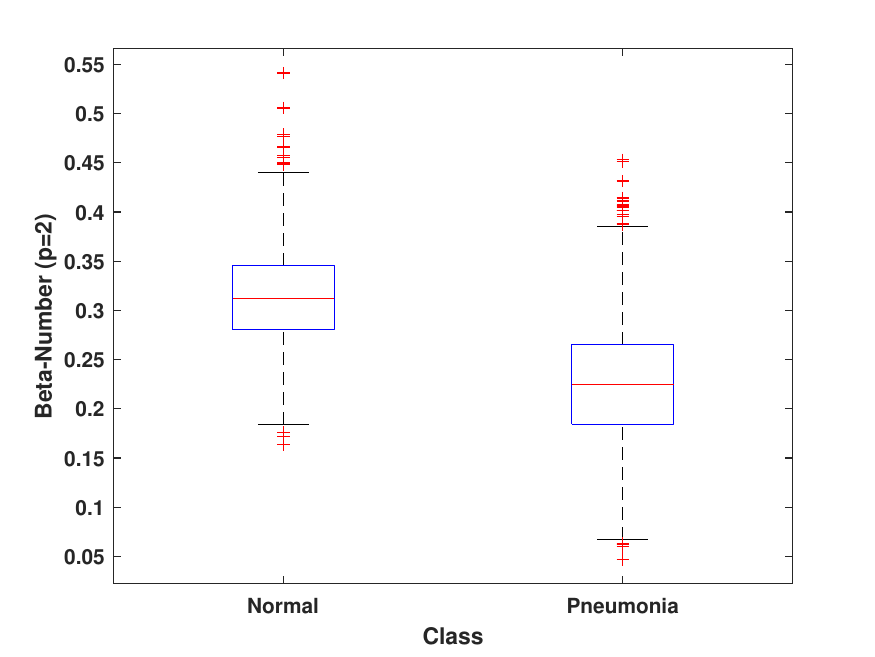}
        \caption{Boxplot of $\beta$-number ($p$=2) by class}
    \end{subfigure}
    \caption{Boxplot of $\beta$-number derived from  5,856 PneumoniaMNIST samples, plotted against the image category. The x-axis represents the class (pneumonia and normal), and the y-axis represents the corresponding $\beta$-number (for both $p=1$ and $p=2$ cases).}
    \label{fig15}
\end{figure}

As shown in \cref{fig14b} and \cref{fig14c}, compared to a pneumonia image, the grayscale surface of a  normal image exhibits greater oscillations, deviating further from its optimal two-dimensional plane in terms of the $p$-th order average distance, which results in a larger $\beta$-number. In the PneumoniaMNIST dataset, there are $4,273$ pneumonia images and $1,583$ normal images. The grayscale oscillation degrees, $\beta_{1,2}(K)$ and $\beta_{2,2}(K)$, are calculated for each image, and the overall results are presented in \cref{fig15}.

As illustrated in \cref{fig15}, the proposed $\beta$-number metric effectively captures the grayscale oscillation characteristics of 2D images, demonstrating significant class separability in this binary classification task, regardless of whether $p=1$ or $p=2$. In the classification of the image dataset based on $\beta$-number, all images are randomly divided into training and testing sets at an 80:20 ratio. The optimal threshold for classification accuracy is determined on the training set, and this threshold is subsequently validated on the testing set. For $p=1$, the threshold that maximizes classification accuracy on the training set is 0.3344, with images exceeding this value classified as normal and those below it identified as pneumonia. This threshold results in a classification accuracy of 83.60\% on the testing set. Similarly, for $p=2$, a threshold of 0.2984 is applied, yielding a classification accuracy of 81.13\% on the testing set. In summary, $\beta$-number metric alone achieves over 80\% classification accuracy for this task, underscoring its substantial potential for practical applications.

Additionally, we introduce $\beta$-number as a novel feature, which is subsequently integrated with the existing features extracted via machine learning methods to collectively guide the classification process. Specifically, we first apply the Histogram of Oriented Gradients (HOG) technique \cite{HOG} for image feature extraction, resulting in a 1296-dimensional feature vector for each image. We then combine the calculated $\beta$-number, treated as an additional one-dimensional feature, with the HOG features. A variety of classifiers are employed for image classification, and the results are presented in \cref{tab1}:

\begin{table}[h!]
  \caption{Classification Accuracy with Different Classifiers and Feature Sets}
  \label{tab1}
  \centering
  \begin{tabular}{lcc}
    \toprule
    \multicolumn{1}{c}{} & \multicolumn{2}{c}{Accuracy} \\
    \cmidrule(r){2-3}
    Classifier          & HOG  & HOG + $\beta$-number \\
    \midrule
    SVM\cite{SVM}                &0.93515      &0.94198                       \\
    Random Forest\cite{RandomForest}      &0.90529      &0.91724                       \\
    Logistic Regression\cite{LogisticRegression} &0.94198      &0.94625                       \\
    Decision Tree\cite{DecisionTree}      &0.83788      &0.86007                       \\
    Naive Bayes\cite{Naive_Bayes}          &0.87457      &0.87799                       \\
    \bottomrule
  \end{tabular}
\end{table}

The results presented in the table indicate that the inclusion of $\beta$-number as an additional feature, alongside the existing feature set, leads to a significant improvement in classification performance. To further illustrate the importance of $\beta$-number, we conduct SHapley Additive exPlanations (SHAP) feature importance analysis \cite{Lundberg_shap_2017} on the 1296 HOG features and $\beta$-number. Notably, $\beta$-number achieves the highest SHAP contribution score 0.034057 among all features. This score is substantially higher than the contribution values of the next top-four features. For instance, the contribution value of HOG-feature$_{1130}$ is 0.016845, that of HOG-feature$_{932}$ is 0.013232, that of HOG-feature$_{1099}$ is 0.012599 and that of HOG-feature$_{959}$ is 0.012061. Collectively, these results confirm that $\beta$-number not only functions as an effective standalone geometric feature but also complements existing HOG features to enhance downstream classification performance. Moreover, its dominant position in SHAP-based feature importance further validates its advantages over individual traditional features, underscoring its critical role in robust data feature representation.

\section{Conclusion}\label{Sect7}

This paper first introduces the concept of the traditional $\beta$-number in geometric analysis and presents our modified version. Then we provide a theoretical analysis of its upper and lower bounds, and explore its relationship with independent variables. By identifying the $d$-dimensional optimal plane, $\beta$-number quantitatively evaluates the minimum $p$-th order loss incurred when projecting set $K$ in $\mathbb{R}^n$ onto a $d$-dimensional plane. A set with the Hausdorff dimension $k$ that is flatter or a more dispersed discrete point set will have a smaller $\beta$-number value. Therefore, we define $\beta$-number as the $p$-th order $d$-dimensional oscillation degree of the set $K$.
Subsequently, we design algorithms to compute the two forms of $\beta$-number and present results in low-dimensional spaces. Moreover, we identify a strong connection between $\beta$-number and Principal Component Analysis (PCA). Specifically, when $p=2$, the $d$-dimensional plane derived from our $\beta$-number is equivalent to the $d$-dimensional subspace in PCA. This finding, validated both theoretically and experimentally, suggests that $\beta$-number is a higher-order extension of PCA.
Finally, we demonstrate the potential of $\beta$-number as a new index for describing data characteristics, applying it successfully to a classification task.


Although our $\beta$-number has yielded satisfactory results in several low-dimensional examples, its computation relies on solving a non-convex optimization problem, which may become a bottleneck in high-dimensional scenarios. Additionally, the task-specific selection of parameters $p$ and $d$ remains an open question for further investigation. In future work, we plan to develop more efficient algorithms for calculating $\beta$-number and explore its application in high-dimensional data analysis.

Beyond the above directions, our $\beta$-number also holds promising potential for intrinsic dimension estimation. For datasets $K \subseteq \mathbb{R}^n$ whose intrinsic structure is strictly contained within a $d_{\text{int}}$-dimensional affine subspace, intrinsic dimension estimation can be performed using the current definition of $\beta$-number. Specifically, we incrementally increase $d$ from 1 to $n$ and compute $\beta_{p,d}(K)$ (the $\beta$-number for $d$-dimensional subspaces under the $L_p$-norm) for each $d$. The smallest $d$ at which $\beta_{p,d}(K) = 0$ serves as the estimator of $d_{\text{int}}$. This is grounded in the fact that if $d < d_{\text{int}}$, $K$ cannot be fully contained in any $d$-dimensional affine subspace, thus $\beta_{p,d}(K) > 0$; in contrast, if $d \geq d_{\text{int}}$, there exists a $d$-dimensional affine subspace that exactly contains $K$, resulting in $\beta_{p,d}(K) = 0$. Here, $p$ may be chosen arbitrarily or fixed (e.g., $p=2$ for Euclidean settings), as the validity of the estimator is independent of $p$. For datasets $K \subset \mathbb{R}^n$ lying on a $d_{\text{int}}$-dimensional manifold, which is a more general scenario where the structure is not strictly contained in any lower-dimensional affine subspace, the current $\beta$-number definition (formulated with respect to $d$-dimensional affine subspaces) has limitations, given that a manifold cannot be perfectly approximated by an affine subspace. To address this, we propose extending the $\beta$-number in future work by generalizing $L$ from $d$-dimensional affine subspaces to $d$-dimensional manifolds.

\newcommand\strttab{\noindent\begin{tabular}{@{}p{10pc}@{}p{21pc}@{}}}

\backhead{Acknowledgements}
Not applicable.

\backhead{Funding statement}
This work was funded by the National Natural Science Foundation of China under Grants 12301661 and 12431018.

\backhead{Competing interests}
The authors state that there is no conflict of interest.

\backhead{Data availability statement}
Data sharing is not applicable to this article.

\appendix
\captionsetup[figure]{font={footnotesize,stretch=1.2},labelfont={footnotesize,bf,it}}
\captionsetup[subfigure]{font={footnotesize,stretch=1.2},labelfont={footnotesize,bf,it}}
\section{Implementation details}\label{Appendix A}
\setcounter{figure}{0}
\renewcommand{\thefigure}{A\arabic{figure}}
\setcounter{table}{0}
\renewcommand{\thetable}{A\arabic{table}}

In this section, we provide the detailed implementation of the numerical algorithms discussed in the main text.

\subsection{Discretization of the integration and standardization of datasets}\label{subsec A.1}

In this section, we introduce a discrete integral form for the $L^p$ version of $\beta$-number and normalize the dataset $K$. Specifically, if $K$ is a bounded set with Hausdorff dimension $k$, the discrete integral form of the $L^p$ version of $\beta$-number is defined as follows:
\begin{equation}\label{Eq A1}
\begin{aligned} 
\beta_{p,d}(K) 
& =\inf _{L}\left(\frac{1}{V(K) \cdot r^{d}} \int_{K}\left(\frac{\operatorname{dist}(y, L)}{r}\right)^{p} d y\right)^{\frac{1}{p}} \\ 
& \approx \inf _{L}\left(\frac{1}{r^{d}} \sum_{i=1}^{n}\left(\left(\frac{\operatorname{dist}\left(y_{i}, L\right)}{r}\right)^{p} \cdot {\frac{\left|\Delta_{i}\right|}{V(K)}}\right)\right)^{\frac{1}{p}}.
\end{aligned}
\end{equation}
Here, $\Delta_{i}$ denotes a volume unit of set $K$, satisfying $\left|\Delta_{i}\right|=\left|\Delta_{j}\right|$ and $\bigcup_{i=1}^{n} \Delta_{i}=K$, where $i, j = 1, 2, \ldots, n$. The point $y_{i}$ represents the center of gravity of each cell $\Delta_{i}$.
Given that $\left|\Delta_{i}\right|=\left|\Delta_{j}\right|$, it follows that $\frac{\left|\Delta_{i}\right|}{V(K)}=\frac{1}{n}$. Therefore, we have transformed the integral over a bounded set with Hausdorff dimension $k$ in the continuous case into a summation form involving a discrete point set with $n$ elements in the discrete case, allowing us to focus solely on the $\beta$-number in the discrete case in subsequent calculations.

Additionally, prior to computation, we normalize the dataset $K$ by scaling it according to the center $o$ of the smallest enclosing ball, such that the radius of the smallest enclosing ball of the scaled set $K_1$ is 1. As demonstrated in section 2.2, scaling $K$ with respect to $o$ does not alter the relative position between $L$ and $K$. This implies that we can search for the optimal plane $L_1$ within the scaled set $K_1$, and then obtain the optimal plane $L$ for $K$ through inverse scaling. Furthermore, the $\beta$-number value of $K$ can be directly computed from the $\beta$-number value of $K_1$. 

\subsection{Optimization methods} \label{subsec A.2}
How can we identify the optimal line and subspace in the $L^\infty$ and $L^p$ versions of $\beta$-number? In this section, we employ optimization methods to search for $L$. For the $L^\infty$ version of $\beta$-number, if $K$ is situated in $\mathbb{R}^2$, an algorithm can be used to find the narrowest cylinder containing $K$. At this stage, according to (\ref{Eq 5}), the corresponding line $L$ serves as the axis of the cylinder. However, identifying the narrowest cylinder containing set $K$ in $\mathbb{R}^3$ or even $\mathbb{R}^n$ proves challenging. For the $L^p$ version of $\beta$-number, no analogous transformation exists as in the $L^\infty$ version to determine the narrowest cylinder. Consequently, in light of these issues and challenges, we aim to convert these problems into optimization problems. 

We utilize a matrix formulation for description. Let $Ay = b$ represent any $d$-dimensional plane in $\mathbb{R}^n$, where $A$ is a full-rank matrix of size $(n-d) \times n$, $y$ and $b$ are column vectors of size $n \times 1$ and $(n-d) \times 1$, respectively. The distance from any point $y_0 \in \mathbb{R}^n$ to $L$ can be expressed as:
\[
\operatorname{dist}(y_0, L) = \| A^T (A A^T)^{-1} (b - A y_0) \|_2.
\]

Thus, the $L^{\infty}$ version of $\beta$-number can be formulated as the following optimization problem:
\begin{equation}\label{Eq A2}
    \beta_{\infty}(K) = \frac{1}{r} \inf_{L \text{ line}} \sup_{x \in K} \operatorname{dist}(x, L) = \min_{A, b} \left( \max_{y \in K} \left( \frac{\| A^T (A A^T)^{-1} (b - A y) \|_2}{r} \right) \right),
\end{equation}
where $A$ is a full-rank matrix in $\mathbb{R}^{(n-1) \times n}$, and $b$ is a column vectors $\mathbb{R}^{(n-1) \times 1}$. The $L^p$ version of $\beta$-number in the discrete case can be expressed as follows:
\begin{equation}
\begin{aligned}
    \beta_{p,d}(K) 
    &= \inf_{L} \left( \frac{1}{m \cdot r^{d}} \sum_{i=1}^{m} \left( \frac{\operatorname{dist}(y_{i}, L)}{r} \right)^{p} \right)^{\frac{1}{p}}\\ 
    &= \min_{A, b} \left( \frac{1}{m \cdot r^{d}} \sum_{i=1}^{m} \left( \frac{\| A^T (A A^T)^{-1} (b - A y_i) \|_2}{r} \right)^{p} \right)^{\frac{1}{p}}.
\end{aligned}
\end{equation}
where $A$ is a full-rank matrix in $\mathbb{R}^{(n-1) \times n}$, and $b$ is a column vector in $\mathbb{R}^{(n-1) \times 1}$. However, since both of these optimization problems are non-convex, using optimization algorithms may result in local minima instead of the global minimum. Therefore, it is necessary to carry out multiple random initializations of the parameters (such as the matrix $A$ and the vector $b$). Standardizing the point set to the unit ball can also help to reduce the search space for $A$ and $b$ to some extent.

Particularly, in the low-dimensional case (where $n \leq 3$), both the definition of the $d$-dimensional plane $L$ and the distance from points in $\mathbb{R}^n$ to $L$ can be explicitly expressed. Therefore, the optimization problem in low-dimensional spaces can be formulated as follows. 
For a finite discrete set $K$ with $m$ elements in $\mathbb{R}^2$, let $B(x, r)$ denote the minimal enclosing ball that contains all points in the set $K$, where $x$ represents the center and $r$ denotes the radius. When $d=1$ , assuming that the straight line $L$ is given by $Ax+By+C=0$, the $L ^\infty$ version and $L ^ p$ version of $\beta$-number can be calculated as follows:
\begin{equation}
    \beta_{\infty}(K) 
    =\dfrac{1}{r} \inf _{L \operatorname{line}} \sup _{x \in K } \operatorname{dist}(x, L) 
    =\min _{A, B, C}\left(\max _{x \in K}\left(\frac{\left|A x_{i 1}+B x_{i 2}+C\right|}{r \sqrt{A^{2}+B^{2}}}\right)\right),
\end{equation} 
\begin{equation}
\beta_{p,1}(K)
=\inf _{L}\left(\frac{1}{m \cdot r^{d}} \sum_{i=1}^{m}\left(\frac{\operatorname{dist}\left(y_{i}, L\right)}{r}\right)^{p}\right)^{\frac{1}{p}}
=\min _{A, B, C}\left(\frac{1}{m \cdot r} \sum_{i=1}^{m}\left(\frac{\left|A x_{i 1}+B x_{i 2}+C\right|}{r \sqrt{A^{2}+B^{2}}}\right)^{p}\right)^{\frac{1}{p}},
\end{equation}
where $x_{i}=(x_{i1},x_{i2}) \in K$. 

Similarly, for $K$ in $\mathbb{R}^3$, when $d=1$, assuming that the optimal straight line is given by $\frac{x-x_{0}}{A}=\frac{y-y_{0}}{B}=\frac{z-z_{0}}{C}$, and the optimal plane is given by $Ax+By+CZ+D=0$, the $L ^\infty$ and $L ^ p$ version of $\beta$-number can be calculated as follows:
\begin{equation}
    \beta_{\infty}(K) 
    =\min _{A, B, C, x_{0}, y_{0}, z_{0}}
\left(\max _{x \in K}\left(\frac{\left|\left(x_{0}-x_{i 1}, y_{0}-x_{i 2}, z_{0}-x_{i 3}\right) \times(A, B, C)\right|}{r \sqrt{A^{2}+B^{2}+C^{2}}}\right)\right),
\end{equation}
\begin{equation}
\beta_{p,1}(K) 
=\min _{A, B, C, x_{0}, y_{0}, z_{0}}
\left(\frac{1}{m \cdot r^{2}}\sum_{i=1}^{m} \left(\frac{\left|\left(x_{0}-x_{i 1}, y_{0}-x_{i 2}, z_{0}-x_{i 3}\right) \times(A, B, C)\right|}{r \sqrt{A^{2}+B^{2}+C^{2}}}\right)^{p}\right)^{\frac{1}{p}},
\end{equation}
\begin{equation}
\beta_{p,2}(K) 
=
\min _{A, B, C, D}\left(\frac{1}{m \cdot r^{2}} \sum_{i=1}^{m}\left(\frac{\left|A x_{i 1}+B x_{i 2}+C x_{i 3}+D\right|}{r \sqrt{A^{2}+B^{2}+C^{2}}}\right)^{p}\right)^{\frac{1}{p}},
\end{equation}
where $x_{i}=(x_{i1},x_{i2},x_{i3}) \in K$.

In the low-dimensional case (where $n \leq 3$), after transforming the problem of solving $L$ into an optimization problem, we use MATLAB's built-in \texttt{fminimax} and \texttt{fminunc} functions to find the minimum of the aforementioned unconstrained optimization problem. Additionally, both functions utilize local optimization algorithms, which may result in locally optimal solutions near the initial values rather than globally optimal solutions. Therefore, it is necessary to appropriately traverse their initial values to ensure the acquisition of the global optimal solution. Our previous standardization of the set $K$ to the unit ball can also somewhat reduce the number of initial point traversals required. It is worth noting that in subsequent experiments, we find that the $\beta$-number value of $K$ in the unit ball is insensitive to the choice of the initial value. This implies that we only need to traverse a very small number of initial value points to obtain the global optimal solution, thereby significantly reducing the computation time.

\subsection{Domain partition and approximation algorithms of $\beta$-number}
Prior to proceeding with the computation, it is necessary to perform a region decomposition of the set $K$. For $K$ lying in $\mathbb{R}^2$ and $\mathbb{R}^3$, we employ triangulation and tetrahedral meshing, respectively.

\begin{figure}[ht]
    \centering
    \begin{subfigure}{.5\textwidth}
        \centering
        \includegraphics[width=0.8\textwidth]{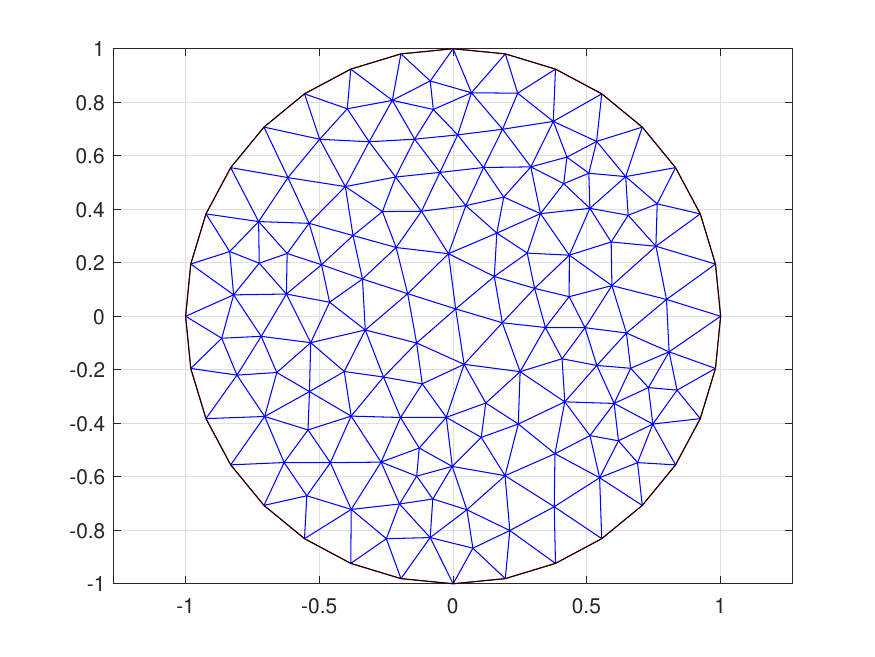}
        \caption{Regular set}
        \label{subfig:1a}
    \end{subfigure}%
    \begin{subfigure}{.5\textwidth}
        \centering
        \includegraphics[width=0.8\textwidth]{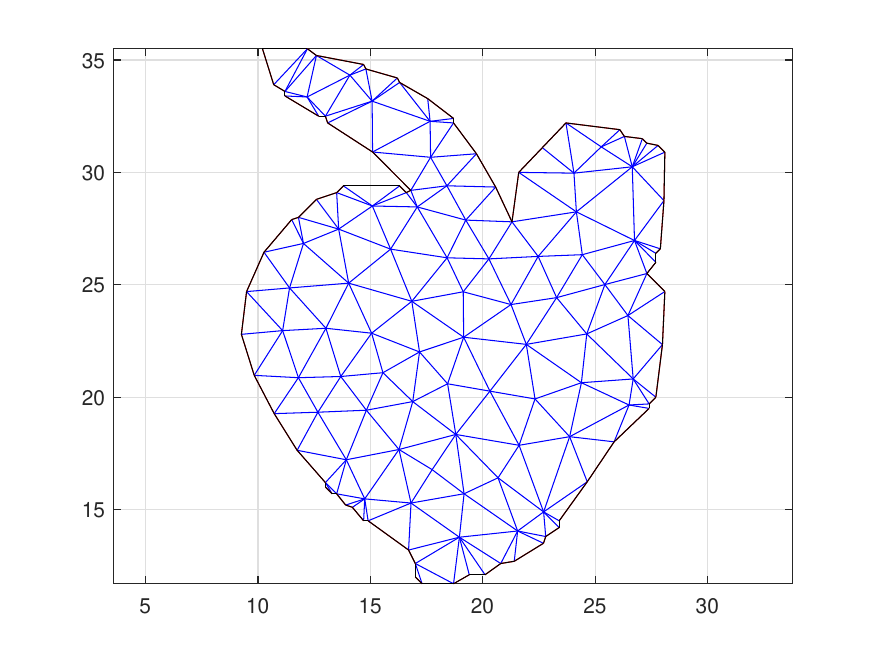}
        \caption{Irregular set}
        \label{subfig:1b}
    \end{subfigure}
    \caption{Illustration of 2D region triangulation. The black lines represent the contours of the region, and the blue lines represent the subdivision lines of the region. (\textbf{a}) Triangulation of a regular-shaped region, such as a unit circle. (\textbf{b}) Triangulation of an irregular-shaped region, such as an arbitrary hand-drawn area. }
    \label{fig1}
\end{figure}

When $K$ is in $\mathbb{R}^2$, if $K$ is a regularly shaped set, which we can obtain its contour in advance, then its triangulation can be easily obtained using the pdeModel plugin available in MATLAB. For instance, the triangulation of a unit circle in $\mathbb{R}^2$ is illustrated in \cref{subfig:1a}. However, for certain irregularly shaped sets, such as a randomly generated region, we propose to extract region information from an image. Specifically, we begin by drawing the desired region on a drawing board and subsequently process the image to extract the relevant information. After reading and binarizing the image, we extract its contour information to obtain the region's outline and the points contained within it, which can be simplified according to a specific scale. Finally, the \texttt{triangulation} and \texttt{generateMesh} functions are employed to create the triangular mesh. For instance, the triangulation of any hand-drawn area in $\mathbb{R}^2$ is illustrated in \cref{subfig:1b}.

When $K$ is in $\mathbb{R}^3$, similarly, if $K$ is a regularly shaped set, which we can obtain its contour in advance, we first define several contour points and interior points of $K$. Based on these points, we can utilize the \texttt{delaunay} and \texttt{trimesh} functions to construct a tetrahedral mesh. For instance, the tetrahedral decompositions of a unit sphere and a cube in $\mathbb{R}^3$ are illustrated in \cref{fig2}.
\begin{figure}[ht]
    \centering
    \begin{subfigure}{.5\textwidth}
        \centering
        \includegraphics[width=1\textwidth]{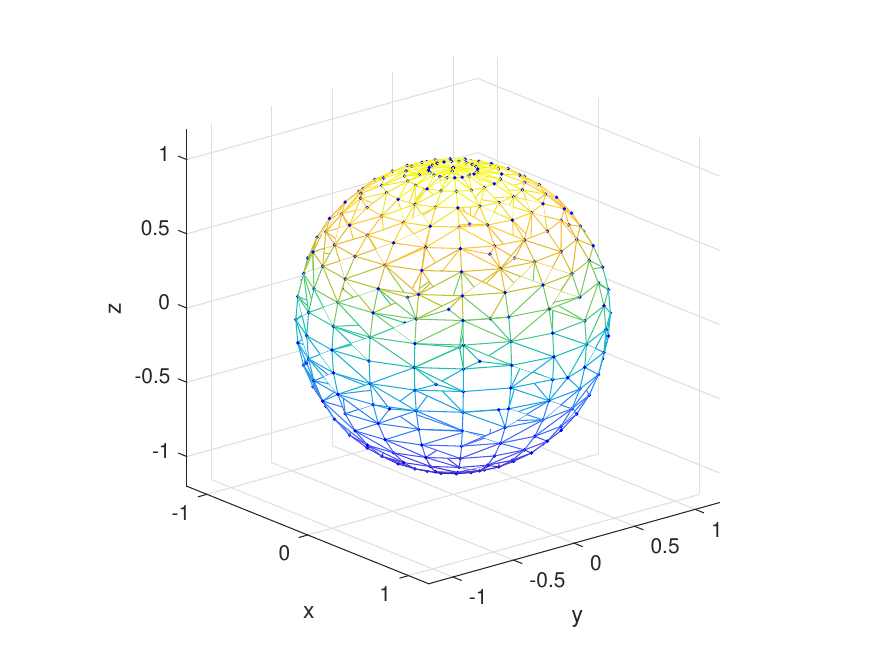}
        \caption{A tetrahedral decomposition of a unit sphere}
        \label{subfig:2a}
    \end{subfigure}%
    \begin{subfigure}{.5\textwidth}
        \centering
        \includegraphics[width=1\textwidth]{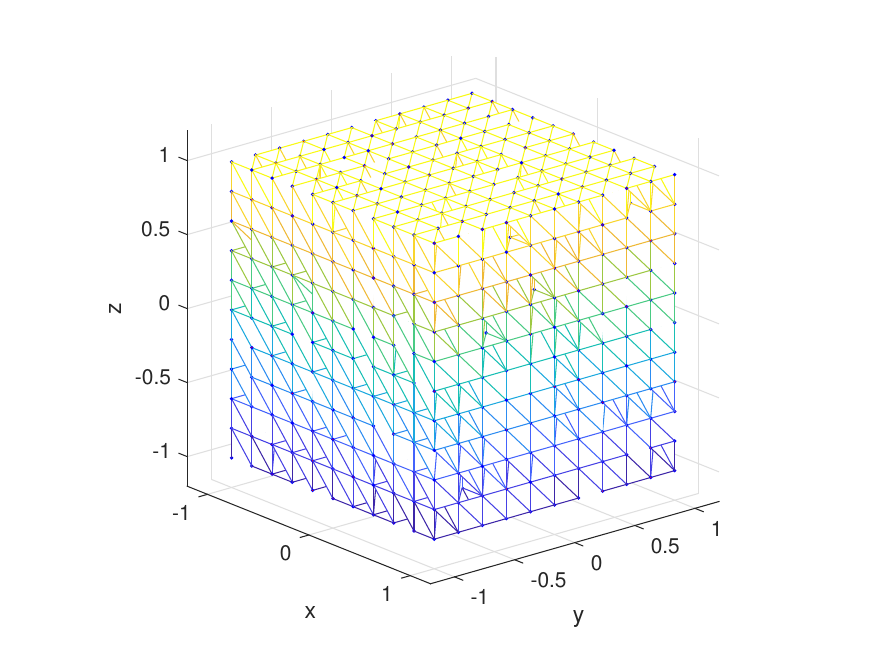}
        \caption{A tetrahedral decomposition of a cube}
        \label{subfig:2b}
    \end{subfigure}
    \caption{Illustration of tetrahedral decompositions for regular regions of 3D shapes. (\textbf{a}) Tetrahedral decomposition of the unit sphere. (\textbf{b}) Tetrahedral decomposition of a cube with edge length of 2.}
    \label{fig2}
\end{figure}

However, for certain irregular sets in $\mathbb{R}^3$ where contour points are not readily available, even if we can obtain their interior and contour points through alternative methods, the tetrahedral mesh constructed by using the \texttt{delaunay} and \texttt{trimesh} functions tends to be coarse and fails to capture local details, as illustrated in \cref{subfig.3b}. Although numerous software tools and algorithms exist for tetrahedral partitioning of irregular 3D objects, they all require the input of the object's edge points, which are not readily accessible in advance. To compute $\beta$-number of any 3D volume, we propose to extract region information from a 3D image. Specifically, after loading the 3D image, we first select one of the three dimensions, the $x$, $y$, or $z$ axis, and slice it into several 2D images. Then, similar to the processing of 2D sets mentioned earlier, we obtain the edge points and interior points of each slice. Due to the poor performance of tetrahedral partitioning when integrating points from all slices, we perform tetrahedral partitioning on the points from any two adjacent slices, aiming to reconstruct both the overall and detailed information of the set. We then stack all slices sequentially to obtain the tetrahedral partition of the entire set. After slicing and performing tetrahedral partitioning in the directions of the $x$, $y$, and $z$ axes, we select the dimension that yields the best reconstruction details for subsequent calculations. For instance, for a 3D image of a liver in $\mathbb{R}^3$, its tetrahedral triangulation is illustrated in the \cref{fig3}.

\begin{figure}[htbp]
    \centering
    \vspace{0.5cm} 
    \begin{subfigure}[b]{0.3\textwidth}
        \centering
        \includegraphics[width=1.0\linewidth]{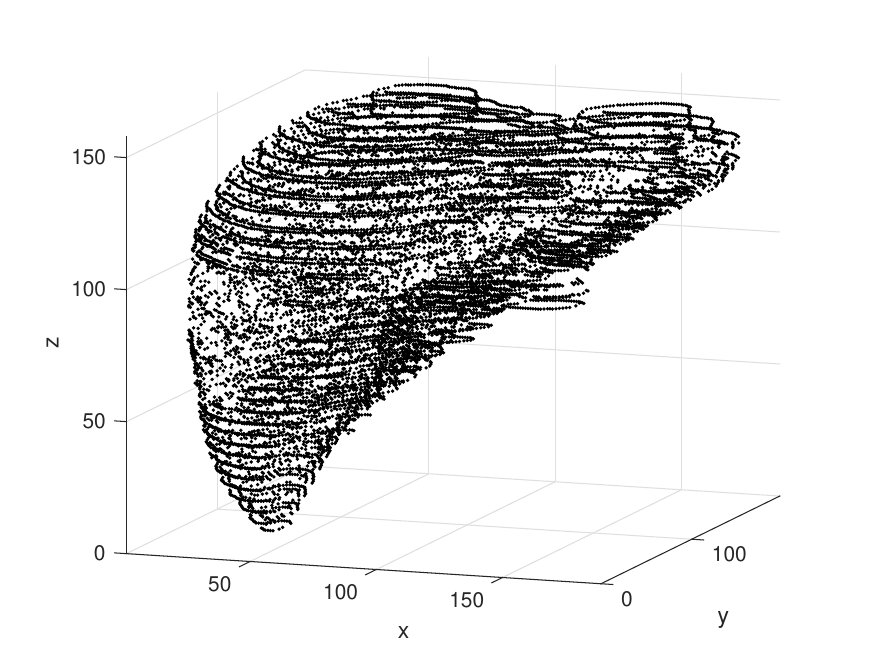} 
        \caption{}
        \label{subfig.3a}
    \end{subfigure}
    \hfill
    \begin{subfigure}[b]{0.3\textwidth}
        \centering
        \includegraphics[width=1.0\linewidth]{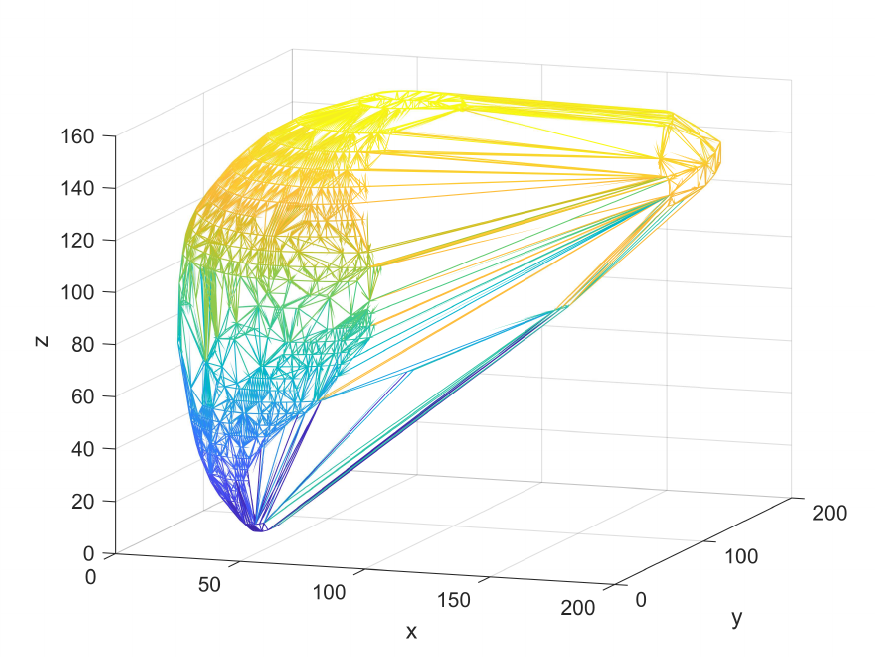} 
        \caption{}
        \label{subfig.3b}
    \end{subfigure}
    \hfill
    \begin{subfigure}[b]{0.3\textwidth}
        \centering
        \includegraphics[width=1.0\linewidth]{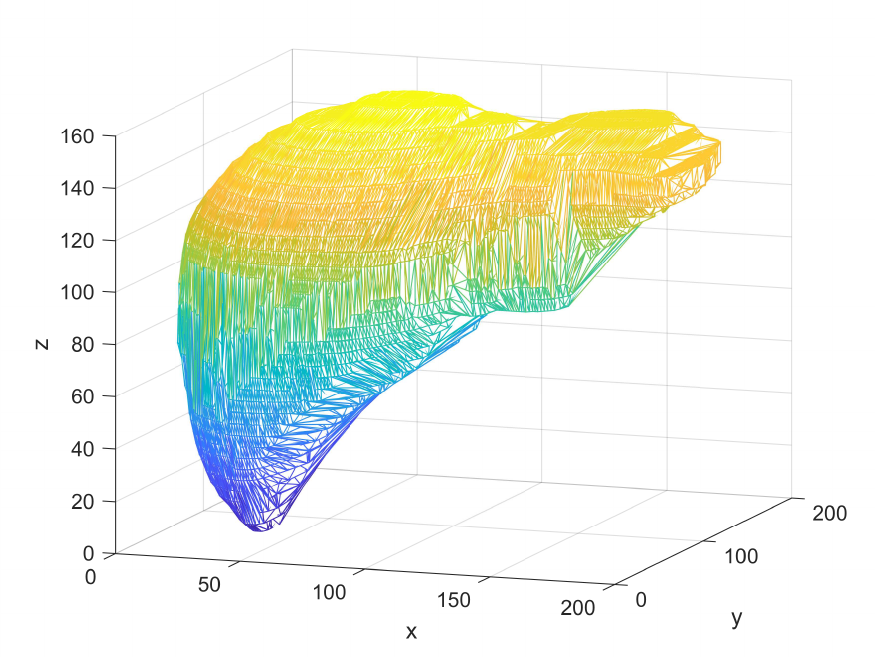} 
        \caption{}
        \label{subfig.3c}
    \end{subfigure}

    \caption{Illustration of tetrahedral decompositions for an irregular volume of 3D shape. (\textbf{a}) Set of voxels in the 3D image. (\textbf{b}) Direct region decomposition of the voxel set.  (\textbf{c}) Tetrahedral decomposition of two adjacent planes in the $xoz$ plane, based on the $y$-axis fusion.}
    \label{fig3}
\end{figure}

Additionally, for a bounded set $K$ in $\mathbb{R}^n$ with Hausdorff dimension $k$, it is generally not feasible to partition $K$ into $n$ units of equal volume, as described in \cref{subsec A.1}  of the theoretical framework. Consequently, in practical computations, the result of the region partitioning is a finite discrete set of points with weights, where each point's weight reflects the percentage of the volume of the microelement containing that centroid relative to the total volume of the set $K$, expressed as follows:
\[
\beta_{p,d}(K) 
\approx \inf _{L}\left(\frac{1}{r^{d}} \sum_{i=1}^{n}\left(w_{i} \cdot \left(\frac{\operatorname{dist}\left(y_{i}, L\right)}{r}\right)^{p} \right)\right)^{\frac{1}{p}},
\]
where $\Delta_{i}$ denotes a partition unit of the set $K$, satisfying $\bigcup_{i=1}^{n} \Delta_{i}=K$, $y_{i}$ represents the centroid of each cell $\Delta_{i}$, and the weight $w_{i}=\left|\Delta_{i}\right|/V(K)$, for $i= 1, 2, \ldots, n$. The computational process aligns with that of a finite discrete point set with uniform weights, except that the weight $1/m$ is replaced by $w_{i}$.

\section{Experiments results}\label{Appendix B}

\setcounter{figure}{0}
\renewcommand{\thefigure}{B\arabic{figure}}
\setcounter{table}{0}
\renewcommand{\thetable}{B\arabic{table}}

In this section, we provide the experiments results of the low dimensional cases discussed in the main text. All experiments were conducted on an Intel Core i7-8550U CPU with 16 GB RAM.

\subsection{Bounded sets $K$ with Hausdorff dimension $k$}
First, consider a bounded set $K$ with Hausdorff dimension $k$. When $k = n$, if the smallest enclosing ball of $K$ coincides with $K$ itself, the symmetry of the sphere implies that for any given $p$ and $d$, the center of the sphere must lie on the $d$-dimensional plane $L$. In low-dimensional spaces, the $\beta$-number of $K$ can be computed directly through integration.

For instance, when $n=k=2$, and $d=1$, suppose $K$ is a disk with a radius of $r$ in $\mathbb{R}^2$. For any given $p$, $L$ is any straight line passing through the center of the disk. Thus, we can integrate directly to obtain:
\begin{equation}\label{Eq 26}
\begin{aligned}
    \beta_{p,1}(K)
    &=\inf _{L}\left(\frac{1}{\pi r^2 \cdot r^{1}} \int_{K}\left(\frac{\operatorname{dist}(y, L)}{r}\right)^{p} d y\right)^{\frac{1}{p}}
    =\left(\frac{1}{\pi r^3} \int_{0}^{r} \int_{0}^{2 \pi} \rho \cdot\left(\frac{\rho|\sin \theta|}{r}\right)^{p} d \theta d \rho\right)^{\frac{1}{p}}\\
    &=\left(\frac{1}{\pi(p+2)r} \cdot\left(\int_{0}^{2 \pi}|\sin \theta|^{p} d \theta\right)\right)^{\frac{1}{p}}.
\end{aligned}
\end{equation}
When $p$ and $r$ are any given positive integers, the above expression is a constant. 

Similarly, when $n=k=3$, suppose $K$ is a ball with a radius of $r$ in $\mathbb{R}^3$. For any given $p$, it can be concluded that:
\begin{equation}\label{Eq 27}
     \beta_{p,1}(K) =
     \left(\frac{3}{r^4} \cdot \int_{0}^{r}\left(\frac{1}{p+2}\left(1-z^{2}\right)^{\frac{p+2}{2}}\right) d z\right)^{\frac{1}{p}},
\end{equation}
\begin{equation}\label{Eq 28}
     \beta_{p,2}(K) =
     \left(\frac{3}{2\pi r^{5}} 
     \int_{0}^{2 \pi}|\sin \theta|^{p} d \theta\cdot \int_{0}^{r}\left(\frac{1}{p+2}\left(1-x^{2}\right)^{\frac{p+2}{2}}\right) d x\right)^{\frac{1}{p}}.
 \end{equation}

If $K$ is a bounded set with Hausdorff dimension $k$, for a set $K$ in $\mathbb{R}^2$, where $k$ could be 1 for a segment of a spatial curve or 2 for a two-dimensional region, we partition the region according to its Hausdorff dimension, discretize the integrals, and compute two forms of $\beta$-number. Specifically, to facilitate comparison, we first scale all sets to fit within a unit circle centered at the center of the minimum enclosing circle, i.e., with $r=1$. Then, if the Hausdorff measure of a set is 1, we perform a line segment partitioning, recording the midpoint and length of each segment, thereby transforming the problem of finding the optimal line into an optimization problem. We utilize the \texttt{fminimax} and \texttt{fminunc} functions in MATLAB to solve the line integral and compute $\beta$-number. If the Hausdorff measure of a set is 2, we perform a triangulation, recording the centroid and area of each triangle, solving the area integral, and computing $\beta$-number. The numerical results for the three example sets are shown in \cref{fig4}.
\begin{figure}[htbp]
    \centering
    \begin{tabular}{ccc}
        \multicolumn{1}{c}{\textbf{One-dimensional Segment}} & \multicolumn{1}{c}{\textbf{Two-dimensional Unit Circle}} & \multicolumn{1}{c}{\textbf{Two-dimensional Arbitrary Shape}} \\
        \subcaptionbox{$\beta_{\infty}(K)$=0.5726}{\includegraphics[width=0.3\textwidth]{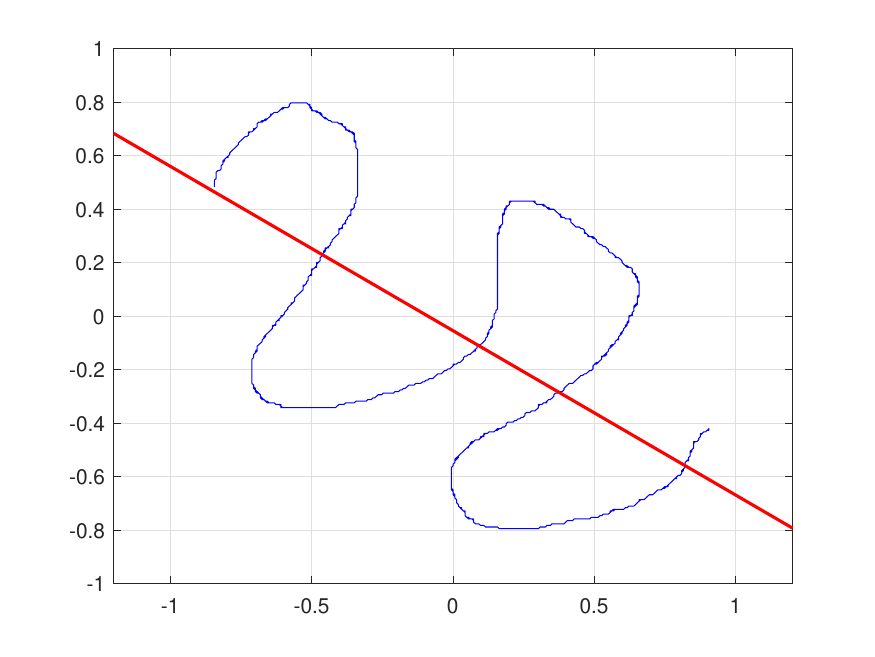}} &
        \subcaptionbox{$\beta_{\infty}(K)$=0.9749}{\includegraphics[width=0.3\textwidth]{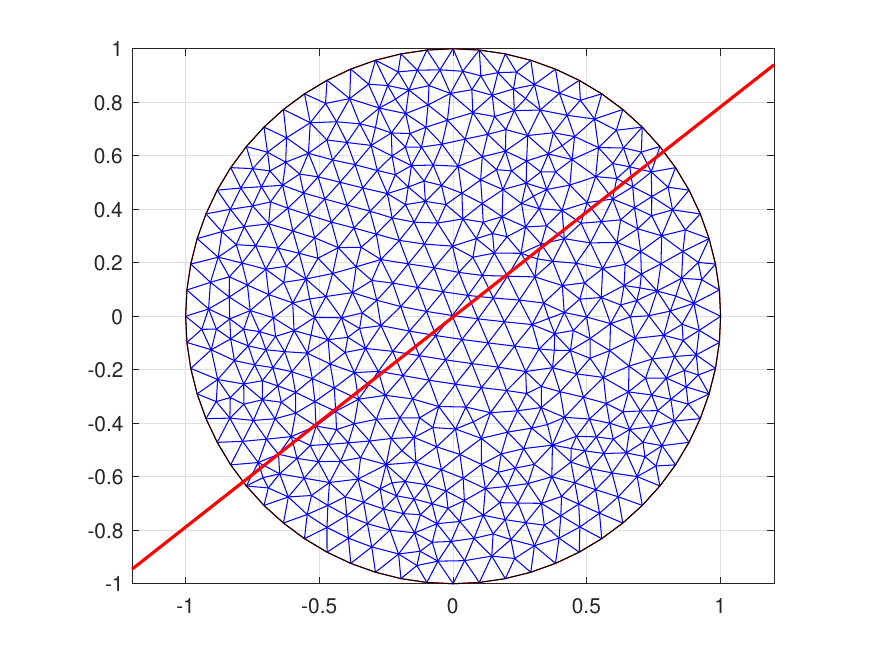}} &
        \subcaptionbox{$\beta_{\infty}(K)$=0.7512}{\includegraphics[width=0.3\textwidth]{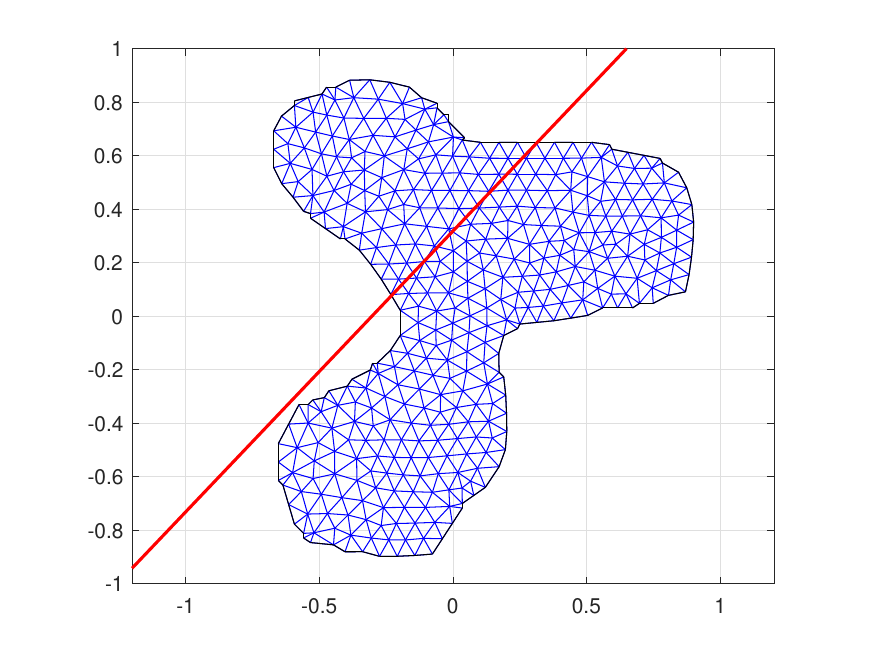}} \\
        \subcaptionbox{$\beta_{2,1}(K)$=0.3502}{\includegraphics[width=0.3\textwidth]{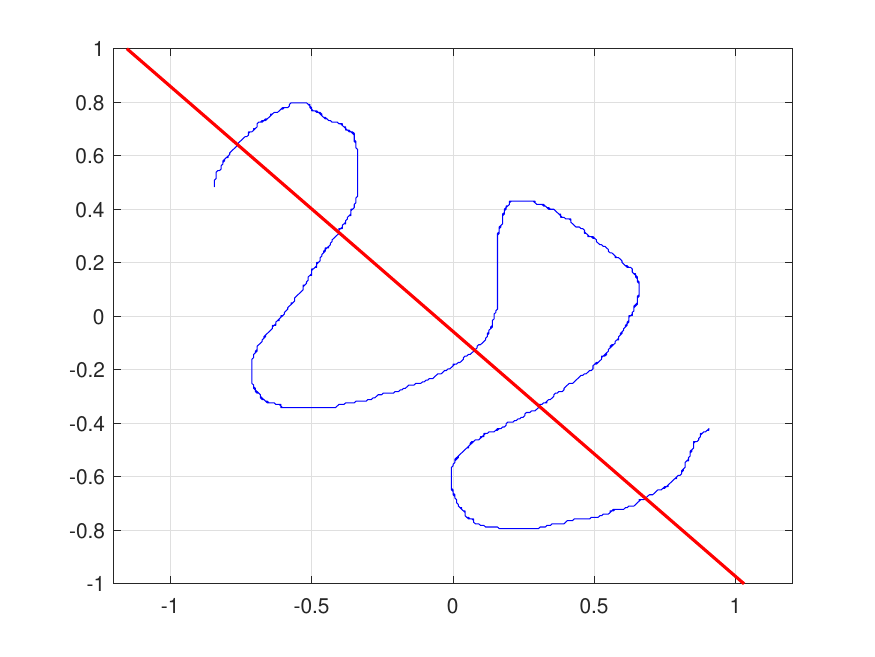}} &
        \subcaptionbox{$\beta_{2,1}(K)$=0.4993}{\includegraphics[width=0.3\textwidth]{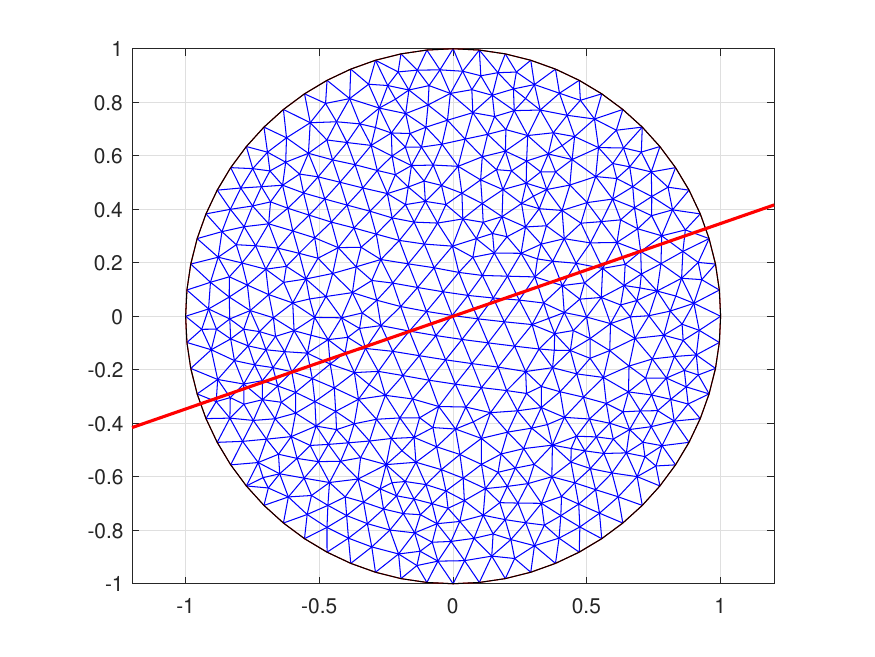}} &
        \subcaptionbox{$\beta_{2,1}(K)$=0.3522}{\includegraphics[width=0.3\textwidth]{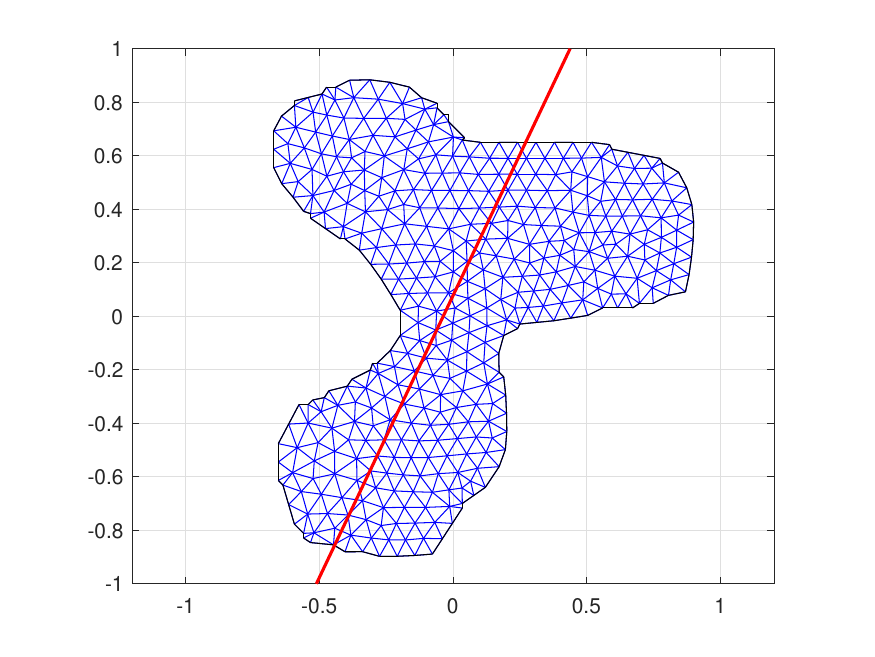}} \\
    \end{tabular}
    \caption{Illustration of the computation results for the set $K$ in $\mathbb{R}^2$. In each figure, black lines represent the boundaries of the region, blue lines represent the subdivision lines, and red lines represent the optimal subspaces $L$. At the bottom of each figure, the corresponding $\beta$-number value is presented. \textbf{(a),(d)} For a planar line segment, the $L^\infty$ version of $\beta$-number value is 0.5726, and the $L^2$ version of $\beta$-number is 0.3502. \textbf{(b),(e)} For a unit circle, the $L^\infty$ version of $\beta$-number is 0.9749, and the $L^2$ version of $\beta$-number is 0.4993. \textbf{(c),(f)} For a planar region, the $L^\infty$ version of $\beta$-number is 0.7512, and the $L^2$ version of $\beta$-number is 0.3522.
}
    \label{fig4}
\end{figure}

As shown in \cref{fig4}, for the two-dimensional unit circle, when $p=2$ and $d=1$, the optimal subspace $L$ is any line passing through the center of the disk. The $L^2$ version of $\beta$-number, calculated as 0.4993, numerically matches the theoretical result of 0.5 derived from (\ref{Eq 26}). Moreover, for any given planar line segment and planar region depicted in the figure, the calculated $\beta$-numbers are less than the theoretical upper bound derived in \cref{subsec 2.1}.

Additionally, consider the three elliptical regions $K_1$, $K_2$, and $K_3$ in $\mathbb{R}^2$, as shown in \cref{fig5}. They represent a circle with radius 1, an ellipse with major axis 2 and minor axis 1.6, and an ellipse with major axis 2 and minor axis 0.8, all centered at the origin. When $p=2$ and $d=1$, $\beta_{2,1}(K_1)=0.4993$, $\beta_{2,1}(K_2)=0.3994$, $\beta_{2,1}(K_3)=0.1998$. It is evident that their optimal line $L$ is the line along their major axes. Based on the definition of $\beta$-number and considering the geometric configurations of the three ellipses, we observe that the flatter the ellipse, the more closely it aligns with its major axis, leading to a smaller reduced oscillation and $\beta$-number. This observation is consistent with our perspective in \cref{sub 3.1}, where we characterize the $L^p$ version of $\beta$-number for set $K$ as the "$p$-th order $d$-dimensional oscillation degree" of $K$.

\begin{figure}[h]
    \centering
    \begin{minipage}{0.45\textwidth}
        \centering        \includegraphics[width=\linewidth]{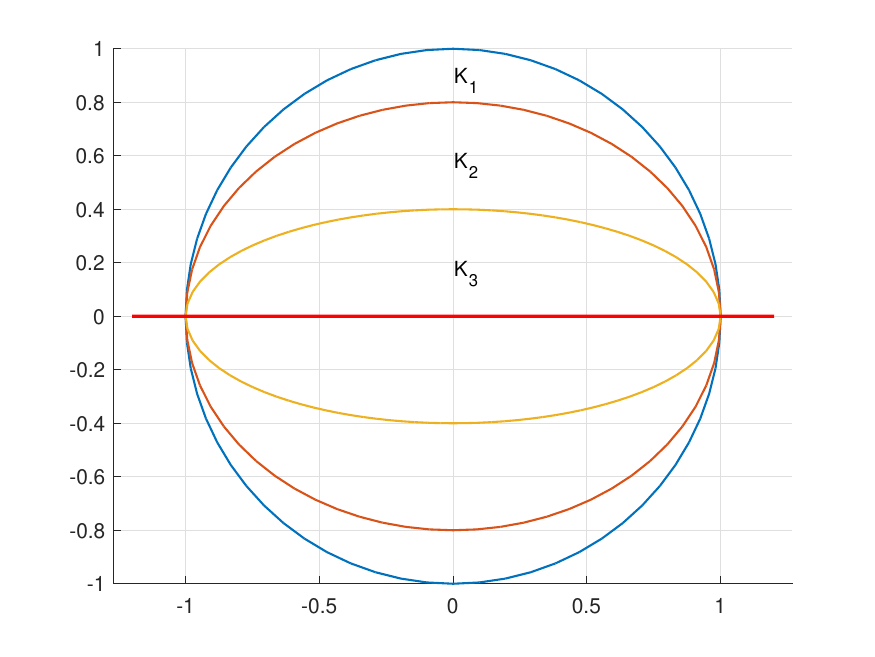}
    \end{minipage}%
    \begin{minipage}{0.45\textwidth}
        \centering
        \captionof{figure}{The schematic diagram of the calculation results for $K_1$, $K_2$, and $K_3$. The three elliptical regions are represented by different colors. When $p=2$, the optimal lines for the three sets coincide with the major axis of the ellipse and are represented by a red line.}
        \label{fig5}
    \end{minipage}
\end{figure}
 
For a set $K$ in $\mathbb{R}^3$, it can represent a one-dimensional spatial curve, a two-dimensional surface, or a three-dimensional solid. The methodology for spatial curve segments and two-dimensional planes is analogous to that in the two-dimensional case. For two-dimensional surfaces within a three-dimensional space, we first perform a subdivision based on their projections onto the coordinate planes and subsequently map these subdivisions back onto the surface to achieve triangular meshing. In the case of three-dimensional solids, if the shapes are regular, meaning that the interior and boundary points can be predetermined, we can directly apply tetrahedral meshing. Conversely, for irregular shapes, we extract set information from 3D images and construct tetrahedral meshing between adjacent slices along the coordinate planes to finalize the complete tetrahedral meshing. After determining the centroid and volume of each tetrahedron in the three-dimensional solid, we transform the tasks of finding the optimal line and plane into optimization problems. We then evaluate the volume integral using the discrete form of the three-dimensional integral and compute the $\beta$-number. \cref{fig6} presents numerical experiment examples for several sets in the three-dimensional space.

\begin{figure}[h]
    \centering
    \begin{tabular}{ccc}
        \multicolumn{1}{c}{\textbf{One-dimensional curve segment}} & \multicolumn{1}{c}{\textbf{Two-dimensional surface}} & \multicolumn{1}{c}{\textbf{Three-dimensional solid}} \\
        \subcaptionbox{$\beta_{\infty}(K)$=0.8000}{\includegraphics[width=0.3\textwidth]{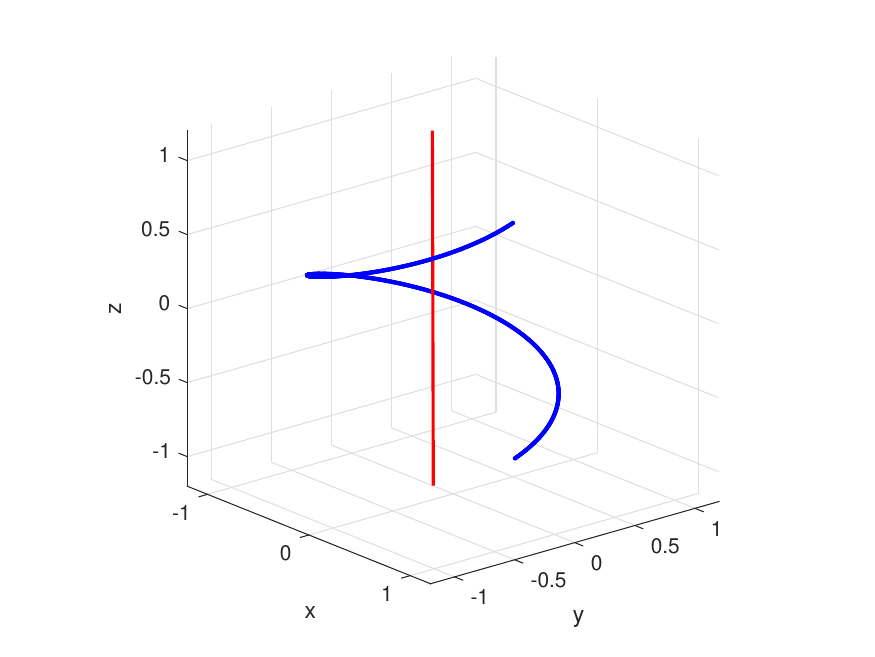}} &
        \subcaptionbox{$\beta_{\infty}(K)$=0.6973}{\includegraphics[width=0.3\textwidth]{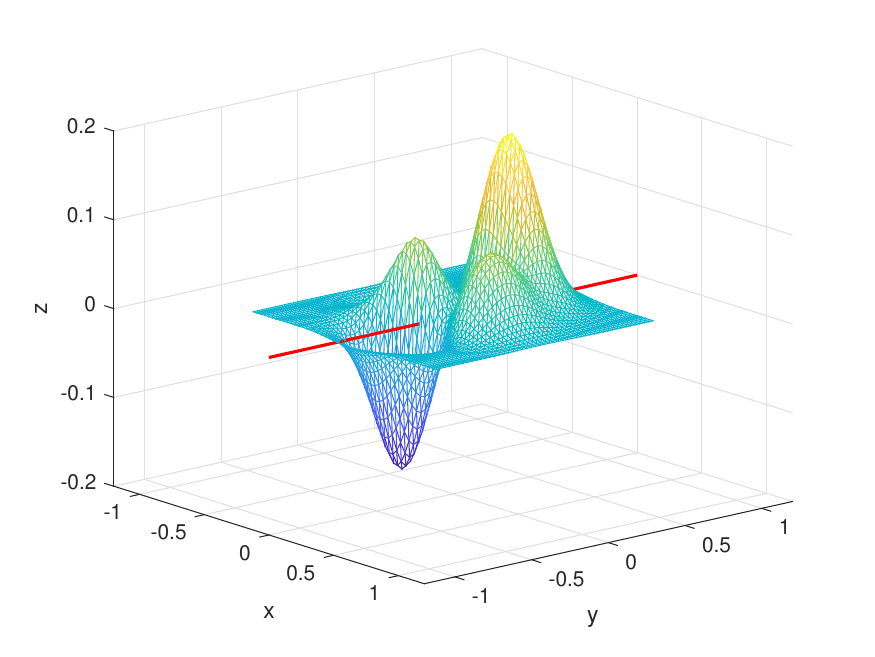}} &
        \subcaptionbox{$\beta_{\infty}(K)$=0.9975}{\includegraphics[width=0.3\textwidth]{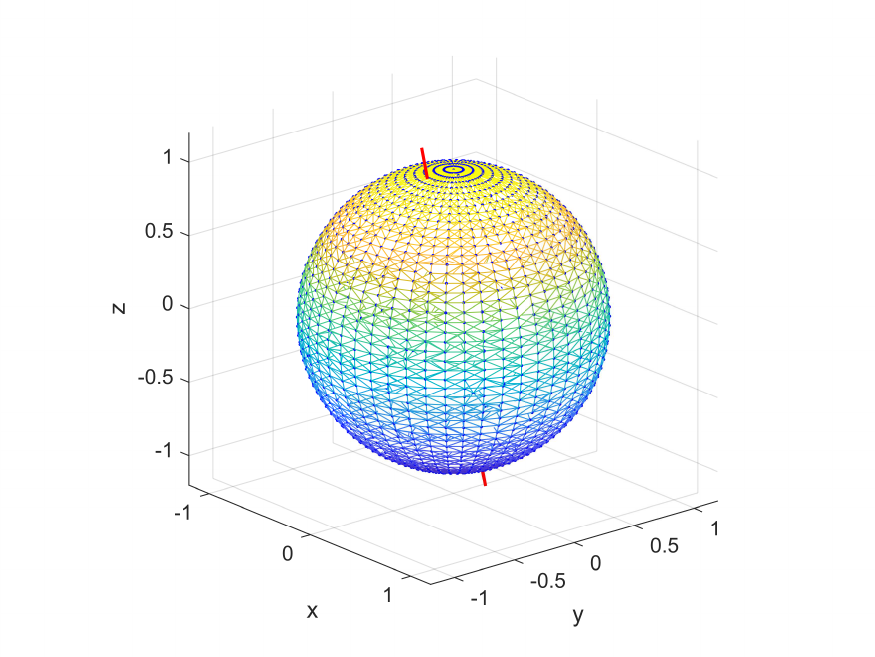}} \\
        \subcaptionbox{$\beta_{2,1}(K)$=0.6133}{\includegraphics[width=0.3\textwidth]{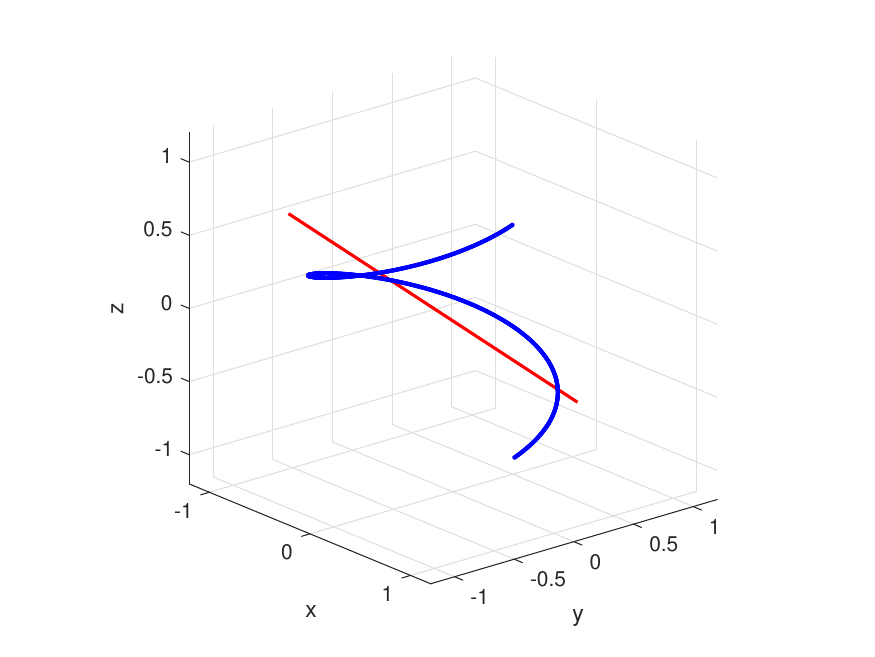}} &
        \subcaptionbox{$\beta_{2,1}(K)$=0.4000}{\includegraphics[width=0.3\textwidth]{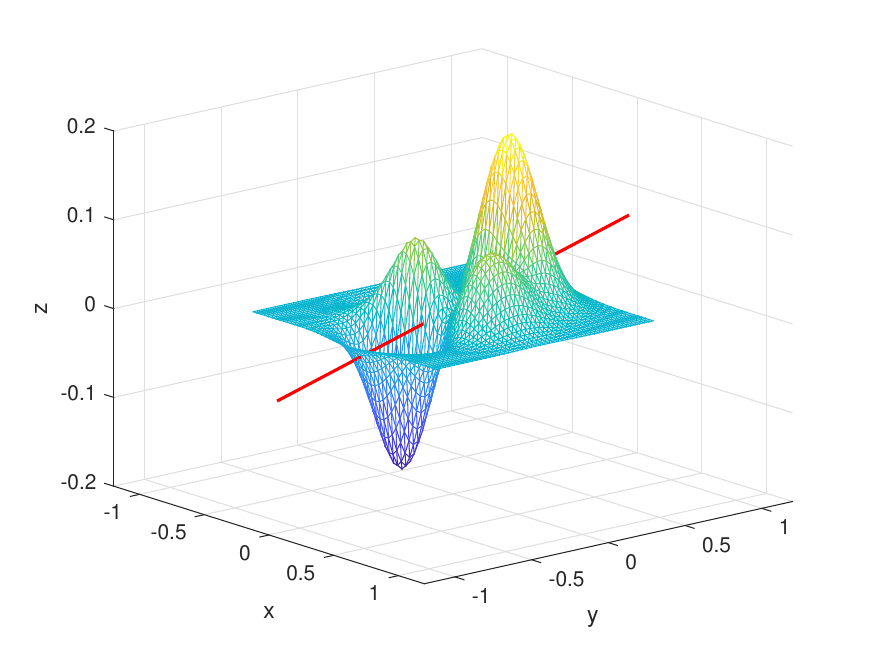}} &
        \subcaptionbox{$\beta_{2,1}(K)$=0.6301}{\includegraphics[width=0.3\textwidth]{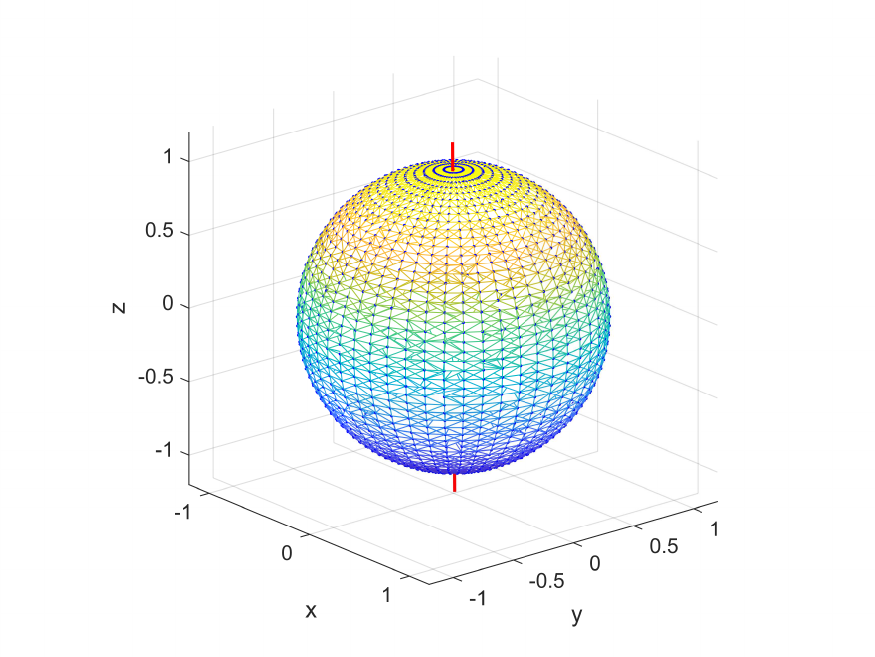}} \\
        \subcaptionbox{$\beta_{2,2}(K)$=0.2368}{\includegraphics[width=0.3\textwidth]{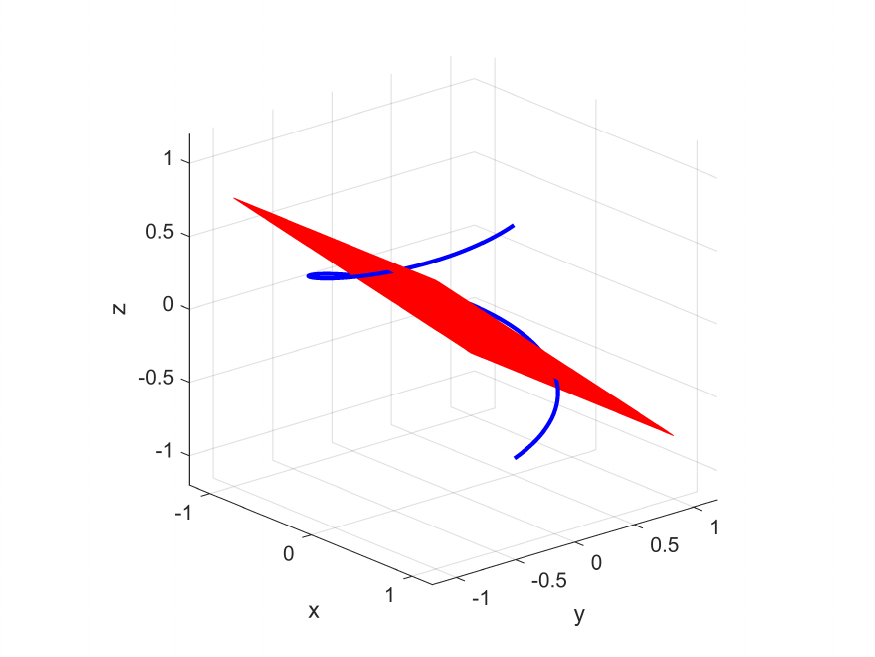}} &
        \subcaptionbox{$\beta_{2,2}(K)$=0.0427}{\includegraphics[width=0.3\textwidth]{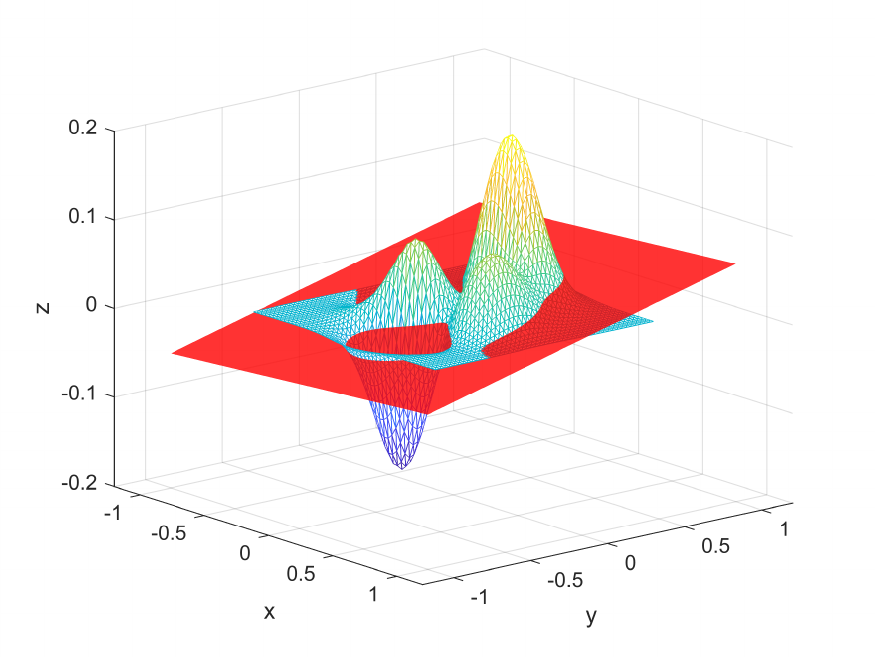}} &
        \subcaptionbox{$\beta_{2,2}(K)$=0.4456}{\includegraphics[width=0.3\textwidth]{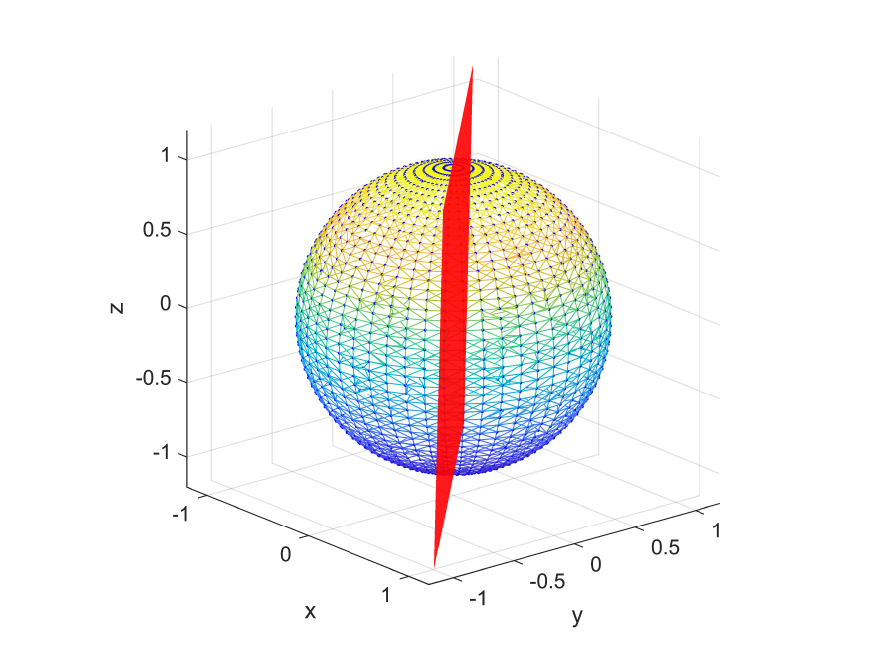}}
    \end{tabular}
    \caption{Illustration of the computation results for the set $K$ in $\mathbb{R}^3$. In each figure, red lines represent the optimal subspaces $L$. At the bottom of each figure, the corresponding $\beta$-number value is shown. The first column, second column, and third column represent the experimental results for the sets of one-dimensional curve segments, two-dimensional surfaces, and three-dimensional solids, respectively. The first row, second row, and third row calculate $\beta$-number in the $L^{\infty}$ form, $L^p$ form ($p=2, d=1$), and $L^p$ form ($p=2, d=2$), respectively.    
}
    \label{fig6}
\end{figure}

From the above experiments, we can also observe that for a set $K$ in $\mathbb{R}^3$, if $k = 1$, corresponding to the first column in \cref{fig6}, then when $ d = 1 $, the line $ L $ corresponding to $\beta$-number can be considered the optimal subspace for dimensionality reduction of $K$. When $d = 2$, since $k < d$, the plane $L$ identified by the $\beta$-number can be considered the optimal embedding subspace for the set $K$. If its $\beta$-number is non-zero for $d = 1$ but zero for $d = 2$, it can be concluded that $K$ is a planar curve. Furthermore, when $K$ is a three-dimensional unit ball and $p=2$, the calculated values $\beta_{2,1}(K)=0.6301$ and $\beta_{2,1}(K)=0.4456$ numerically match the results of 0.6324 and 0.4472 derived from (\ref{Eq 27}) and (\ref{Eq 28}).

In addition, if the set $K$ consists of multiple spatial curve segments or multiple planar regions within $\mathbb{R}^3$, the computation of their $\beta$-numbers requires partitioning each subset according to its dimension. Subsequently, the partitioned results are integrated to solve the optimization problems. The optimal subspaces identified during the computation of the $L^p$ version of $\beta$-number are illustrated in \cref{fig7}.

\begin{figure}[htbp]
    \centering
    \begin{subfigure}{.5\textwidth}
        \centering
        \includegraphics[width=1\textwidth]{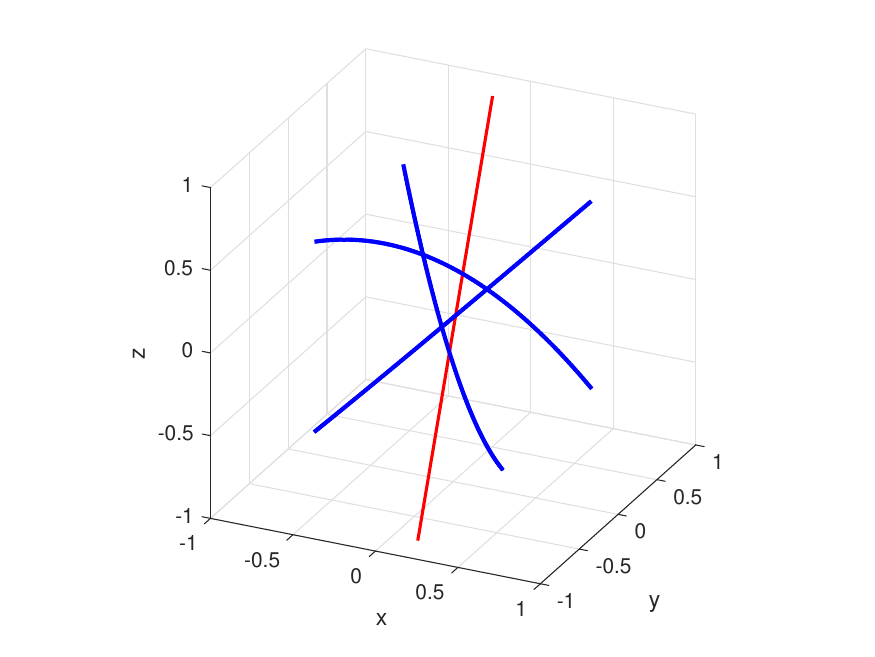}
        \caption{$\beta_{2,1}(K)$=0.4546}
        \label{subfig:7a}
    \end{subfigure}%
    \begin{subfigure}{.5\textwidth}
        \centering
        \includegraphics[width=1\textwidth]{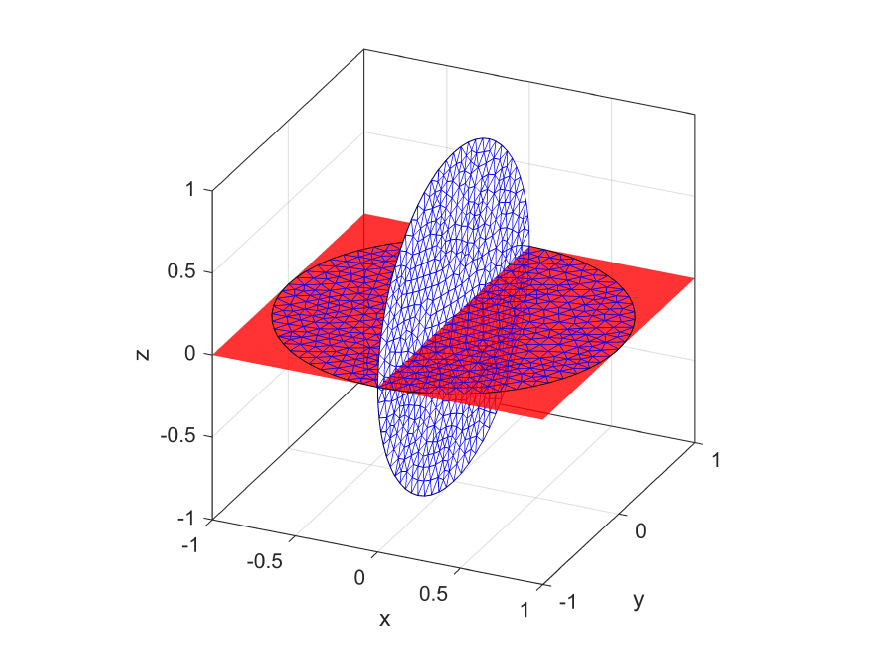}
        \caption{$\beta_{2,2}(K)$=0.3530}
        \label{subfig:7b}
    \end{subfigure}
    \caption{Experiment examples of multiple line segments and planes in $\mathbb{R}^3$. Blue ones represents the partitioned segments and regions, while red ones denotes the optimal lines and planes. (\textbf{a}) The optimal dimensionality reduction line for three spatial curve segments. (\textbf{b}) The optimal dimensionality reduction plane for two planes. }
    \label{fig7}
\end{figure}

\subsection{Finite discrete point sets}
If $K$ is a finite discrete point set, we can directly transform the $L^\infty$ version and the discrete $L^p$ version of $\beta$-number into optimization problems for computation. The numerical results for the two example sets are shown in \cref{fig8}.

\begin{figure}[htbp]
    \centering
    \begin{subfigure}[b]{0.45\textwidth}
        \centering
        \includegraphics[width=0.7\linewidth]{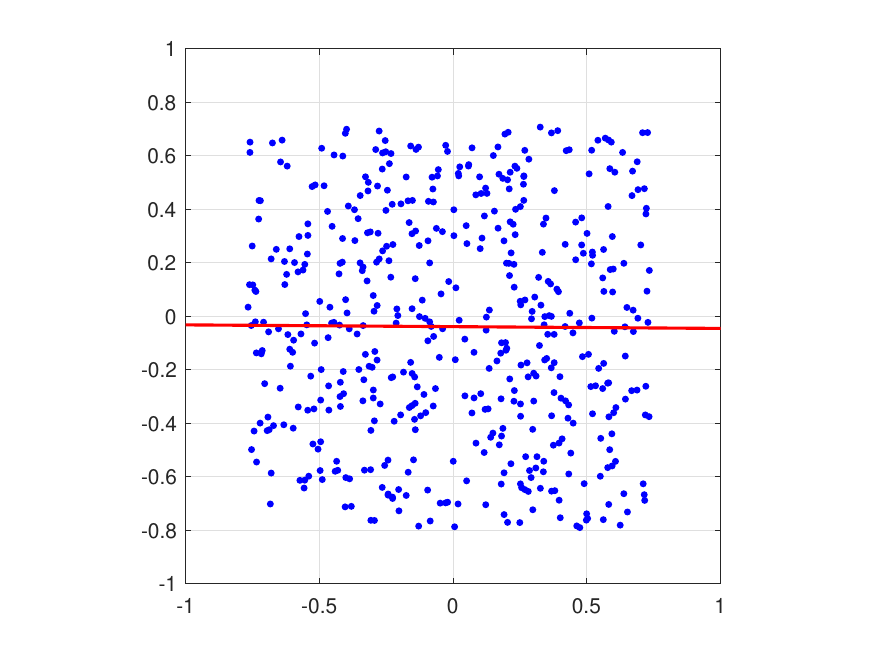} 
        \caption{$\beta_{\infty}(K)$=0.7499}
    \end{subfigure}
    \hfill
    \begin{subfigure}[b]{0.45\textwidth}
        \centering
        \includegraphics[width=0.7\linewidth]{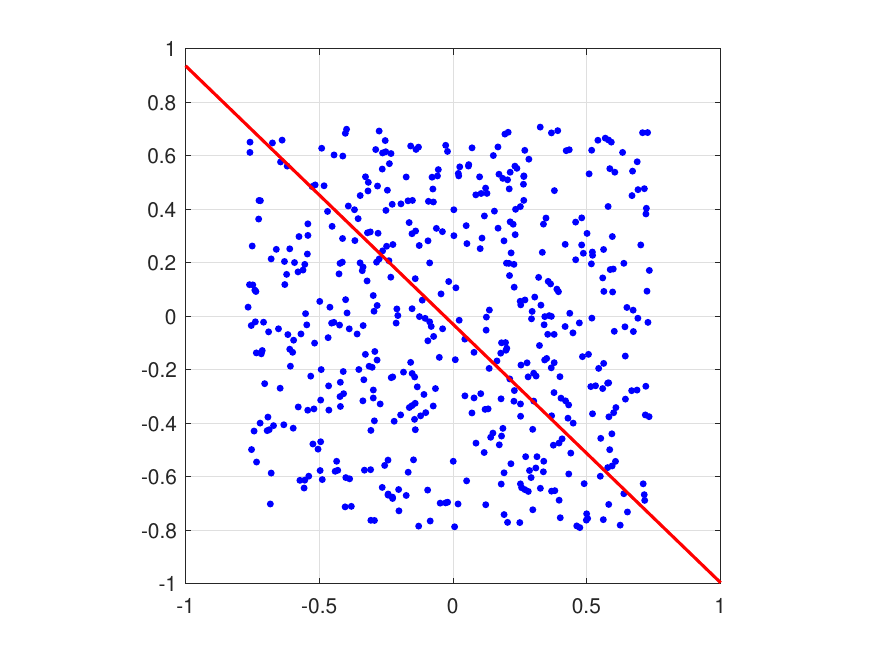} 
        \caption{$\beta_{2,1}(K)$=0.4146}
    \end{subfigure}

    \vspace{0.5cm} 
    \begin{subfigure}[b]{0.32\textwidth}
        \centering
        \includegraphics[width=1\linewidth]{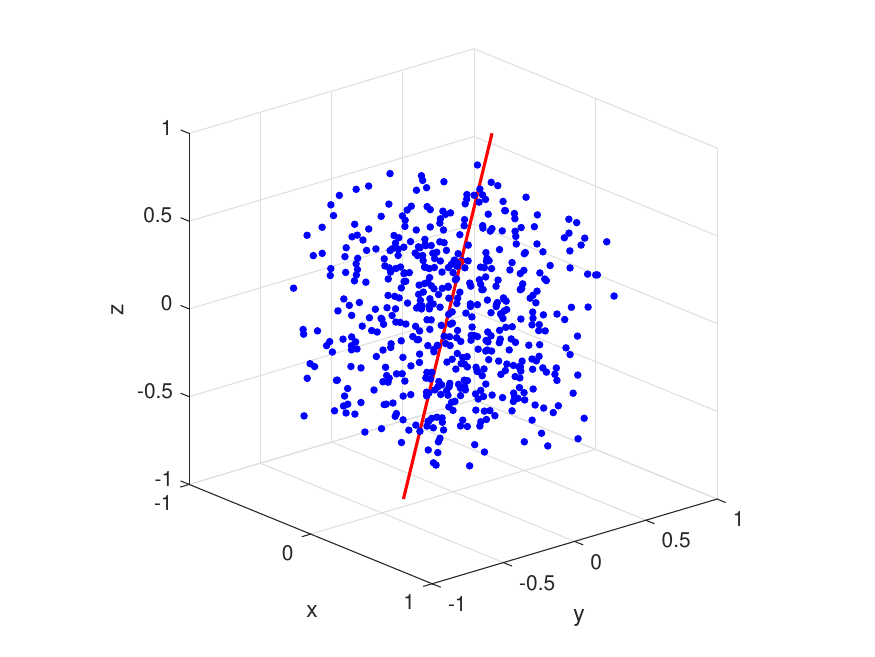} 
        \caption{$\beta_{\infty}(K)$=0.8381}
    \end{subfigure}
    \hfill
    \begin{subfigure}[b]{0.32\textwidth}
        \centering
        \includegraphics[width=1\linewidth]{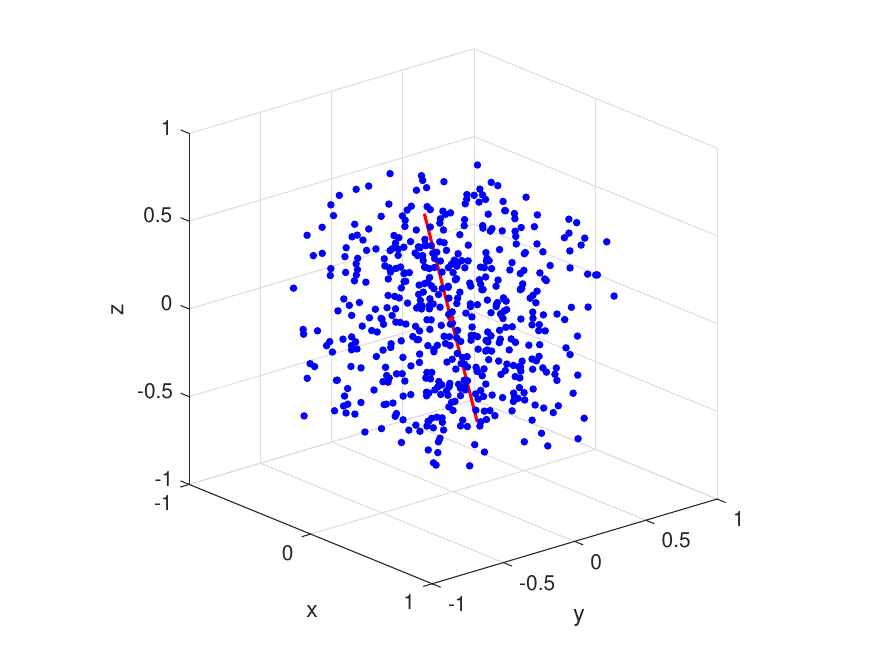} 
        \caption{$\beta_{2,1}(K)$=0.5004}
    \end{subfigure}
    \hfill
    \begin{subfigure}[b]{0.32\textwidth}
        \centering
        \includegraphics[width=1\linewidth]{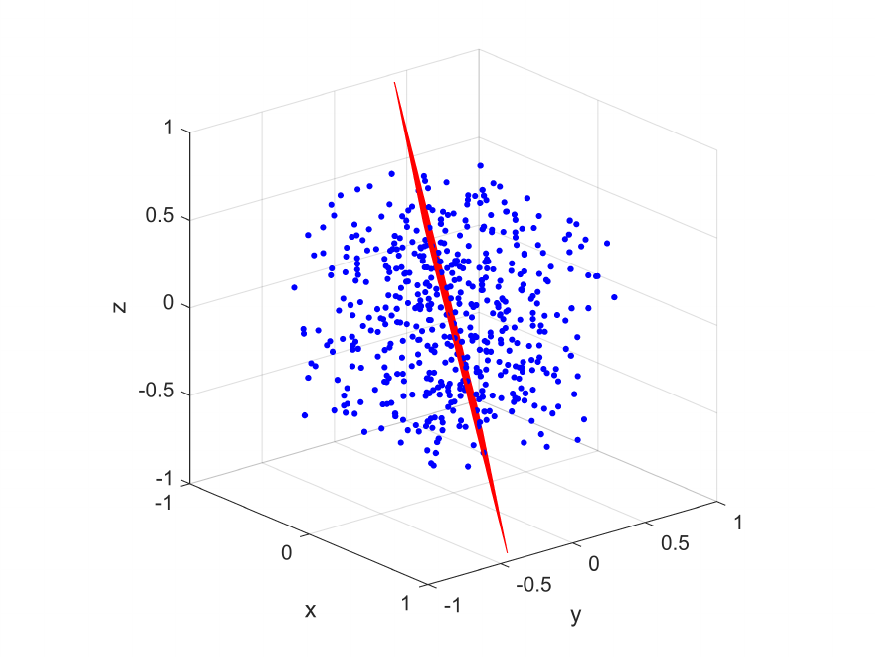} 
        \caption{$\beta_{2,2}(K)$=0.3440}
    \end{subfigure}

    \caption{Illustration of the computation results for finite discrete point sets. In each figure, data points are represented in blue, while the optimal subspace $L$ is represented in red. The corresponding $\beta$-number value is displayed at the bottom of each figure. The first row corresponds to finite discrete point sets in $\mathbb{R}^2$, while the second row corresponds to finite discrete point sets in $\mathbb{R}^3$.}
    \label{fig8}
\end{figure}

\subsection{Experiments on MNIST dataset}
In the previous section, we presented results from experiments conducted on toy examples within low-dimensional spaces. Here, we extend our analysis to the MNIST (Modified National Institute of Standards and Technology) dataset \cite{mnist}, utilizing $\beta$-number to effectively capture the shape characteristics of various handwritten digits. MNIST is a widely used handwritten digit dataset for training image processing systems. It consists of 70,000 handwritten digit images, representing digits from 0 to 9, evenly distributed across the dataset. Each image has a resolution of 28$\times$28 pixels and is presented as a single-channel grayscale image. The digits are handwritten by various individuals, resulting in a variety of writing styles, skewness, sizes, and spacing. Each image is accompanied by a corresponding label, ranging from 0 to 9, which denotes the digit contained within the image. Given the significant morphological differences between digits and the relatively high shape similarity among different handwriting styles of the same digit, we employ $\beta$-number to extract the morphological features of the handwritten digits. Specifically, for each image in the MNIST dataset, we first extract the coordinates of pixels with grayscale values greater than 150, then normalize these coordinates to the unit circle, forming a point set $K$ that represents the shape of the digit. Subsequently, we calculate the shape oscillation measures $\beta_{1,1}(K)$ for this two-dimensional point set $K$, with a schematic \cref{fig9} shown below:
\begin{figure}[htbp]
    \centering
    \begin{tabular}{ccccc}
         \\
        \subcaptionbox{\parbox{0.13\textwidth}{
        \centering
            \scriptsize $\beta_{1,1}(K)$=0.4725
        }}{\includegraphics[width=0.17\textwidth]{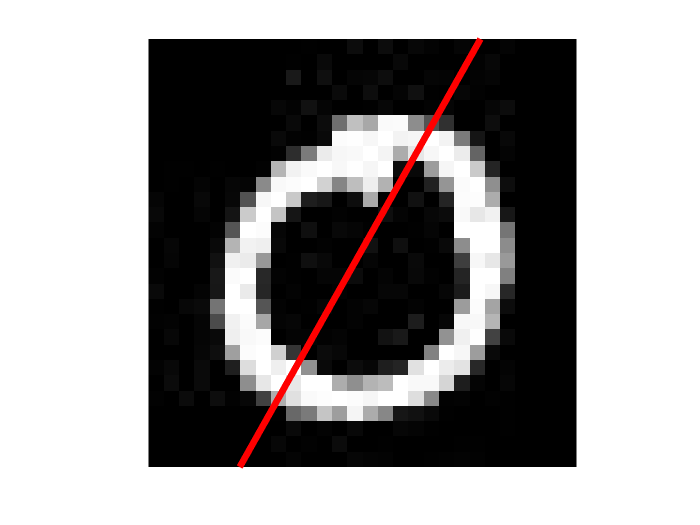}} &
        
        \subcaptionbox{\parbox{0.13\textwidth}{
        \centering
            \scriptsize  $\beta_{1,1}(K)$=0.0687
        }}{\includegraphics[width=0.17\textwidth]{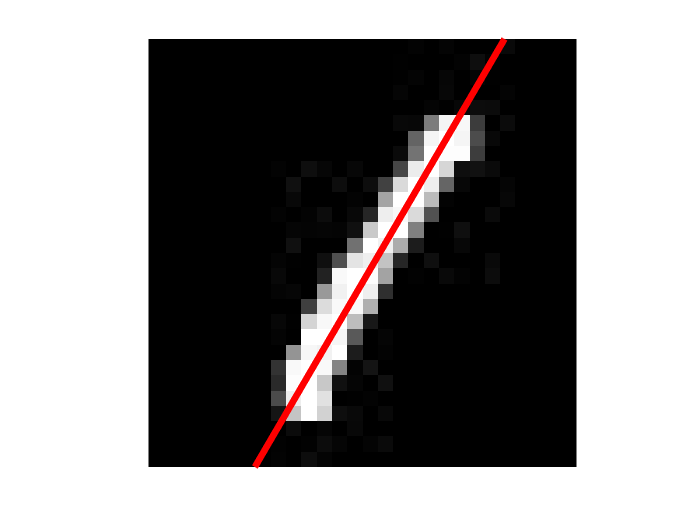}} &
        
        \subcaptionbox{\parbox{0.13\textwidth}{
        \centering
            \scriptsize  $\beta_{1,1}(K)$=0.3373
        }}{\includegraphics[width=0.17\textwidth]{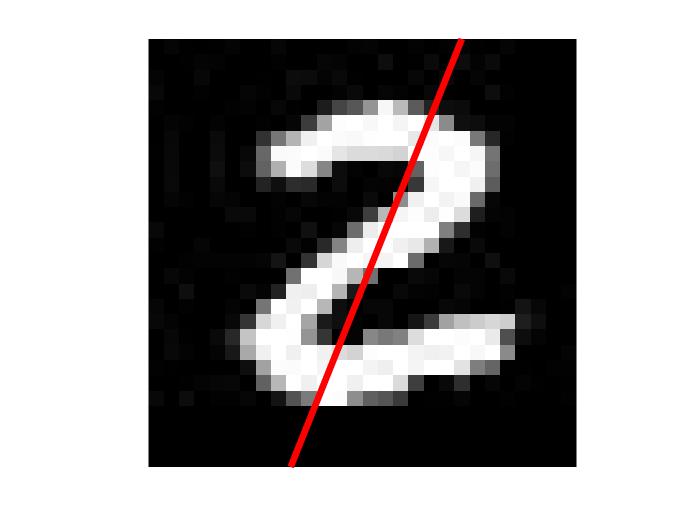}} &

        \subcaptionbox{\parbox{0.13\textwidth}{
        \centering
            \scriptsize  $\beta_{1,1}(K)$=0.2416
        }}{\includegraphics[width=0.17\textwidth]{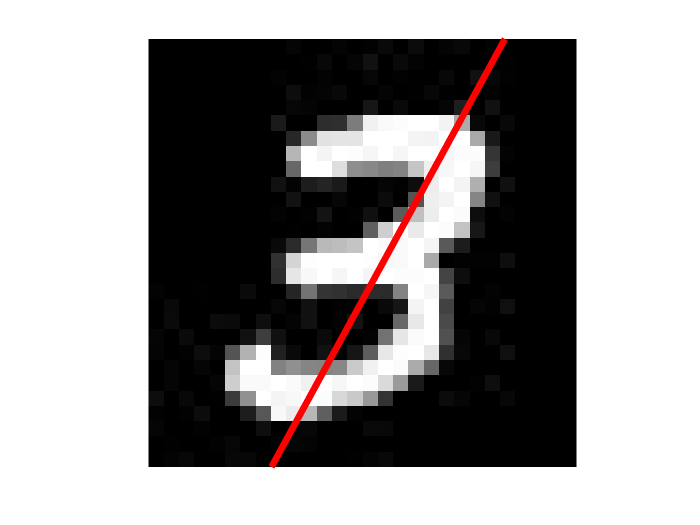}} &
        
        \subcaptionbox{\parbox{0.13\textwidth}{
        \centering
            \scriptsize  $\beta_{1,1}(K)$=0.2894
        }}{\includegraphics[width=0.17\textwidth]{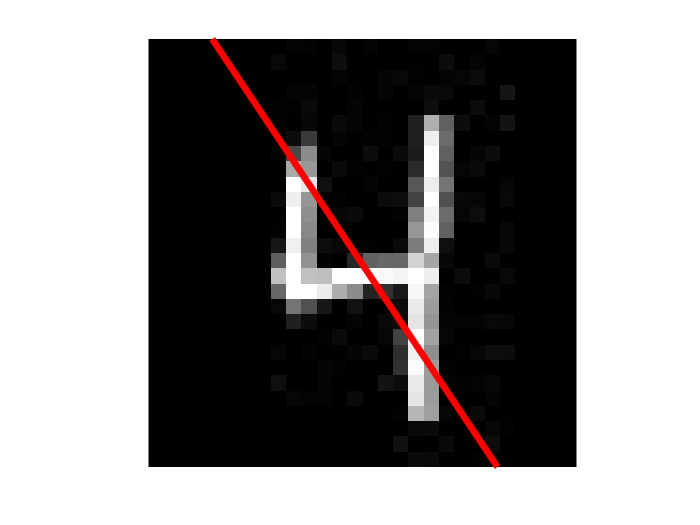}}
        
        \\
        \subcaptionbox{\parbox{0.13\textwidth}{
        \centering
            \scriptsize  $\beta_{1,1}(K)$=0.1802
        }}{\includegraphics[width=0.17\textwidth]{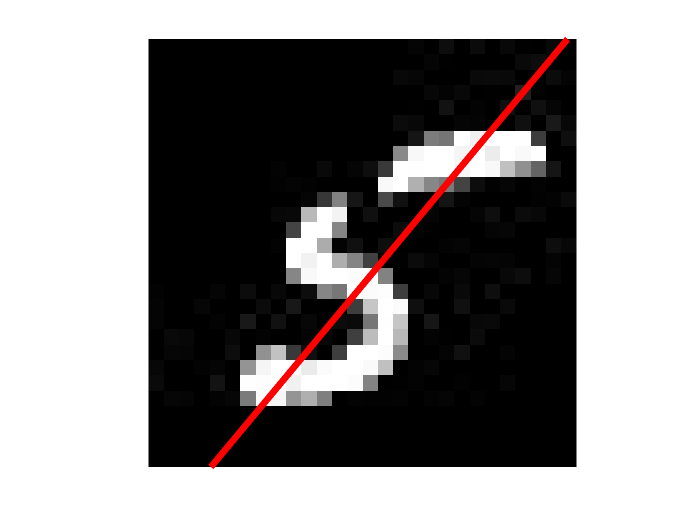}} 
&       
        \subcaptionbox{\parbox{0.13\textwidth}{
        \centering
            \scriptsize  $\beta_{1,1}(K)$=0.2627
        }}{\includegraphics[width=0.17\textwidth]{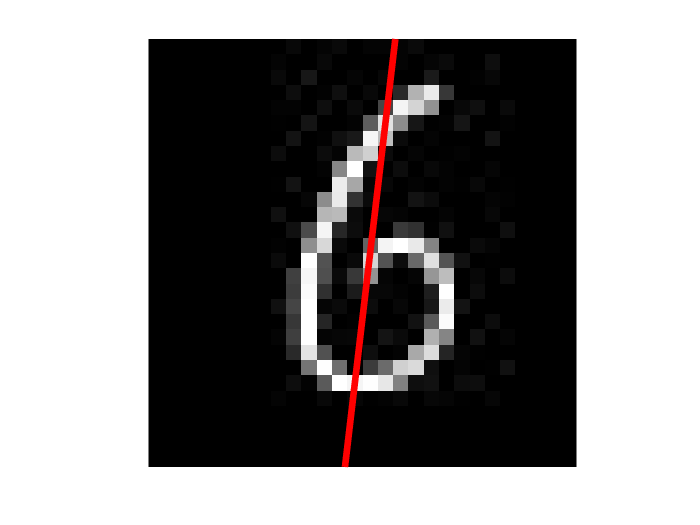}} &
        
        \subcaptionbox{\parbox{0.13\textwidth}{
        \centering
            \scriptsize  $\beta_{1,1}(K)$=0.2512
        }}{\includegraphics[width=0.17\textwidth]{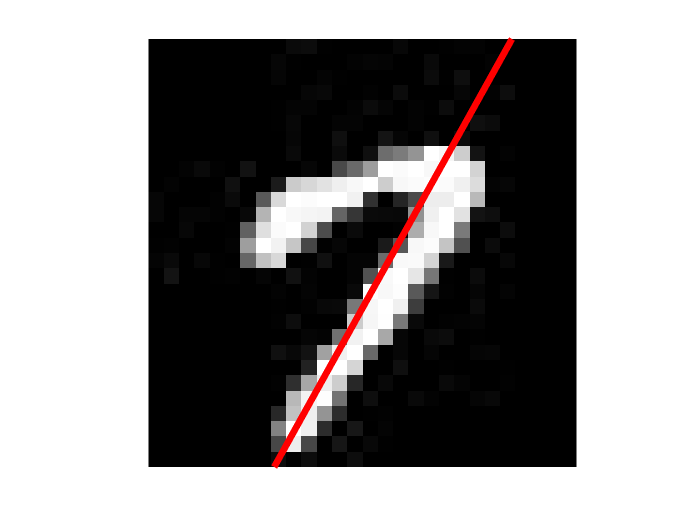}} &

        \subcaptionbox{\parbox{0.13\textwidth}{
        \centering
            \scriptsize  $\beta_{1,1}(K)$=0.1802
        }}{\includegraphics[width=0.17\textwidth]{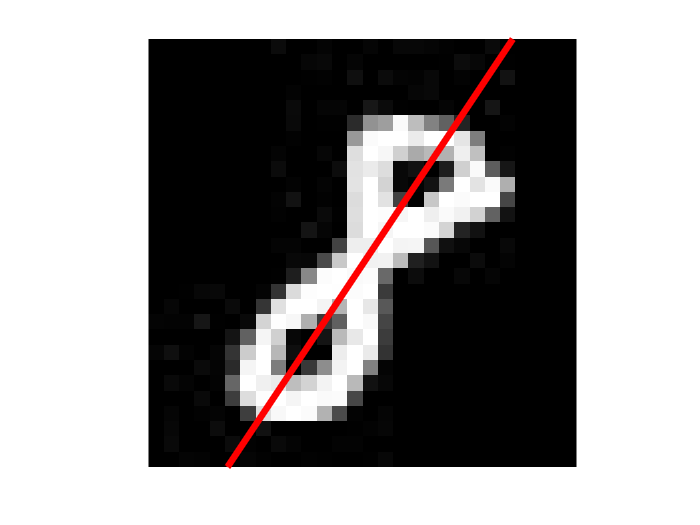}} &
        
        \subcaptionbox{\parbox{0.13\textwidth}{
        \centering
            \scriptsize  $\beta_{1,1}(K)$=0.2638}}{\includegraphics[width=0.17\textwidth]{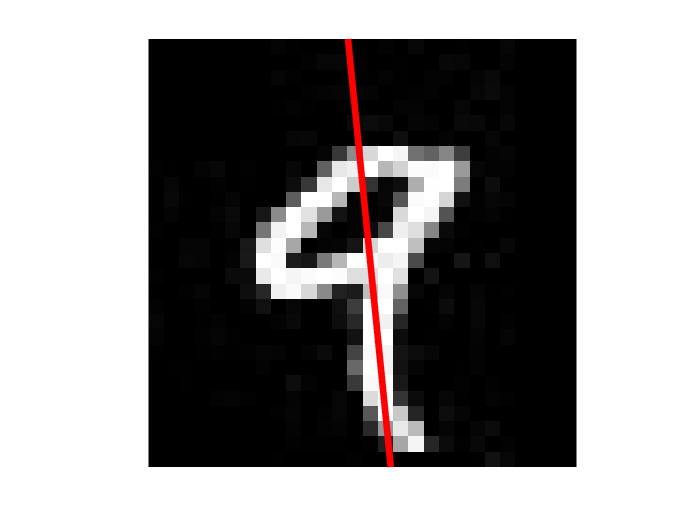}} 
    \end{tabular}
    \caption{Illustration of the computation results for the MNIST dataset. In each figure, red line represents the optimal fitting line $L$ when $p=1$. At the bottom of each diagram, the corresponding $\beta$-number is displayed, which characterizes the morphological features of the image. 
}
    \label{fig9}
\end{figure}

From \cref{fig9}, it is evident that the shape of the digit 1 closely resembles a straight line, resulting in a significantly lower $\beta$-number compared to other digits, while the $\beta$-number for digits 0 is markedly higher. The computational results obtained from the 10,000 training images in the MNIST dataset are shown in \cref{fig10}.
\begin{figure}[ht]
    \centering
    \begin{subfigure}{.5\textwidth}
        \centering
        \includegraphics[width=1\textwidth]{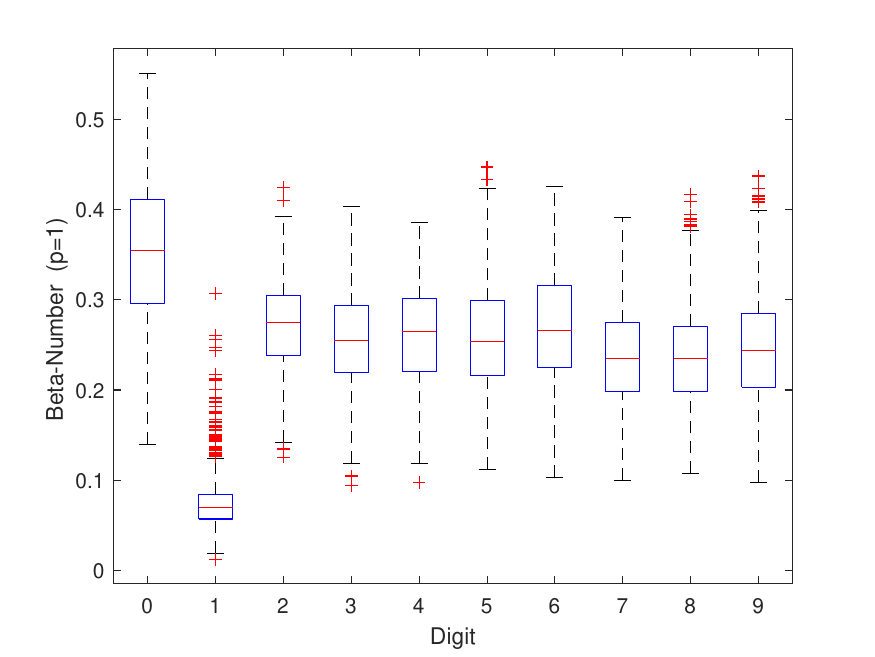}
        \caption{Boxplot of $\beta$-number ($p$=1) by digit}
    \end{subfigure}%
    \begin{subfigure}{.5\textwidth}
        \centering
        \includegraphics[width=1\textwidth]{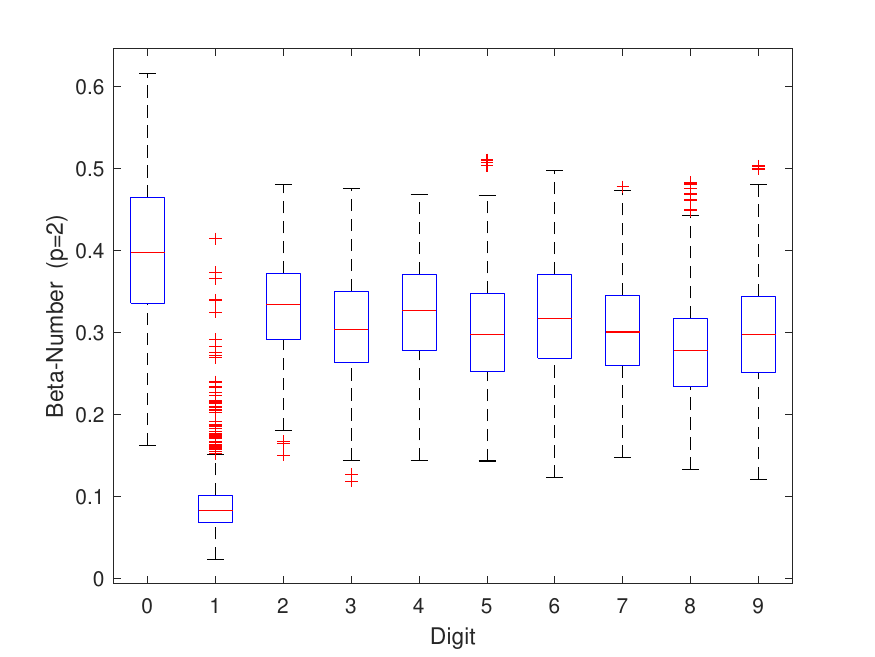}
        \caption{Boxplot of $\beta$-number ($p$=2) by digit}
    \end{subfigure}
    \caption{Boxplot of $\beta$-number derived from 10,000 MNIST samples, plotted against the image labels. The x-axis represents the digit labels (ranging from 0 to 9), and the y-axis represents the corresponding $\beta$-number (for both $p=1$ and $p=2$ cases).}
    \label{fig10}
\end{figure}

From \cref{fig10}, it is evident that, for both $p=1$ and $p=2$, $\beta$-number of digit "1" is significantly smaller than that of the other digits. Furthermore, when $p=1$, the $\beta$-number of digit "0" is significantly larger than that of the other digits. 
Except for digits "0" and "1", no significant differences are observed between the boxplots of the remaining digits. This indicates that the shape oscillation of the other eight digits are similar, lying between those of digits "1" and "0". Furthermore, for digits "0" and "1", when $p=1$, we observe that the boxplot for digit "0" has a notably longer box and dashed line compared to the other digits, suggesting that the shape variability of digit "0" is greater across different handwriting styles. On the other hand, digit "1" exhibits more outliers, indicating that many instances of digit "1" show abnormal fluctuations or noise. To further investigate this phenomenon, we extract the images corresponding to the maximum and minimum values of $\beta_{1,1}$ for digits "0" and "1", as presented in the \cref{fig11}.
\begin{figure}[htbp]
    \centering
    \begin{tabular}{cccc}
         \\
        \subcaptionbox{$\beta_{1,1}(K)$=0.1399}{\includegraphics[width=0.22\textwidth]{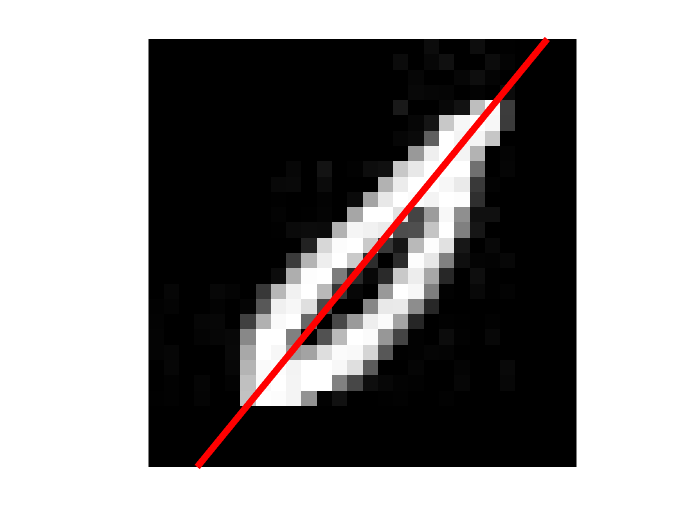}} &

        \subcaptionbox{$\beta_{1,1}(K)$=0.5510}{\includegraphics[width=0.22\textwidth]{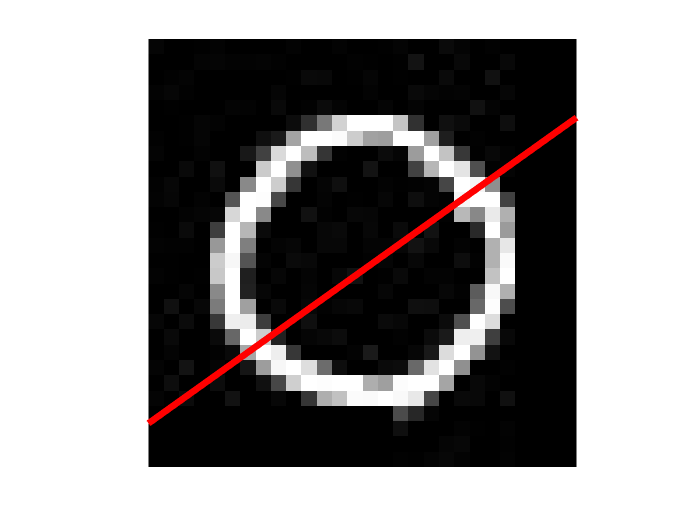}} &
        
        \subcaptionbox{$\beta_{1,1}(K)$=0.0123}{\includegraphics[width=0.22\textwidth]{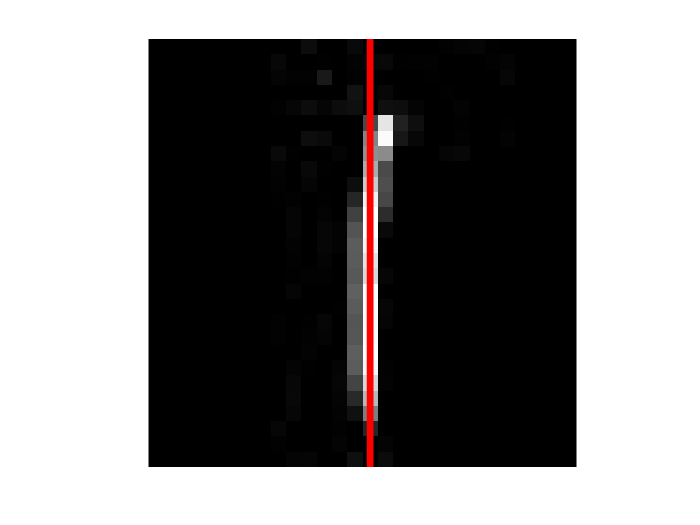}} &     \subcaptionbox{$\beta_{1,1}(K)$=0.3068 \label{fig11d}}{\includegraphics[width=0.22\textwidth]{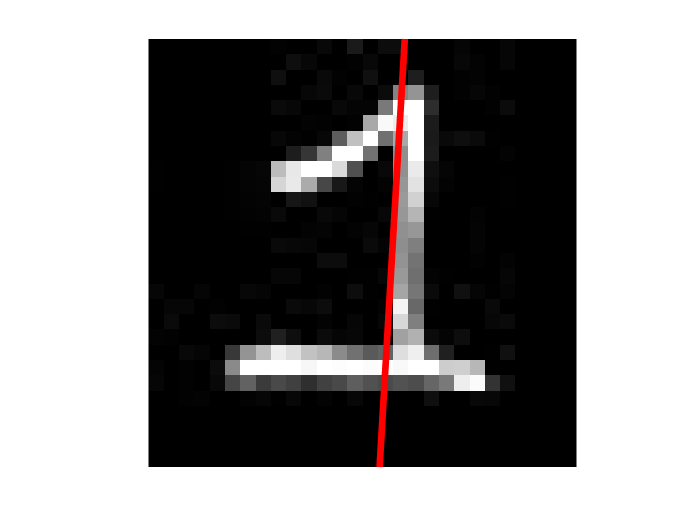}}
    \end{tabular}
    \caption{The corresponding images of the minimum and maximum $\beta_{1,1}$ values for digits "0" and "1". In each figure, red line represents the optimal fitting line $L$ when $p=1$, and the $\beta$-number displayed at the bottom characterizes the morphological features of the image.
}
    \label{fig11}
\end{figure}

As shown in \cref{fig11}, for digit "0", when its shape becomes more flattened and approaches a straight line, the shape oscillation decreases, resulting in a smaller $\beta_{1,1}$. Conversely, when the shape approaches a more circular form, the oscillation increases, leading to a larger $\beta_{1,1}$. For digit "1", due to handwriting style variations, some individuals write it in the form shown in \cref{fig11d}, exhibiting greater shape oscillation, which leads to a larger $\beta_{1,1}$ and the formation of outliers for digit "1".

In summary, $\beta$-number effectively captures the morphological characteristics of handwritten digits, specifically the $p$-th order one-dimensional shape oscillation degree as defined in \cref{sub 3.1}.

\label{lastpage}


\end{document}